\documentclass{article}

\usepackage[main, final, nonatbib]{neurips/neurips_2026} 

\usepackage[numbers]{natbib}
\usepackage{eccvabbrv}

\usepackage{graphicx}
\usepackage{booktabs}
\usepackage{amsmath,amssymb,amsfonts}
\usepackage{makecell}
\usepackage{multicol,multirow}
\usepackage{pgfplots}
\usepackage{adjustbox}
\usepackage{diagbox}
\usepackage{cite}
\usepackage{subcaption}
\usepackage{placeins}

\usepackage[accsupp]{axessibility}  
\usepackage[table,RGB]{xcolor}
\usepackage{tikz}
\usepackage{fontawesome5}

\usetikzlibrary{arrows, calc, shapes,  positioning, arrows.meta,3d, intersections, arrows.meta, decorations.markings}

\usepackage{custom_style}

\usepackage[pagebackref,breaklinks,colorlinks]{hyperref}
\usepackage{cleveref}

\usepackage{orcidlink}

\renewcommand{\cite}[1]{\citep{#1}}

\begin{document}

\title{Counting Trees from Satellite Imagery \\ with Noisy Supervision}



\author{
Dimitri Gominski\textsuperscript{1}\!\!\And
Maurice Mugabowindekwe \textsuperscript{1,2}\!\!\And
Qiue Xu \textsuperscript{3}\!\!\And 
Xiaowei Tong\textsuperscript{3}\!\!\And
Martin Brandt \textsuperscript{1}\!\!\And
Hieu Le \textsuperscript{4}\!\!\And
Rasmus Fensholt \textsuperscript{1}\!\!\And
Dimitris Samaras \textsuperscript{5}\!\!\And
Loic Landrieu \textsuperscript{6}\!\!\and \textsuperscript{1}University of Copenhagen \and \textsuperscript{2}University of Rwanda \and \textsuperscript{3}University of Chinese Academy of Sciences \and \textsuperscript{4}University of North Carolina at Charlotte \and \textsuperscript{5}Stony Brook University \and \textsuperscript{6}LIGM, CNRS, Univ Gustave Eiffel, ENPC, IPP }

\maketitle

\begin{abstract}
Counting individual trees is a fundamental task for environmental monitoring, yet remains largely unexplored with satellite imagery.
At these resolutions, isolated trees may still be identifiable, but crown boundaries become ambiguous in dense forests, making the notion of an individual tree inherently ill-defined.
Moreover, large-scale manual annotations of individual trees are prohibitively expensive. While scalable supervision can be derived from airborne LiDAR, the resulting annotations are noisy and difficult to exploit effectively.
We address these challenges by formulating tree counting as a spatial density matching problem supervised through Unbalanced Optimal Transport.
This formulation naturally accommodates both precise localization of isolate trees and robust density estimation in dense forests.
We further introduce a self-correction mechanism that leverages transport residuals to progressively refine noisy supervision during training.
We evaluate our approach on \textsc{TinyTrees}, a new benchmark spanning three continents and three satellite sensors, comprising over 216 million tree annotations (including 639k manually verified instances) across $25\,890\,\mathrm{km}^2$.
Our method consistently outperforms detection-based, regression-based, and transport-based distribution-matching baselines, demonstrating the effectiveness of unbalanced transport and reliability-aware supervision for large-scale tree counting from satellite imagery.
Code, data and models are available at \href{https://github.com/dgominski/treematch}{github.com/dgominski/treematch}.
\end{abstract}

\section{Introduction}
\label{sec:intro}

Most large-scale tree monitoring efforts focus on forest cover~\cite{hansen2013high}, biomass~\cite{santoro2021global}, or canopy height~\cite{tolan2024very,fogel_open-canopy_2025,pauls_estimating_2024}, which provide aggregate metrics and do not capture tree-level structure.
However, estimating the number and spatial distribution of individual trees is central to environmental monitoring and land management.
Indeed, many key applications require tree-level analysis: including reforestation tracking~\cite{cao2011greening}, biomass and carbon estimation~\cite{kankare_individual_2013}, biodiversity assessment~\cite{beloiu_individual_2023}, as well as the monitoring of anthropogenic tree structures such as hedgerows~\cite{muro_hedgerow_2025} and urban trees~\cite{ventura_individual_2024,beery_auto_2022}.

Despite this, large-scale tree counting from satellite imagery remains largely unexplored, with existing datasets restricted to small-scale aerial imagery and narrow geographic coverage~\cite{chen_transformer_2022, yao2021tree}, limiting both systematic evaluation and the development of dedicated methods.

\begin{figure}[t]
    \centering
    \begin{tabular}{c@{\;}c@{\;}c@{\;}c}
\begin{subfigure}{0.24\linewidth}
\includegraphics[width=\linewidth]{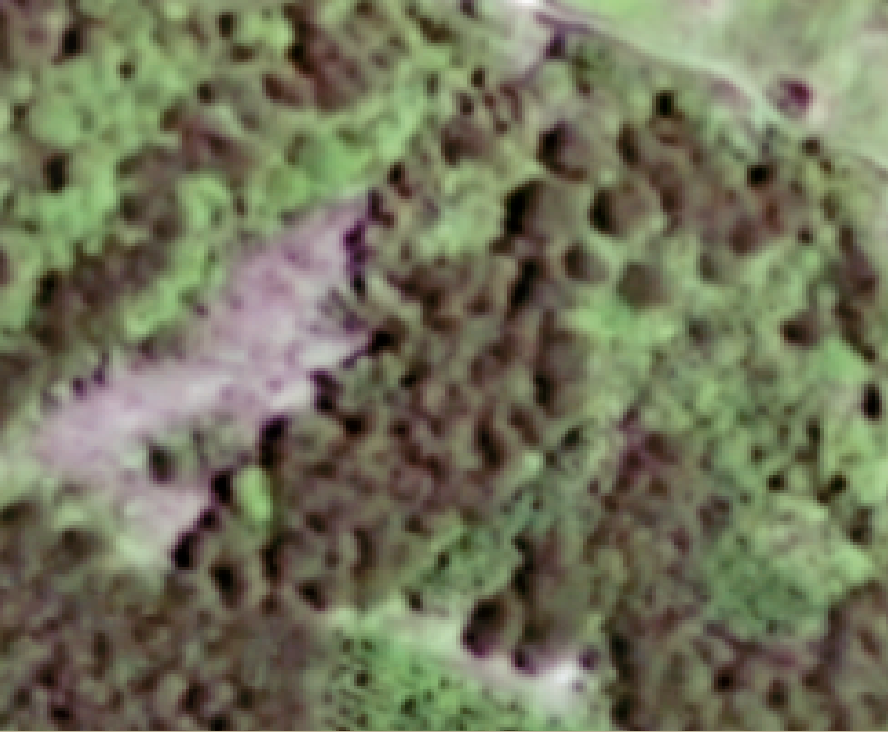}
\caption{Satellite Image \\~}
\label{fig:teaser:sat}
\end{subfigure}
&
\begin{subfigure}{0.24\linewidth}
\includegraphics[width=\linewidth]{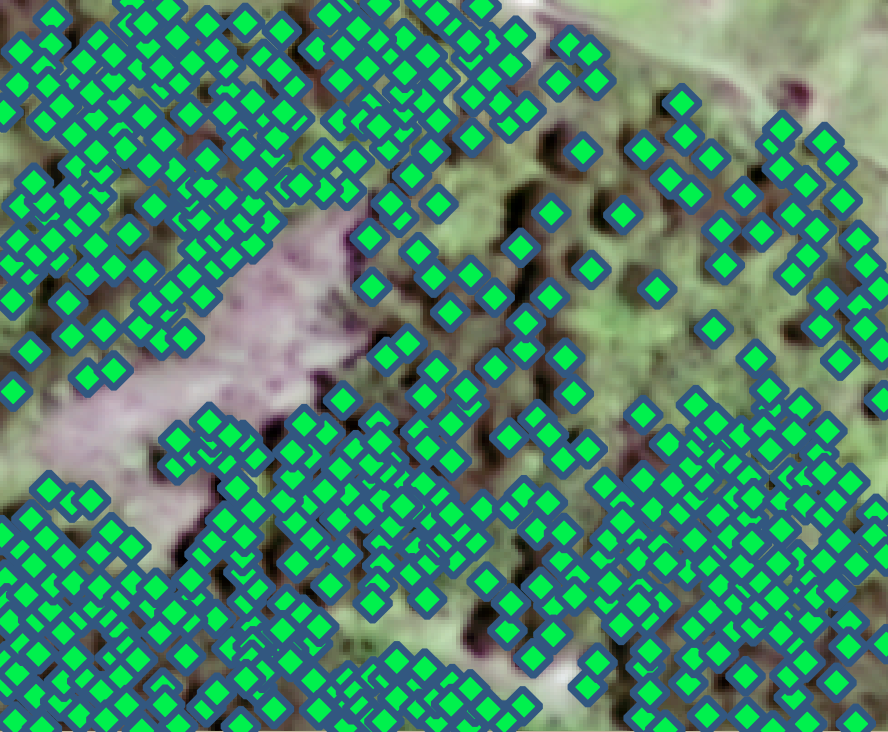}
\caption{Strong Labels \\ count: 361 trees}
\label{fig:teaser:strong}
\end{subfigure}
&
\begin{subfigure}{0.24\linewidth}
\includegraphics[width=\linewidth]{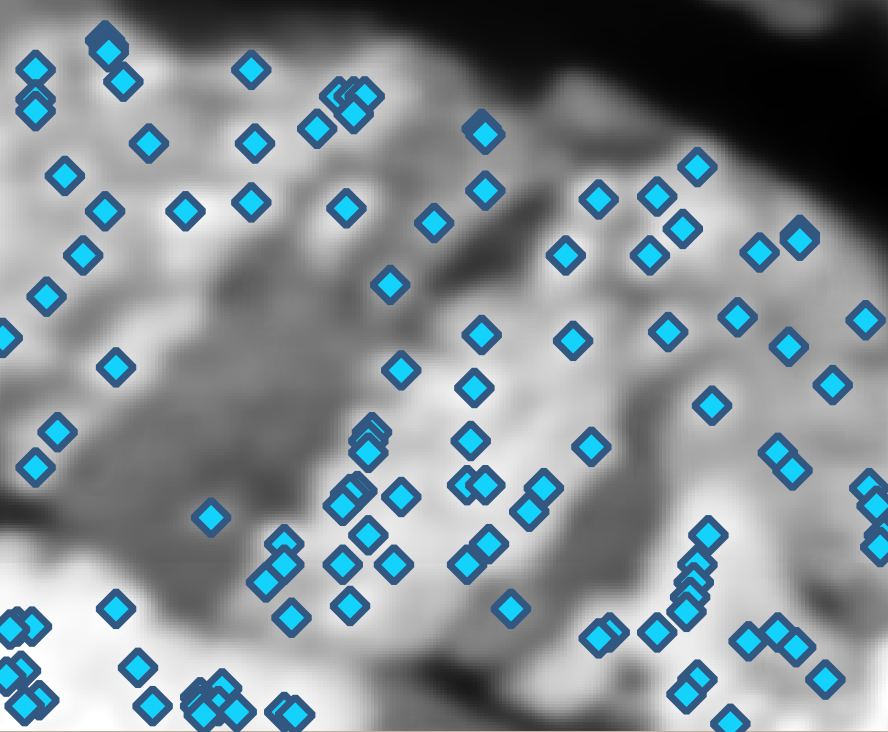}
\caption{Weak Labels \\ count: 77 trees}
\label{fig:teaser:weak}
\end{subfigure}
&
\begin{subfigure}{0.24\linewidth}
\includegraphics[width=\linewidth]{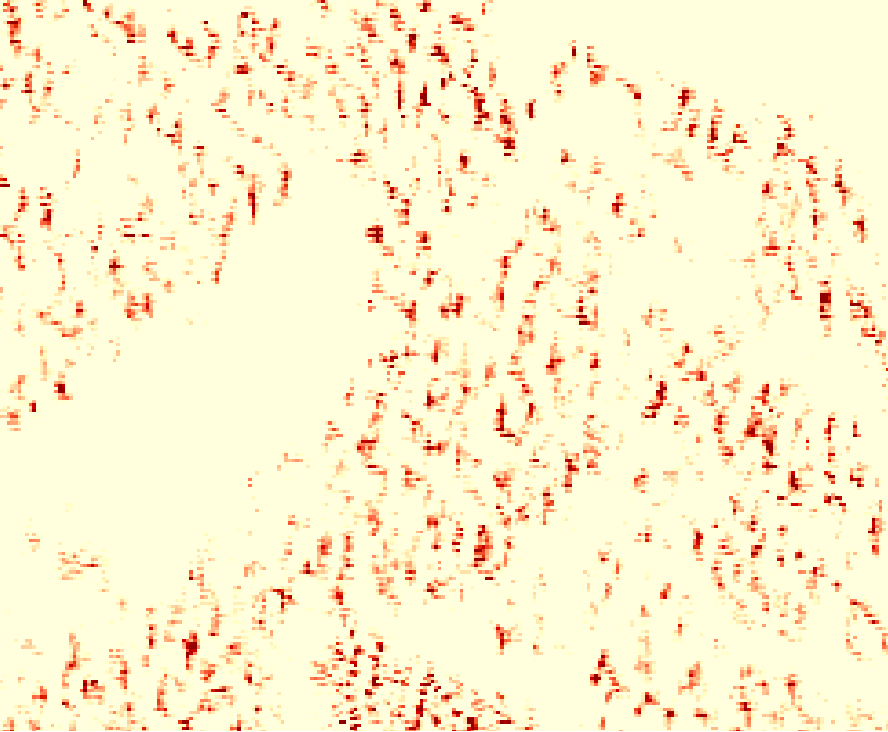}
\caption{Predicted Density \\ count: 395 trees}
\label{fig:teaser:pred}
\end{subfigure}
\end{tabular}
    \vspace{-3mm}
    \caption{{\bf Counting Trees From Space.}
    We introduce \textsc{TreeMatch}, a model to estimate the number of trees in a VHR satellite image (\subref{fig:teaser:sat}).
    Training relies on two supervision sources:
    \emph{strong labels} consisting of manually annotated tree centers (\subref{fig:teaser:strong}),
    and \emph{weak labels} automatically derived from airborne laser scanning (\subref{fig:teaser:weak}).
    The model predicts a per-pixel tree density map (\subref{fig:teaser:pred}), enabling precise localization of isolated trees while accommodating dense forest regions where individual boundaries are ambiguous.}
    \label{fig:teaser}
\end{figure}

\paragraph{\textbf{ Challenges.}}
Counting trees from satellite imagery raises several unique technical challenges.
Tree crowns vary widely in size, from less than $1\,\mathrm{m}^2$ to several hundred square meters, and do not always admit a clear object center.
They occur in heterogeneous spatial configurations, ranging from isolated individuals to dense forest canopies.
At satellite resolution, individual crowns often span only a few pixels, and sensor blur can merge neighboring trees.
As a result, delineation becomes ill-defined in dense regions while remaining feasible for isolated trees.
Because real-world scenes combine separable instances, dense clusters, and intermediate regimes, tree counting is difficult to address with a single globally consistent formulation.

Precise manual annotations for tree counting are costly and typically limited in scale.
However, large-scale supervision can be derived automatically from Airborne Laser Scanning (ALS), albeit with substantially lower reliability.
Effective learning at the satellite scale therefore requires formulations that explicitly accommodate noisy and heterogeneous annotations.

\paragraph{\textbf{Contributions.}}
To address these challenges, we make three contributions.
First, we introduce \textsc{TinyTrees}, the first large-scale benchmark for tree counting from satellite imagery.
The dataset spans three continents and three satellite sensors, covers $25\,890\,\mathrm{km}^2$, and contains more than 216 million tree annotations, including 639k manually verified instances.
Second, we propose \textsc{TreeMatch}, a density-matching framework based on unbalanced optimal transport.
This formulation naturally handles heterogeneous counting regimes, enabling precise localization of isolated trees while remaining robust in dense canopies where instance boundaries are ambiguous.
Third, we leverage marginal residuals of the unbalanced transport solution to automatically correct weak annotations during training.
This mechanism enables effective learning from large-scale noisy supervision alongside sparse high-quality labels, providing a principled solution to heterogeneous annotation quality.

\paragraph{\textbf{ Results.}}
Across three continents and three satellite sensors, \textsc{TreeMatch} consistently outperforms detection-based, regression-based, and distribution-matching baselines.
Our formulation demonstrates improved robustness to weak supervision and maintains strong performance across diverse forest types and imaging conditions.

\section{Related Work}
\label{sec:related}

\begin{table}[t]
\caption{{\bf Tree-Counting Datasets.}
\textsc{TinyTrees} substantially expands geographic coverage, annotation scale, and sensor diversity compared to existing datasets.
GSD denotes Ground Sampling Distance (spatial resolution). $\dagger$ data not accessible.}
\resizebox{\linewidth}{!}{
\begin{tabular}{l cccc@{\quad}c}
 \toprule
 & \multirow{2}{*}{Location} & \multirow{2}{*}{\makecell{sensor\\ GSD (m)}} & \multirow{2}{*}{\makecell{area\\  (km²)}} & \multicolumn{2}{c}{labels}\\[-1mm]\cmidrule(lr){5-6}
   &&  & & manual & automatic\\
\midrule
Yosemite Tree \cite{chen_transformer_2022} & California & aerial (0.1m) & \hphantom{1}$\sim$10 & $\sim$100k & -\\
\rowcolor{gray!10}$\dagger$ VHRTrees \cite{topgul2025vhrtrees} & Turkey & Google earth (0.5m) & \hphantom{$\sim$1}23 & \hphantom{$\sim$1}26k & - \\
KCL-London & London, UK & aerial (0.2m) & \hphantom{$\sim$1}38 & \hphantom{$\sim$1}95k & - \\
\rowcolor{gray!10}$\dagger$ GF-II \cite{yao2021tree} & China & Gaofen2 (0.8m) & \hphantom{$\sim$1}56 & \hphantom{1}$\geq$100k &  \\
\textsc{DRIFT} \cite{li_get_2024} & France & \makecell{aerial, SkySat \\ (0.2-0.5m)} & \hphantom{$\sim$1}62 & - & 526k \\
\rowcolor{gray!10}\textsc{NEON} Benchmark \cite{weinstein_benchmark_2021} & USA & aerial (0.1m) & $\leq$100 & \hphantom{$\sim$1}12k & 9.3M  \\\greyrule
\bf \multirow{1}{*}{\textsc{TinyTrees} (ours)}
& \makecell{China, Africa,\\ France} & \makecell{Gaofen2, Planet,\\ SPOT6 (0.8-4.2m)} & \bf 25.9k &\bf \hphantom{$\sim$}639k & \bf 215M \\\bottomrule
\end{tabular}
}
\end{table}

\paragraph{\textbf{ Object Counting.}}
Object counting is a well-established problem in computer vision.
A common approach converts object annotations into density maps by convolving point locations with Gaussian kernels and training models using pixel-wise regression losses~\cite{lian_density_2019, onoro-rubio_towards_2016}.
Subsequent work adapts kernel sizes based on local density~\cite{wan_adaptive_2019}.
Alternatively, detection-based methods localize instances via bounding boxes~\cite{liu_point_2019} or points~\cite{song_rethinking_2021}.
While effective when object boundaries are distinct, detection-based approaches degrade under heavy overlap or clutter, where separability assumptions fail—conditions typical of dense forest canopies.

\paragraph{\textbf{ Tree Mapping.}}
Recent advances in remote sensing have enabled large-scale tree analysis, including isolated tree detection~\cite{tucker_sub-continental-scale_2023}, canopy height estimation~\cite{lang_high-resolution_2023,tolan2024very}, and species mapping~\cite{mu_national-scale_2025}.
These efforts are supported by public datasets providing crown delineations, point annotations~\cite{veitch-michaelis_oam-tcd_2024}, LiDAR point clouds~\cite{ign_lidar_nodate}, and species labels~\cite{gaydon_pureforest_2024, ahlswede_treesatai_2023}.
However, tree counting itself is typically treated as a by-product of detection~\cite{brandt_unexpectedly_2020, chen_transformer_2022}, and no dedicated large-scale benchmark exists for satellite imagery.
Moreover, trees in continuous canopy exhibit overlapping and irregular crowns~\cite{voulgaris_bridging_2025}, making precise delineation difficult.
Although high-resolution data and specialized models can improve segmentation~\cite{wielgosz_segmentanytree_2024, firoze_tree_2023}, such approaches are costly and difficult to scale.
Consequently, recent work increasingly relies on airborne LiDAR to derive large-scale supervision~\cite{gominski_trees_2026,fogel_open-canopy_2025}.
Yet LiDAR-derived labels introduce structured and correlated errors, motivating noise-robust formulations.

\paragraph{\textbf{Distribution Matching for Counting.}}
Crowd counting faces similar challenges of density variation, overlap, and annotation uncertainty.
This has motivated probabilistic formulations that replace strict pixel-wise regression with distribution-matching objectives~\cite{ma_bayesian_2019, wang_distribution_2020}, improving robustness to spatial ambiguity.
Closest to out work, Ma \etal \cite{ma2021learning} propose to use UOT for crowd counting.
Despite these parallels, distribution-matching approaches have not been systematically explored for large-scale tree counting from satellite imagery.

\section{Method}    
\label{sec:method}

We first review unbalanced optimal transport (\cref{sec:primer}), then describe how it can be used for tree density regression (\cref{sec:ottree}). We then present a method to bootstrap the results of marginal transport to correct weak annotations (\cref{sec:slack}). Finally, we present our approach to leverage both strong and weak tree annotations to train our model (\cref{sec:weak}).

\subsection{Primer on Unbalanced Optimal Transport}
\label{sec:primer}
We briefly review optimal transport and its unbalanced variant, focusing on the elements required for our formulation. Readers already familiar with optimal transport may skip this subsection.

\paragraph{\textbf{Optimal Transport.}}
Optimal Transport (OT)~\cite{villani_optimal_2009} measures the cost of transforming one probability distribution into another while accounting for the geometry of the underlying space.
Let $\mu \in \mathbb{R}_+^n$ and $\nu \in \mathbb{R}_+^n$ be two discrete probability distributions satisfying
$\mathbf{1}^\top \mu = \mathbf{1}^\top \nu = 1$, where $\mathbf{1}$ denotes a vector of ones.
Let $C \in \mathbb{R}_+^{n \times n}$ be a cost matrix, where $C_{ij}$ represents the cost of transporting a unit mass from location $i$ to location $j$.
OT seeks a transport plan $\gamma$ that matches the marginals of $\mu$ and $\nu$:
$
\Gamma(\mu,\nu)
=
\left\{
\gamma \in \mathbb{R}_+^{n \times n}
\;\middle|\;
\gamma \mathbf{1} = \mu,\;
\gamma^\top \mathbf{1} = \nu
\right\}.
$
The Monge--Kantorovich OT problem is then defined as
\begin{align}
W^\text{OT}_C(\mu,\nu)
=
\min_{\gamma \in \Gamma(\mu,\nu)}
\langle C, \gamma \rangle,
\label{eq:ot-primal}
\end{align}
where $\langle C, \gamma \rangle = \sum_{i,j} C_{ij} \gamma_{ij}$.
Intuitively, OT computes the minimum effort required to reshape one distribution into the other.

\paragraph{\textbf{Entropy-Regularized Optimal Transport.}}
Solving~\eqref{eq:ot-primal} exactly is computationally expensive.
Cuturi \etal~\cite{cuturi_sinkhorn_2013} introduce a negative entropy penalty, leading to the relaxed objective
\begin{align}
W^\text{ROT}_{C,\varepsilon}(\mu,\nu)
=
\min_{\gamma \in \Gamma(\mu,\nu)}
\langle C, \gamma \rangle
+
\varepsilon \, \mathrm{H}(\gamma),
\qquad
\mathrm{H}(\gamma)
=
\sum_{i,j} \gamma_{ij} \left( \log \gamma_{ij} - 1 \right),
\label{eq:rot}
\end{align}
where $\varepsilon > 0$ controls the smoothness of the transport plan.
The solution of~\eqref{eq:rot} can be efficiently approximated using Sinkhorn iterations.
This procedure is fast, GPU-friendly, and differentiable, which makes entropy-regularized OT
widely used as a loss in deep learning.

\paragraph{\textbf{Unbalanced Optimal Transport.}}
In many applications, including density regression, the two distributions may not have the same total mass.
Enforcing strict marginal constraints is therefore unrealistic and can destabilize optimization.
Unbalanced Optimal Transport (UOT)~\cite{chizat_scaling_2018} relaxes these constraints by allowing mass creation or destruction at a controlled cost.
Given two nonnegative measures $\mu$ and $\nu$ with arbitrary total mass, UOT solves
\begin{align}\label{eq:uot}
W_{C,\varepsilon,\tau}^\text{UOT}(\mu,\nu)
=
\min_{\gamma \ge 0}
\;
\langle C, \gamma \rangle
+
\varepsilon \, \mathrm{H}(\gamma)
+
\tau \, \text{KL}\!\left( \gamma \mathbf{1} \,\|\, \mu \right)
+
\tau \, \text{KL}\!\left( \gamma^\top \mathbf{1} \,\|\, \nu \right).
\end{align}
where $\tau>0$ controls the penalty for deviating from mass conservation with respect to the source distribution $\mu$ and the target distribution $\nu$.
and $\text{KL}(\cdot\,\|\,\cdot)$ denotes the generalized Kullback--Leibler divergence
between nonnegative measures \cite{chizat_scaling_2018}.

\begin{figure}[t]
    \centering
    \input{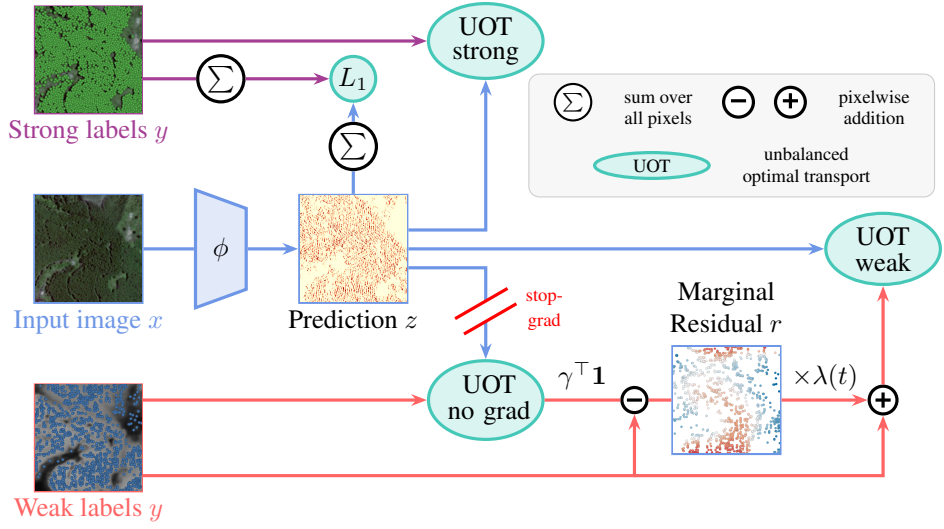}
\caption{{\bf \textsc{TreeMatch}.}
Given a VHR satellite image, the network $\phi$ predicts a continuous tree density map.
Training uses two supervision sources:
\protect\legendline{stronglbl} strong labels (manual point annotations) and
\protect\legendline{weaklbl} weak labels (automatically derived tree-cover masks).
Strong labels supervise both a global count loss and a UOT-based alignment loss.
For weak labels, we first solve a UOT problem (without backpropagation) to compute marginal residuals.
These residuals are used to correct the weak annotations.
A second UOT loss is then applied between the corrected labels and the prediction, from which gradients are propagated to update the model.}
    \label{fig:pipeline}
\end{figure}

\subsection{Learning to Count Trees with Optimal Transport}
\label{sec:ottree}

Optimal transport offers a natural way to compare spatial measures by jointly accounting for geometry and mass distribution. The unbalanced formulation is especially appropriate in our setting, as it decouples geometric alignment from exact mass conservation, thereby accommodating counting errors and annotation noise.

\paragraph{\textbf{Setting.}}
We consider an input image $x \in \mathbb{R}^{H \times W \times C}$ defined on a pixel grid of size $N = H \times W$.
Ground-truth annotations are represented as a nonnegative density map $y \in \mathbb{R}_+^{N}$.
Depending on the supervision source, $y$ may either encode a dense tree-cover map or a sparse binary map, where $y_i = 1$ indicates the annotated center of a tree crown at pixel $i$.
This representation defines a discrete measure over the image domain, rather than a probability distribution.
The total number of annotated trees is given by $\sum_i y_i$.

We train a model $\phi$ that maps $x$ to a nonnegative density map
$z \in \mathbb{R}_+^{N}$,
where $z_i$ represents the predicted mass at pixel $i$.
As for $y$, the prediction is not normalized.
The predicted tree count is obtained by summation, $\sum_i z_i$.

\paragraph{\textbf{Count Supervision.}}
A first supervision signal enforces consistency between predicted and ground-truth counts:
\begin{align}
\mathcal{L}^{\text{count}}(z,y)
=
\left|
\sum_i z_i
-
\sum_i y_i
\right|.
\end{align}
While this loss enforces global count consistency, it provides no spatial supervision.

\paragraph{\textbf{Density Matching with Unbalanced Optimal Transport.}}
Directly comparing the predicted density $z$ and the annotated measure $y$ using a KL divergence is inappropriate in our setting.
Indeed, KL heavily penalizes spatial misalignment: assigning mass to a nearby but unannotated pixel can incur arbitrarily large loss. In our setting, the precise location of tree centers is inherently ambiguous, especially in dense canopies.
Instead, we align $z$ and $y$ using Optimal Transport, which compares spatial measures while explicitly accounting for geometric proximity.
We define the cost matrix $C$ using squared Euclidean distances between pixel locations:
\begin{align}
C_{ij}
=
\left\|
\pos(i) - \pos(j)
\right\|_2^2,
\end{align}
where $\pos(i)$ denotes the 2D coordinates of pixel $i$.
Transporting mass over long distances is penalized, encouraging the predicted density to concentrate near annotated locations while tolerating small localization shifts.

DM-Count~\cite{wang_distribution_2020} applies classical (balanced) OT to crowd counting by normalizing both the prediction and the ground truth into probability distributions.
This normalization implicitly enforces equal total mass between the two measures.
In our setting, however, annotations may be noisy or incomplete, and predicted counts may differ substantially from ground truth.
We instead adopt \emph{Unbalanced} Optimal Transport (UOT), which relaxes the mass conservation constraint.
UOT allows mass to be locally created or removed at a controlled penalty, decoupling geometric alignment from exact mass equality.
The loss is defined as
\begin{align}
\mathcal{L}^{\text{UOT}}
=
W_{C,\varepsilon,\tau}^{\text{UOT}}
\left(
z,
y
\right),
\end{align}
where $\varepsilon$ controls entropic regularization and $\tau$ governs tolerance to mass discrepancies.
This formulation preserves the geometric alignment properties of OT while explicitly accommodating counting uncertainty.
\if 1 0
\subsection{Boostrapping Marginal Slack}
\label{sec:slack}
Tree annotations derived from automated pipelines are inherently noisy.
They may contain spurious detections in dense forest cover while remaining reliable for isolated trees.
This uncertainty is highly localized and varies across locations, making it difficult to address with a single global relaxation parameter or heuristic.
Rather than introducing external supervision or ad-hoc filtering rules, we leverage a signal that naturally emerges when solving the Unbalanced Optimal Transport problem.

\paragraph{\bf Slack as a Reliability Signal.}
Unbalanced Optimal Transport provides, as a by-product, a measure of how much ground-truth mass is discarded during matching.
We interpret this discarded mass, or \emph{marginal slack}, as a data-driven indicator of annotation reliability.
When solving the UOT problem between the predicted density $z$ and the ground-truth measure $y$, the optimizer may choose not to transport part of the ground-truth mass if aligning it with $z$ is geometrically or statistically implausible.
Under a reasonably calibrated model, annotations that are consistently discarded are therefore less likely to correspond to true trees.

Let $\gamma$ denote the optimal transport plan obtained from UOT between $z$ and $y$.
The transported ground-truth mass corresponds to the column marginals $\gamma^\top \mathbf{1}$.
We define the \emph{marginal slack} on the ground-truth side as
\begin{align}
    s(y,\gamma)
    &=
    \max\!\left(0,\; y - \gamma^\top \mathbf{1}\right), \\
\end{align}
computed elementwise.
For an annotated pixel $i$ with $y_i=1$, $s_i=0$ indicates that the annotation is fully consistent with the prediction, while larger values indicate that the annotation is not explained by the model and is effectively discarded during the transport computation.

\paragraph{\bf Confidence from Slack.}
We convert the slack into a per-annotation confidence weight:
\begin{align}\label{eq:confidence}
    c_i = \alpha + (1-\alpha)(1-s_i),
\end{align}
where $\alpha>0$ defines a minimum confidence floor.
Annotations with small slack therefore receive weights close to $1$, while those with large slack are smoothly downweighted but never fully removed.

\paragraph{\bf Using Confidence in Training.}
Confidence weights are used only to modulate supervision for weak labels.
Given a weak annotation map $y$, we define a reweighted target
\begin{align}
    \tilde{y}(t)
    =
    y \odot \bigl((1-\lambda(t)) + \lambda(t)\,c\bigr),
\end{align}
where $\lambda(t)\in[0,1]$ is a deterministic schedule increasing from $0$ to $1$ over training.
At early epochs, $\lambda(t)\approx 0$ and training relies on the raw annotations.
As training progresses, and the model becomes better calibrated, $\lambda(t)$ increases up to $1$ to progressively incorporate the slack-derived confidence into $\tilde{y}$.
Note that the slack is always computed with respect to the raw annotations $y$, not $\tilde{y}$,
ensuring that confidence estimation does not depend on the reweighting itself.
\fi

\subsection{Bootstrapping Marginal Residuals}
\label{sec:slack}

\begin{figure}[t]
    \centering
    \setlength{\fboxsep}{1pt} 

\begin{tabular}{c@{\;\;}c@{\;\;}c@{\;\;}c@{\;\;}l}
\begin{subfigure}{0.22\linewidth}
    \fbox{\includegraphics[width=1\linewidth, trim=2 15 2 15,
    clip]{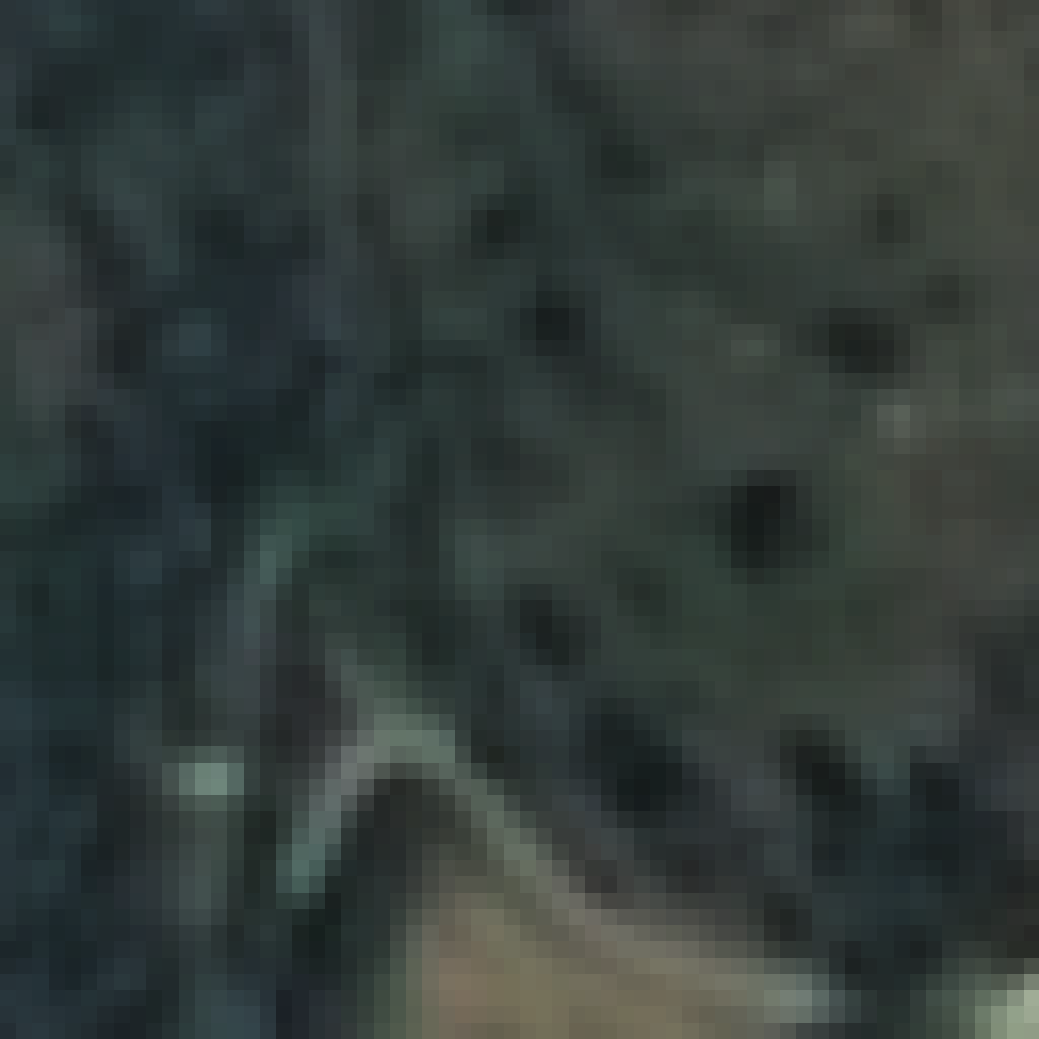}}
    \caption{Satellite image}
    \label{fig:residuals:a}
\end{subfigure}
& 
\begin{subfigure}{0.22\linewidth}
    \fbox{\includegraphics[width=1\linewidth, trim=2 15 2 15,
    clip]{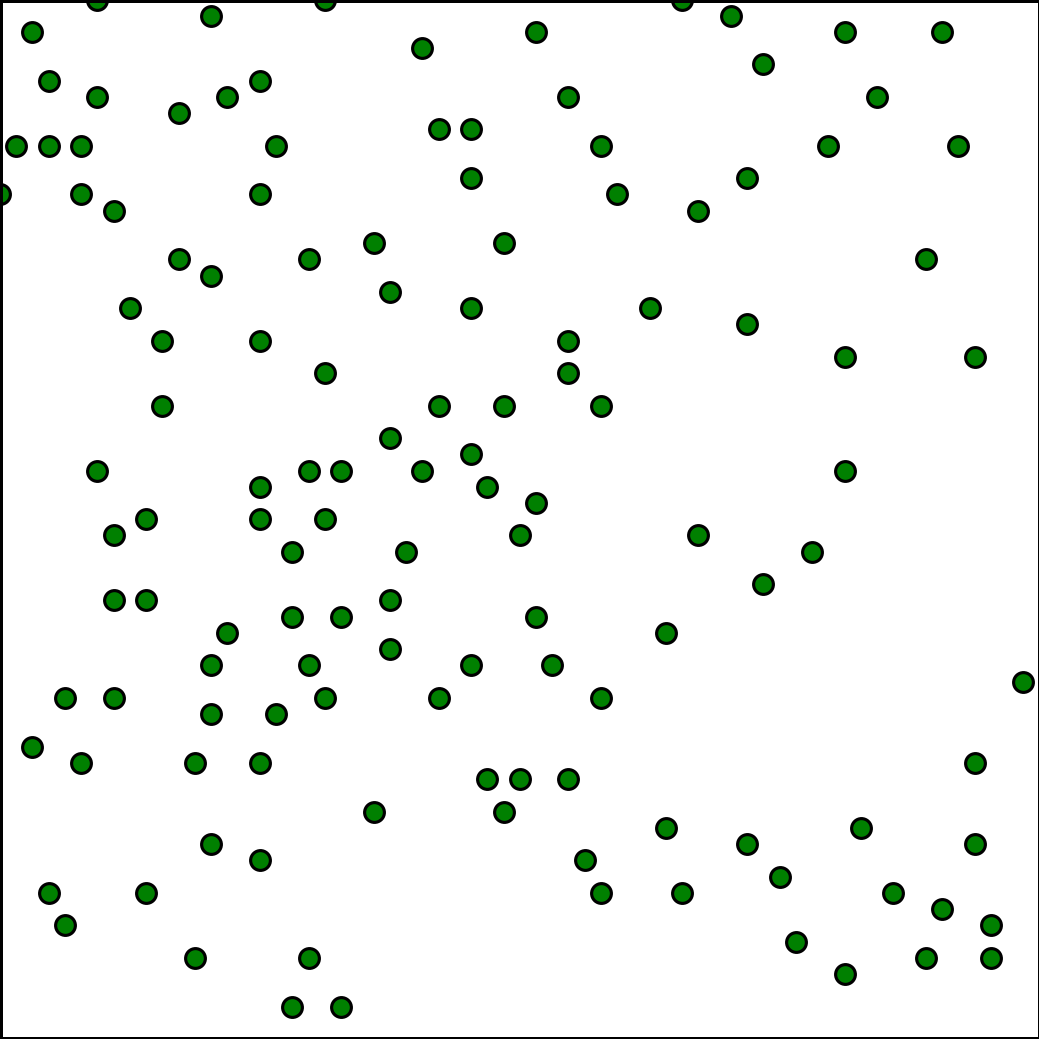}}
    \caption{Weak Labels}
     \label{fig:residuals:b}
\end{subfigure}
&
\begin{subfigure}{0.22\linewidth}
    \fbox{\includegraphics[width=1\linewidth, trim=2 15 2 15,
    clip]{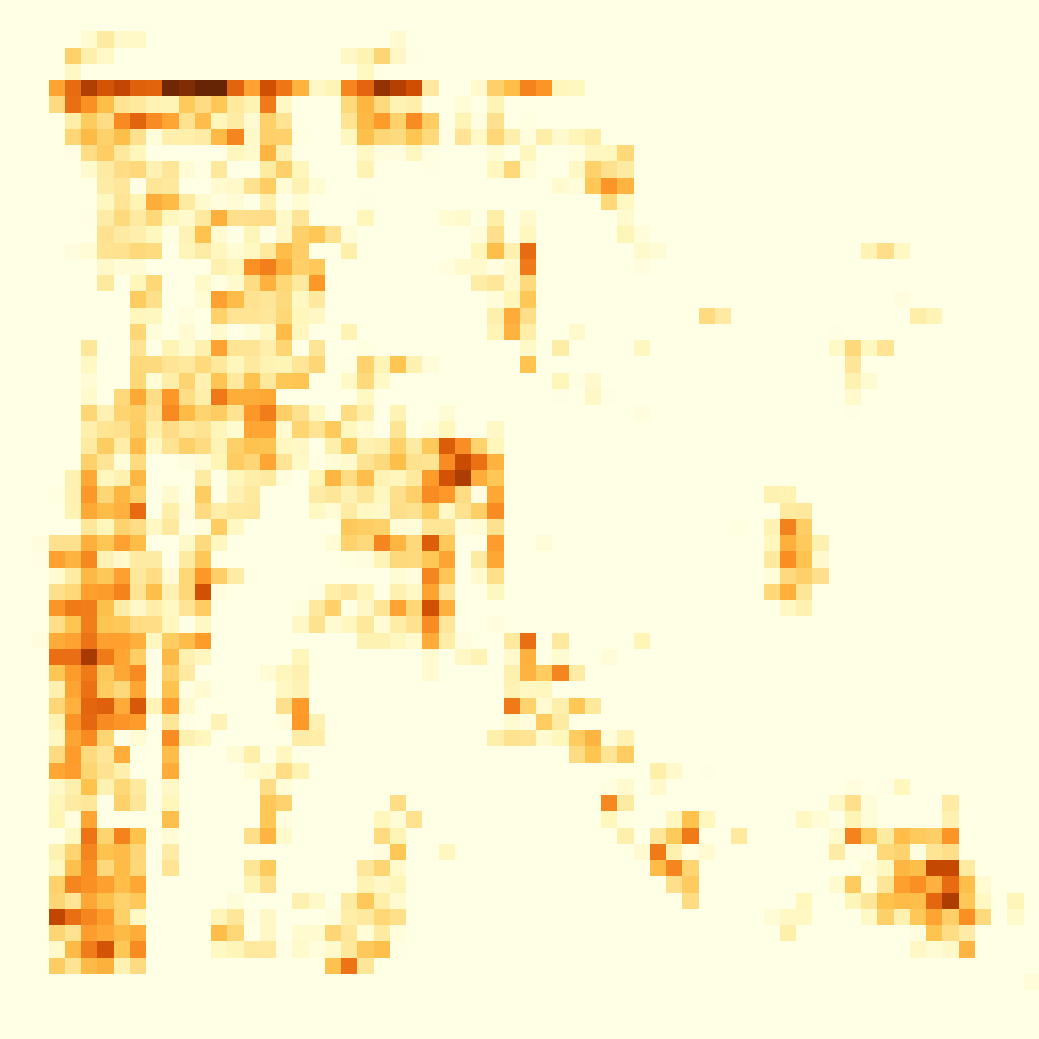}}
    \caption{Prediction}
     \label{fig:residuals:c}
\end{subfigure}
& 
\hspace{-3mm} 
\begin{subfigure}[t]{0.275\linewidth} 
\vspace{-32.2mm} 
{ \begin{tabular}{c@{}c} \fbox{\includegraphics[width=0.8\linewidth, trim=2 15 2 15,
    clip]{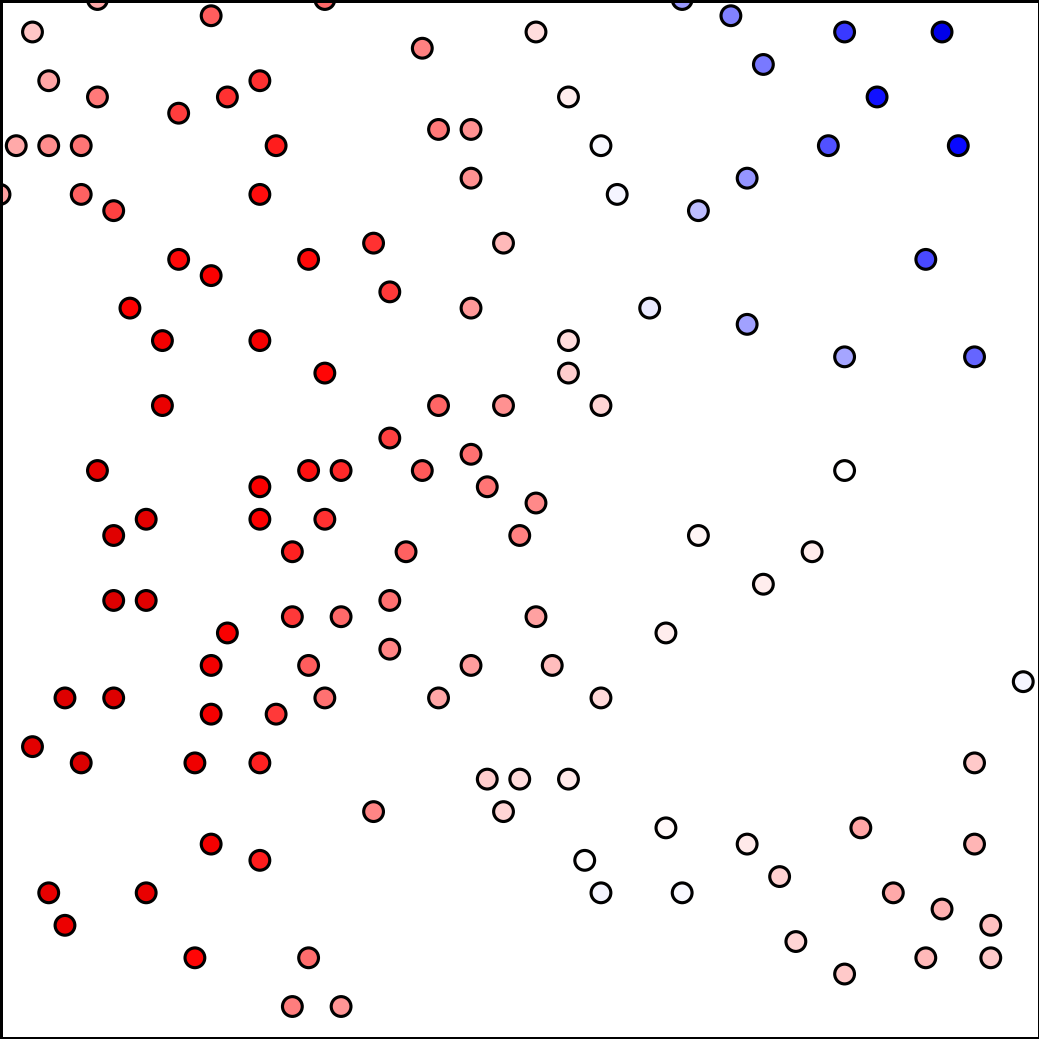}} &\hspace{-2cm} \def\colormapheight{1.7cm}
\pgfplotsset{%
    /pgfplots/colormap={seismic}{%
        rgb255(0cm)=(0,0,128), 
        rgb255(1cm)=(0,0,255),
        rgb255(2cm)=(255,255,255),
        rgb255(3cm)=(255,0,0),
        rgb255(4cm)=(128,0,0)
    }%
}%
\begin{tikzpicture}
    \begin{axis}[
        hide axis,
        scale only axis,
        height=5pt,
        width=50pt,
        xlabel={in m},
        colormap name=seismic,
        colorbar,
           colorbar,
            colorbar style={
            width=.2cm,
            height=\colormapheight,
            ytick={0,2,4},
            yticklabels={-1,0,1},
            yticklabel style={font=\tiny},
            major tick length=1.5pt, 
            line width=.05mm,
            grid style={draw=none} 
        },
        point meta min=0,
        point meta max=4
    ]
    \addplot [draw=none] coordinates {(0,0)};
    \end{axis}
\end{tikzpicture} \end{tabular} }
\vspace{-1mm}
\caption{Marginal Residuals} \label{fig:residuals:d} \end{subfigure} 
\end{tabular}


    \caption{{\bf Marginal Residuals.}
A satellite image (\subref{fig:residuals:a}) is annotated with weak labels (\subref{fig:residuals:b}) and processed by the model to produce a density prediction (\subref{fig:residuals:c}).
The resulting marginal residuals (\subref{fig:residuals:d}) highlight disagreements between annotation and prediction, revealing potential missed trees (in red) and spurious annotations (in blue).}
    \label{fig:residuals}
\end{figure}
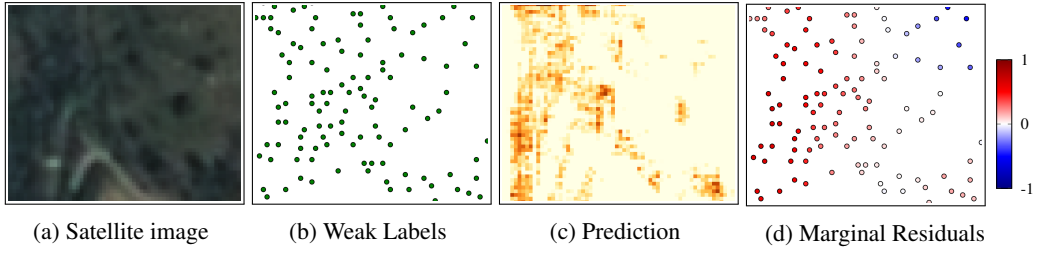

Tree annotations derived from automated pipelines are inherently noisy, often containing spurious detections or missing trees in dense forest regions.
These errors occur at the level of individual trees, making them difficult to address with a single global---or even image-level---relaxation parameter.
Rather than introducing ad-hoc filtering rules, we exploit the local mass discrepancy that naturally arises when solving the Unbalanced Optimal Transport (UOT) problem.

\paragraph{\textbf{Marginal Residuals from Transport.}}
When solving UOT between the predicted density ${z}$ and the annotated measure ${y}$, mass may be both discarded and created at a controlled cost.
Let $\gamma$ denote the optimal transport plan.
The column marginals $\gamma^\top \mathbf{1}$ represent the amount of mass effectively matched to each annotation under the transport geometry.
We define the \emph{marginal residual} as
\begin{align}\label{eq:residuals}
    r = \gamma^\top \mathbf{1} - {y}.
\end{align}
As illustrated in \cref{fig:residuals}, positive values indicate that the transport assigns more mass to a location than originally annotated, suggesting missed detections.
Negative values indicate that annotated mass is not supported by the prediction, suggesting over-annotation.

\paragraph{\textbf{Bootstrapped Target.}}
We use this residual to progressively refine the supervision signal:
\begin{align}
    \tilde{y}(t)
    =
    {y} + \lambda(t)\, r,
\end{align}
where $\lambda(t)\in[0,\alpha]$ increases from $0$ to $\alpha\in[0,1]$ during training.
Early in training, $\lambda(t)\approx 0$ and supervision relies on the raw annotations.
As the model becomes better calibrated, $\lambda(t)$ increases and the target progressively incorporates transport-consistent corrections.
Importantly, the transport plan $\gamma$ is always computed with respect to the original annotations $y$, ensuring that the correction does not depend on previously reweighted targets.

\subsection{Learning with Weak and Strong Tree Labels}
\label{sec:weak}

Manually annotating individual trees at scale is prohibitively expensive.
Instead, approximate tree-center or tree-cover annotations can be derived automatically from airborne laser scanning (ALS) data (see \cref{subsec:dataset}).
While these weak annotations are informative, they become noisy in dense forests and may contain missed or spurious detections.
In contrast, manual annotations are accurate but sparse.
We therefore consider two training sets:
a strong set $\mathcal{D}^{\text{strong}}$ with reliable spatial localization and counts,
and a larger weak set $\mathcal{D}^{\text{weak}}$ with potentially inaccurate annotations.

Given a prediction $\phi(x)$, strong labels supervise both the global count and spatial alignment using the counting loss $\mathcal{L}^{\text{count}}$ and the UOT-based distribution matching loss $\mathcal{L}^{\text{UOT}}$.
For weak labels, we first compute marginal residuals (\cref{sec:slack}) and use them to form corrected annotations $\tilde{y}$.
This correction is computed without backpropagation.
We then apply the UOT loss between the prediction and $\tilde{y}$, with backpropagation.
The overall objective is
\begin{align}
\mathcal{L}
=&
\sum_{(x,y) \in \mathcal{D}^{\text{strong}}}
\Big(
\mathcal{L}^{\text{count}}\!\left( \phi(x), y \right)
+
\mathcal{L}^{\text{UOT}}\!\left( \phi(x), y \right)
\Big)
\\
&+
\sum_{(x,y) \in \mathcal{D}^{\text{weak}}}
\mathcal{L}^{\text{UOT}}\!\left( \phi(x), \SG(\tilde{y}) \right),
\end{align}
where $\SG$ denotes a stop-gradient operator preventing gradients from flowing through the correction step.
This formulation enables large-scale learning from noisy supervision while preserving reliable count and spatial constraints where clean annotations are available. Although weak labels are typically far more abundant than strong ones, we sample $\mathcal{D}^{\text{weak}}$ and $\mathcal{D}^{\text{strong}}$ to have comparable sizes.
This ensures accurate count supervision from strong annotations while stabilizing training and preventing circular feedback from the correction mechanism applied to noisy labels.

\paragraph{\textbf{Implementation Details.}}
The transport plan smoothness $\varepsilon$ is set to $0.005$, inducing an effective interaction radius of approximately $\sqrt{\varepsilon} \approx 0.07$ in normalized coordinates.
The mass deviation penalty $\tau$ is set between 0.1 and 0.5 depending on the dataset, balancing the cost of creating unit mass against the cost of transporting unit mass across one unit spatial distance (in normalized coordinates). Smaller values of $\tau$ reduce mass flexibility, while larger values allow moderate shifts in density.
The scheduler $\lambda(t)$ increases from 0 to $\alpha=0.8$ over the first 400 optimization steps using a tempered sigmoid schedule.
For all data splits, we use AdamW \cite{loshchilov_decoupled_2019} with default parameters. We train for 500 epochs with a cosine annealing learning rate scheduler with an initial value of 8e-5. Weak and strong training samples are drawn in  stratified 1:1 ratio.

\if 1 0
Unbalanced Optimal Transport (UOT), introduced by Chizat et al.~\cite{chizat_scaling_2018},
relaxes the marginal constraints by allowing mass creation or destruction at a controlled cost.
Given two nonnegative measures $\mu$ and $\nu$ with possibly different total mass,
UOT solves
\begin{equation}
    W_{\varepsilon,\tau}(\mu,\nu)
    \;=\;
    \min_{\gamma \ge 0}
    \Bigl(
        \langle C , \gamma \rangle
        + \varepsilon\, \mathrm{H}(\gamma)
        + \tau\, D_\varphi(\gamma \mathbf{1}_m \,\|\, \mu)
        + \tau\, D_\varphi(\gamma^\top \mathbf{1}_n \,\|\, \nu)
    \Bigr),
\end{equation}
where $\varepsilon > 0$ is the entropic regularization strength,
$\tau > 0$ controls the penalty for modifying total mass,
and $D_\varphi$ is the KL divergence.
When $\tau \rightarrow \infty$, the classical balanced OT constraints are recovered.

In density regression,
$\mu$ corresponds to the predicted density map and
$\nu$ corresponds to the empirical distribution of annotated points.

\textbf{UOT provides a key advantage: the parameter $\tau$ controls the trade-off between enforcing strict Optimal Transport constraints (assuming equal total mass) and allowing soft, noise-robust assignments that tolerate differences in total counts.}

Optimal Transport (OT) \cite{villani_optimal_2009} refers to the optimal cost of transforming one 
probability distribution into another. We review here the 
Monge--Kantorovich formulation.
Let $X = \{x_i \mid x_i \in \mathbb{R}^d\}_{i=1}^n$ and 
$Y = \{y_j \mid y_j \in \mathbb{R}^d\}_{j=1}^n$ 
be two sets of points in a $d$-dimensional vector space. 
Let $\mu$ and $\nu$ be two probability measures defined on 
$X$ and $Y$, respectively; 
$\mu,\nu \in \mathbb{R}_+^n$ and 
$\mathbf{1}_n^\top \mu = \mathbf{1}_n^\top \nu = 1$ 
(where $\mathbf{1}_n$ is the $n$-dimensional vector of all ones).

Let $c : X \times Y \to \mathbb{R}_+$ be the cost function for moving a unit mass 
from a point in $X$ to a point in $Y$, and let $C$ be the corresponding 
$n \times n$ cost matrix with entries
\[
C_{ij} = c(x_i, y_j).
\]

Let $\Gamma$ be the set of all transport plans:
\[
\Gamma = \{\gamma \in \mathbb{R}_+^{n \times n} 
\; : \; \gamma \mathbf{1}_n = \mu,\; \gamma^\top \mathbf{1}_n = \nu \}.
\]

The Monge--Kantorovich Optimal Transport cost between $\mu$ and $\nu$ is defined as
\begin{equation}
W(\mu, \nu) 
= \min_{\gamma \in \Gamma} \langle C, \gamma \rangle ,
\label{eq:ot-primal}
\end{equation}
where $\langle C, \gamma \rangle = \sum_{i,j} C_{ij} \gamma_{ij}$.

Intuitively, if the probability distribution $\mu$ is viewed as a unit amount 
of “dirt” piled on $X$ and $\nu$ as a unit amount of dirt piled on $Y$, the OT cost 
is the minimum effort required to transform one pile into the other. 
The OT cost quantifies the dissimilarity between two probability distributions 
while taking into account the geometry of the underlying space.

Although theoretically appealing, this problem is computationally expensive due to the
high-dimensional linear program. Entropy-regularized OT introduces a negative
entropy penalty, proposed by Cuturi~\cite{cuturi_sinkhorn_2013}, yielding the problem
\begin{equation}
    W_\varepsilon(\mu,\nu)
    \;=\;
    \min_{\gamma \in \Gamma(\mu,\nu)}
    \Bigl(
        \langle C, \gamma \rangle
        + \varepsilon \, \mathrm{H}(\gamma)
    \Bigr),
    \qquad
    \mathrm{H}(\gamma)
    =
    \sum_{i,j} \gamma_{ij} (\log \gamma_{ij} - 1).
\end{equation}
The regularization parameter $\varepsilon>0$ controls the balance between fidelity
to the original OT problem and smoothness of the coupling.

The optimal transport plan admits the well-known scaling form
$\gamma^\star = \mathrm{diag}(u)\, K \, \mathrm{diag}(v)$,
where $K = \exp(-C / \varepsilon)$ is the Gibbs kernel.
The scaling vectors $u$ and $v$ are obtained via the Sinkhorn--Knopp
iterative updates:
\begin{equation}
    u^{(t+1)} = \frac{\mu}{K v^{(t)}},
    \qquad
    v^{(t+1)} = \frac{\nu}{K^\top u^{(t+1)}}.
\end{equation}
This procedure converges rapidly, is GPU-friendly,
and yields a differentiable approximation of OT.
Entropy-regularized OT therefore forms the foundation of most modern OT loss
functions in deep learning.

\subsection{Unbalanced Optimal Transport for Density Regression}

In density regression tasks, the predicted density map may not integrate to the same total mass as the empirical distribution defined by annotated points.
Classical OT requires the input distributions to have equal total mass,
which imposes an unrealistic global normalization constraint
and may lead to unstable optimization when the predicted count is inaccurate.

Unbalanced Optimal Transport (UOT), introduced by Chizat et al.~\cite{chizat_scaling_2018},
relaxes the marginal constraints by allowing mass creation or destruction at a controlled cost.
Given two nonnegative measures $\mu$ and $\nu$ with possibly different total mass,
UOT solves
\begin{equation}
    W_{\varepsilon,\tau}(\mu,\nu)
    \;=\;
    \min_{\gamma \ge 0}
    \Bigl(
        \langle C , \gamma \rangle
        + \varepsilon\, \mathrm{H}(\gamma)
        + \tau\, D_\varphi(\gamma \mathbf{1}_m \,\|\, \mu)
        + \tau\, D_\varphi(\gamma^\top \mathbf{1}_n \,\|\, \nu)
    \Bigr),
\end{equation}
where $\varepsilon > 0$ is the entropic regularization strength,
$\tau > 0$ controls the penalty for modifying total mass,
and $D_\varphi$ is typically a KL divergence.
When $\tau \rightarrow \infty$, the classical balanced OT constraints are recovered.

In density regression,
$\mu$ corresponds to the predicted density map and
$\nu$ corresponds to the empirical distribution of annotated points.

\textbf{UOT provides a key advantage: the parameter $\tau$ controls the trade-off between enforcing strict Optimal Transport constraints (assuming equal total mass) and allowing soft, noise-robust assignments that tolerate differences in total counts.}

Practically, unbalanced Sinkhorn iterations are used to compute the transport plan:
\begin{equation}
    u^{(t+1)} = 
        \left(
            \frac{\mu}{K v^{(t)}}
        \right)^{\frac{\tau}{\tau + \varepsilon}},
    \qquad
    v^{(t+1)} = 
        \left(
            \frac{\nu}{K^\top u^{(t+1)}}
        \right)^{\frac{\tau}{\tau + \varepsilon}}.
\end{equation}
These generalized scaling updates interpolate between normalization and full flexibility,
making UOT a natural and powerful choice for density regression tasks where precise mass
conservation is inappropriate.

Following DM-Count, our overall loss function is a combination of the counting loss, the uOT loss, and the TV loss:
\begin{equation}
\ell(z, \hat{z}) = 
\ell_C(z, \hat{z}) 
+ \lambda_1 \, \ell_{\mathrm{OT}}(z, \hat{z}) 
+ \lambda_2 \, \|z\|_1 \, \ell_{\mathrm{TV}}(z, \hat{z}),
\label{eq:total-loss}
\end{equation}
where $\lambda_1$ and $\lambda_2$ are tunable hyper-parameters weighting the OT and TV loss terms. 
To ensure that the TV loss has the same scale as the counting loss, this term is multiplied by the total count.
\fi

\section{Experiments}
\label{sec:experiments}

\begin{table*}[t]
\centering
\caption{{\bf Overview of the \textsc{TinyTrees} datasets.}
Dataset statistics by region and split.
Weak annotations are obtained from height/crown predictions derived from aerial LiDAR scans, while strong labels are manual annotations obtained via photointerpretation (\faEye) or field surveys (\faWalking).
\hfill $\star$: use medium density LiDAR, likely undercounted.}

\label{table:datasets}
\begin{tabular}{c@{\quad}cccc@{\quad}c}
\toprule
 Location & Sensor /GSD  & Split & area (km²) & \#trees &label source \\ \midrule
\multirow{3}{*}{China} & \multirow{3}{*}{\makecell{Gaofen-2 \\ 0.8m}}  & Train-weak & 50.3 & 7.68M$^\star$ & height reg.\\
& & Train-strong & 5.9 & 54.6k & manual \scriptsize{\faEye} \\
 & & Test & 7.02 & 58.4k & manual \scriptsize{\faEye}\\\midrule
\multirow{3}{*}{\makecell{Rwanda}} &\multirow{3}{*}{\makecell{PlanetScope \\ 3.4-4.2 m}} & Train-weak & 209.3 & 3.40M & crown seg.\\
& & Train-strong & 48.4 & 309k & manual \scriptsize{\faEye}\\
& & Test & 33.9 & 195k & manual \scriptsize{\faEye}\\
\midrule
 \multirow{3}{*}{\makecell{France}} & \multirow{3}{*}{\makecell{Spot6 \\ 1.5m}} & Train-weak & 23.3k & 204M & high-res ALS\\
& & Train-strong & 0.35 & 11.1k & manual \faWalking\\
& & Test & 0.35 & 10.8k & manual \faWalking\\
 \bottomrule
  \end{tabular}
\end{table*}

We first introduce the \textsc{TinyTrees} datasets (\cref{subsec:dataset}), and present our baselines and metrics (\cref{subsec:baselines}). We then detail quantitative results and analysis (\cref{subsec:results}), followed by an ablation study (\cref{subsec:ablation}).

\subsection{The \textsc{TinyTrees} Dataset}
\label{subsec:dataset}

We introduce \textsc{TinyTrees}, the first large-scale dataset for tree counting from satellite imagery.
The dataset spans three continents and three satellite sensors with ground sampling distances (GSD) ranging from 0.8\,m to 4.2\,m.
In total, it covers over 25k\,km$^2$ and contains more than 216 million tree annotations.
An overview is provided in \cref{table:datasets}.

\paragraph{\textbf{Areas of Interest.}}
\textsc{TinyTrees} covers three geographically and ecologically distinct regions.
\begin{itemize}

\item \textbf{China.}
Very-high-resolution Gaofen-2 imagery (0.8\,m GSD) covering temperate forests.
Manual annotations were produced through expert photo-interpretation, supported by higher-resolution reference imagery.
Canopy height maps from supervised regression trained with medium-density ALS are available ($\sim 2 pt/m\textsuperscript{2}$).

\item \textbf{Rwanda.}
PlanetScope imagery (3--4\,m GSD) covering heterogeneous tropical landscapes.
Manual annotations were created through expert photo-interpretation.
Large-scale tree crown predictions derived from aerial imagery are available.

\item \textbf{France.}
SPOT-6 imagery (1.5\,m GSD, pansharpened from 6\,m) covering temperate forests.
We have access to high-density LiDAR-HD ALS data ($\geq 20 pt/m\textsuperscript{2}$) to derive canopy height maps~\cite{ign_lidar_nodate}, and forest inventory data on 15 m radius plots from the French National Forest Office are available.
\end{itemize}

\paragraph{\textbf{Annotation Types.}}
Each region contains two forms of supervision: strong and weak.
\emph{Strong annotations} are obtained through manual photo-interpretation or \emph{in situ} field measurements, providing high precision but limited geographic coverage.

\emph{Weak annotations} are automatically derived at large scale but are inherently noisy, particularly in dense canopy where tree separation is imperfect. They are obtained from dedicated canopy height models or large-scale crown prediction pipelines.
For France, weak labels are generated from ALS-derived canopy height maps using a fully automated peak detection procedure~\cite{gominski_trees_2026}.
In Rwanda, weak labels originate from a semi-automatic national-scale crown segmentation product~\cite{mugabowindekwe_nation-wide_2023}.
For the China split, we derive pseudo-density targets from canopy height maps: canopy pixels above a 3\,m threshold are assigned the average tree density estimated from the strongly annotated subset.

\paragraph{\textbf{Dataset Splits.}}
All images are extracted as $64 \times 64$ RGB+NIR patches at nominal satellite resolution.
Each region is partitioned into three subsets:
\begin{itemize}
    \item \textbf{Train-weak:} large geographic areas with automatic annotations,
    \item \textbf{Train-strong:} smaller areas with manual annotations,
    \item \textbf{Test:} held-out areas with manual annotations only.
\end{itemize}
To prevent spatial leakage, a 1\,km buffer separates training and test regions.
Because manual annotations cover non-square geometries, a small fraction ($<10\%$) of annotated trees may appear in overlapping patches.

\begin{figure}[t]
    \centering
     \captionsetup[subfigure]{%
   justification=centering,
   labelfont=small,
   textfont=small,
}
\def\wimg{0.14}
\scriptsize
\begin{tabular}{l@{\;}c@{\;}c@{\;}c@{\;}c@{\;}c@{\;}c}
\scriptsize \rotatebox{90}{\quad\;\;\shortstack{China \\ Gaofen2}}
&
\includegraphics[width=\wimg \linewidth, height= \wimg \linewidth]{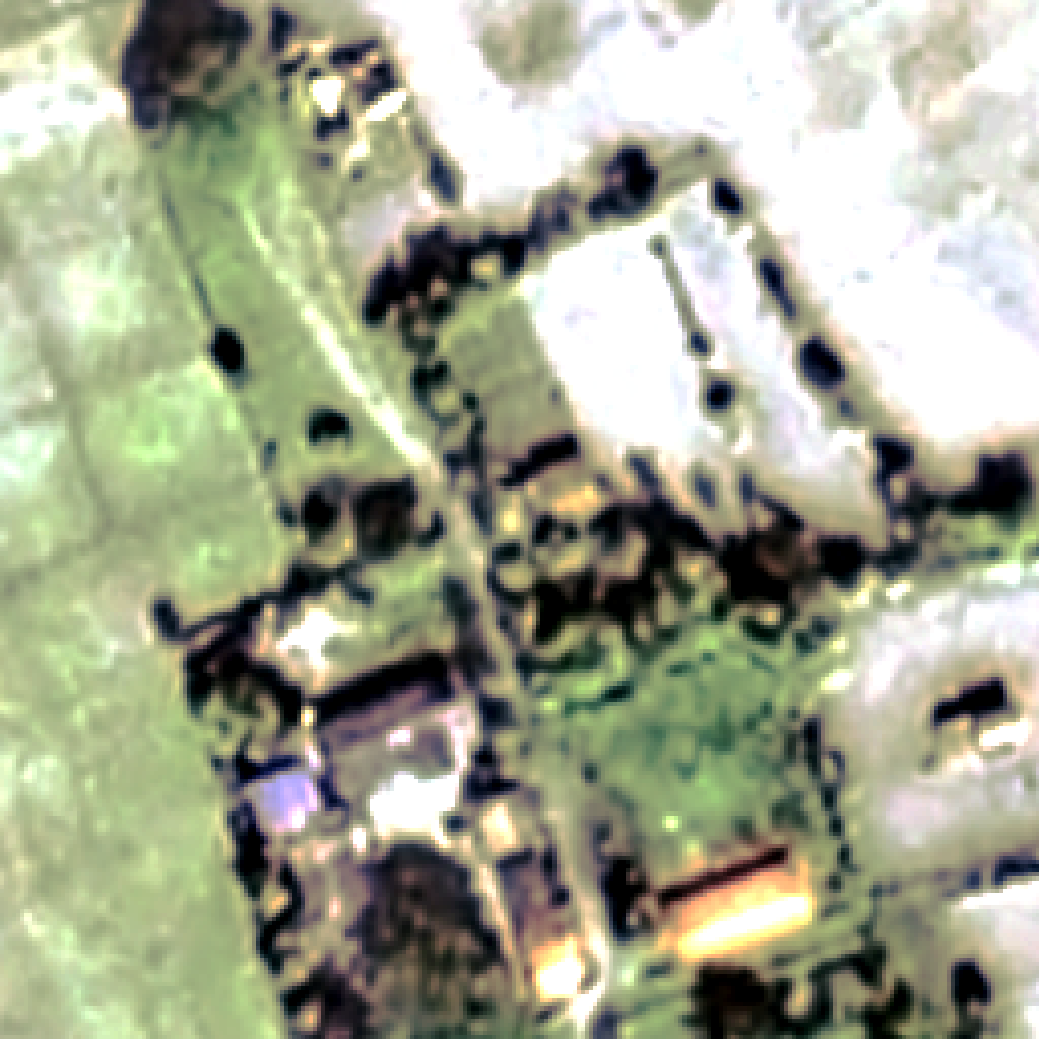}   
&
\includegraphics[width=\wimg \linewidth, height= \wimg \linewidth]{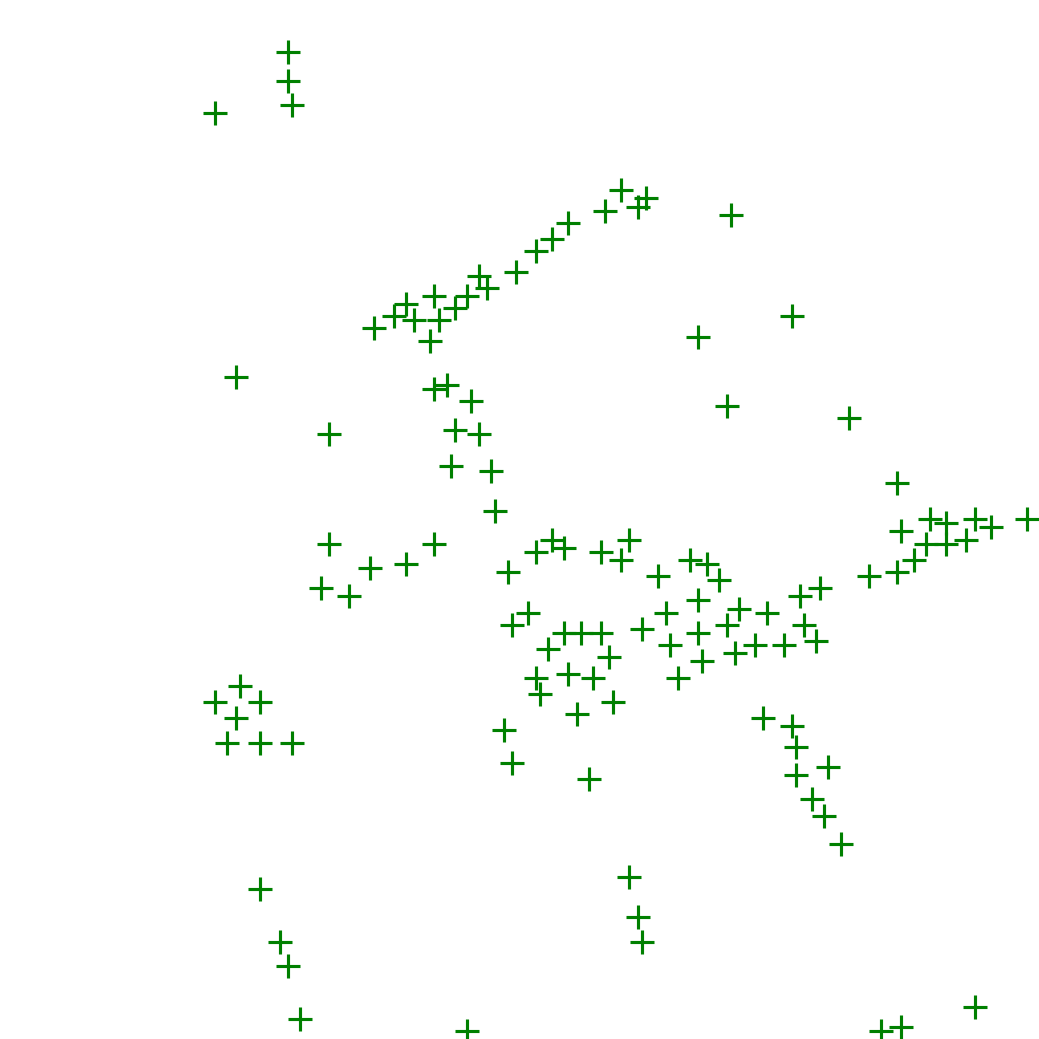} 
&
\includegraphics[width=\wimg \linewidth, height= \wimg \linewidth]{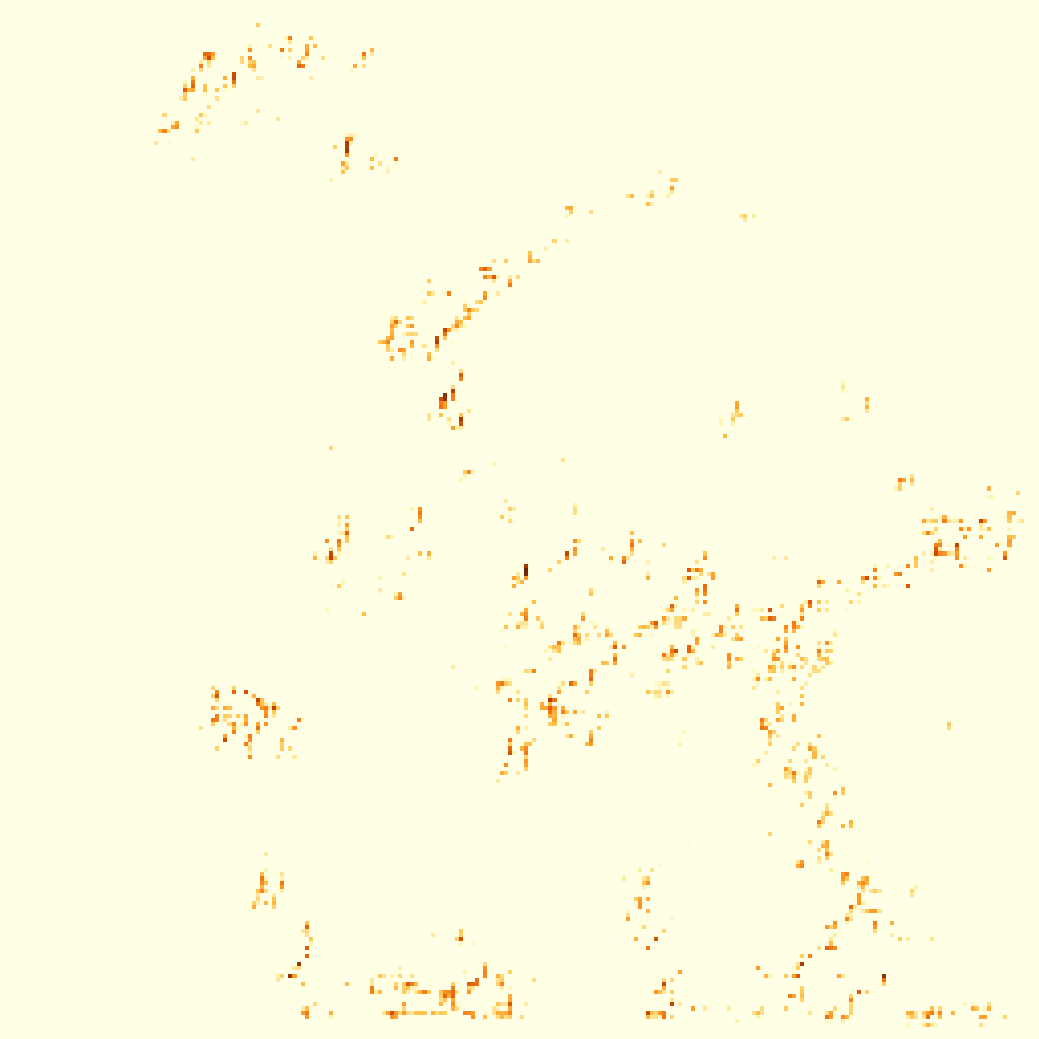} 
&
\includegraphics[width=\wimg \linewidth, height= \wimg \linewidth]{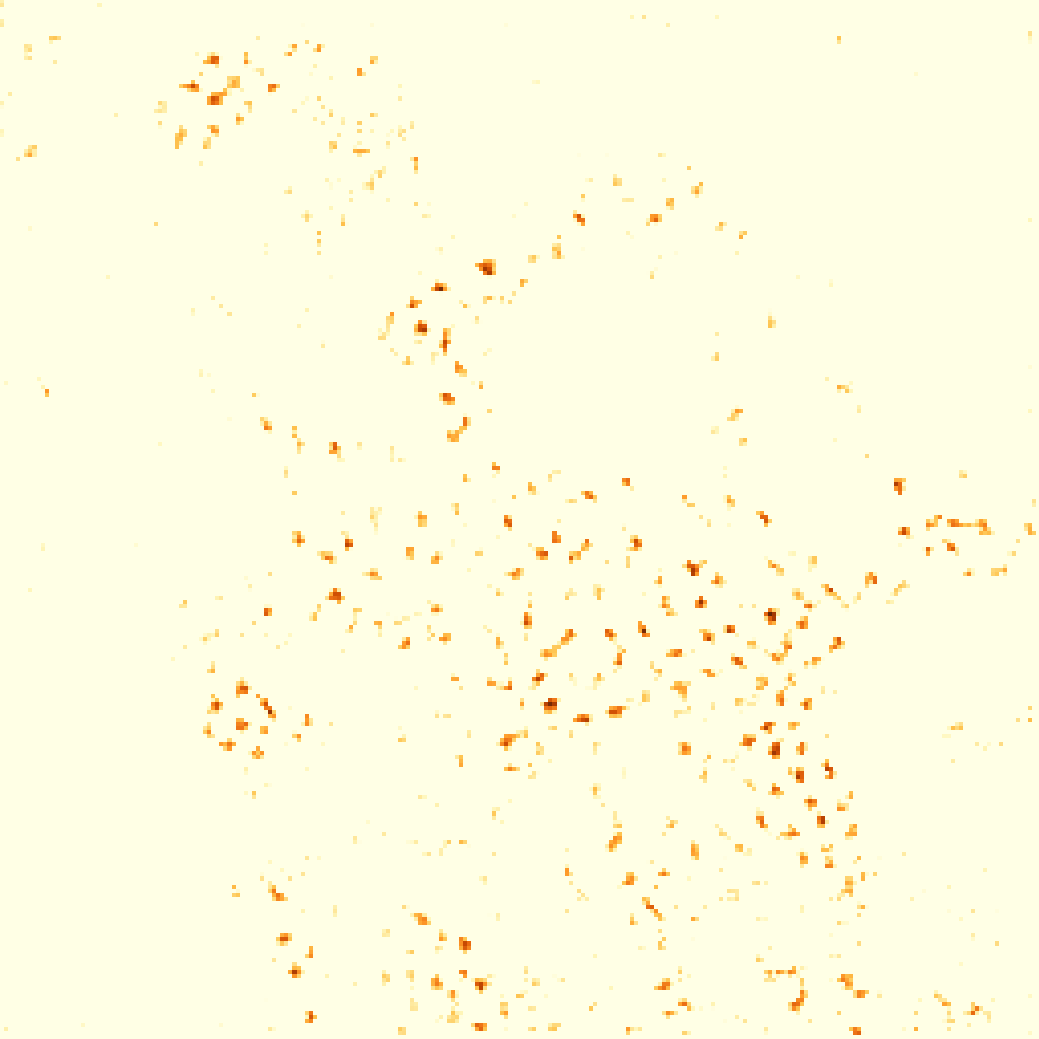} 
&
\includegraphics[width=\wimg \linewidth, height= \wimg \linewidth]{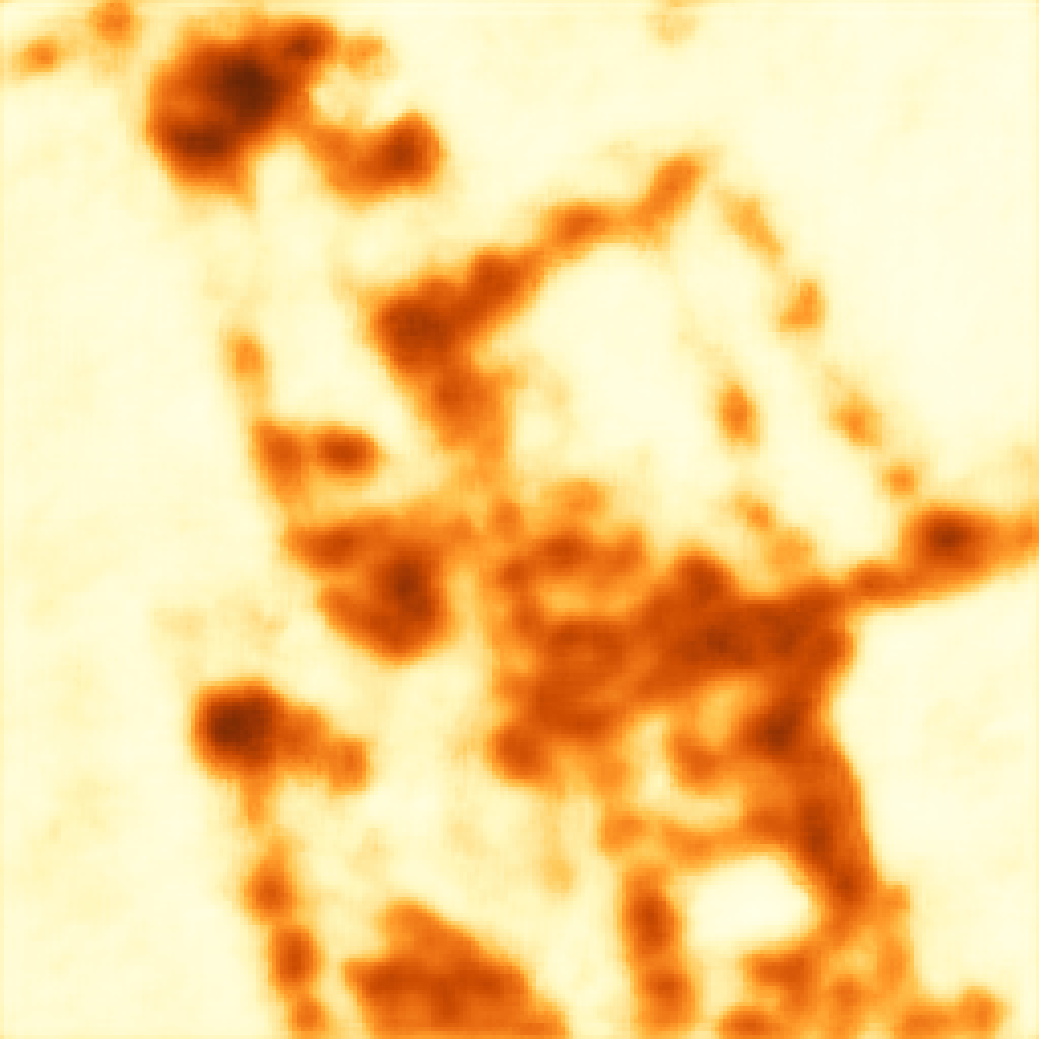} 
&
\includegraphics[width=\wimg \linewidth, height= \wimg \linewidth]{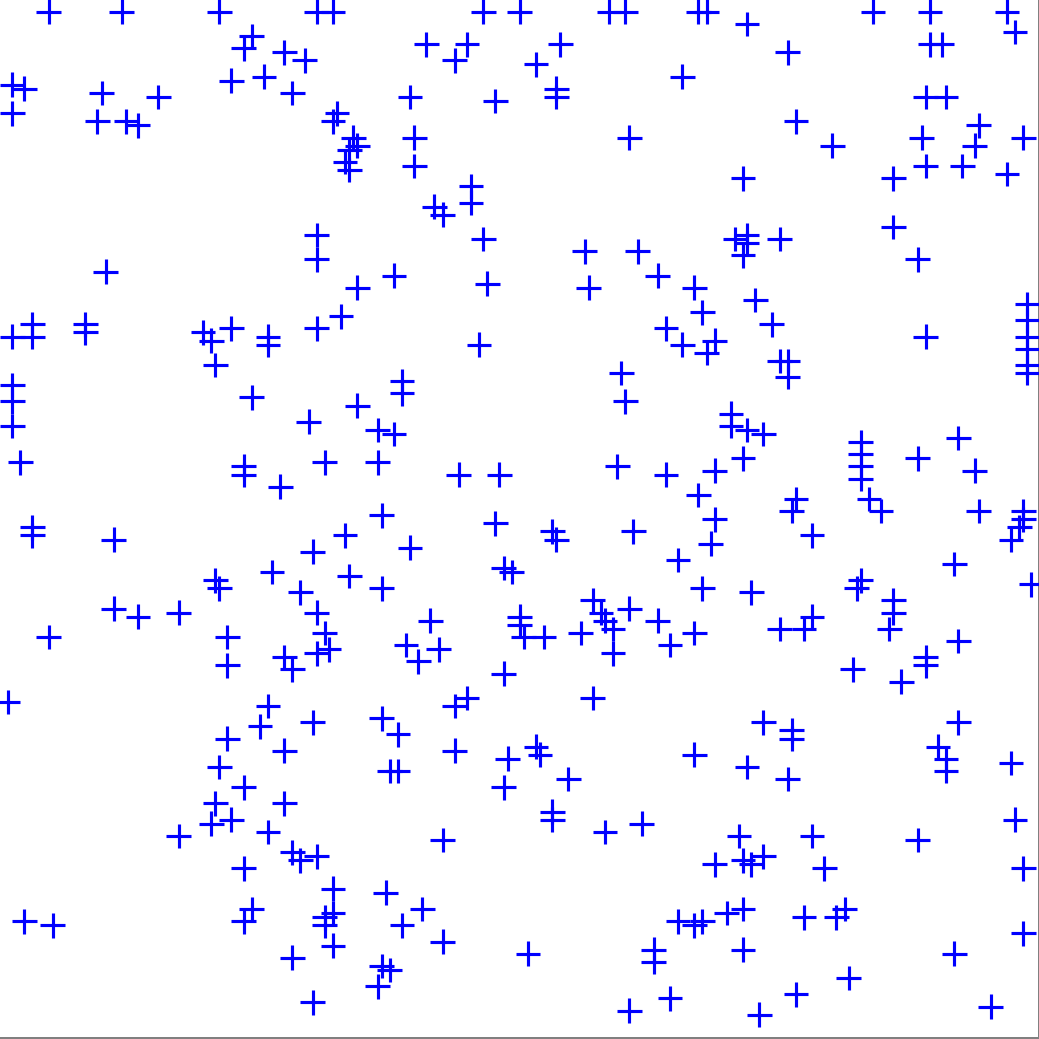} 
\\
& & 127 trees & 143.4 trees & 267.0 trees & 233.3 trees & 332 trees
\\
\scriptsize \rotatebox{90}{\;\;\shortstack{Rwanda \\ PlanetScope}}
&
\includegraphics[width=\wimg \linewidth, height= \wimg \linewidth]{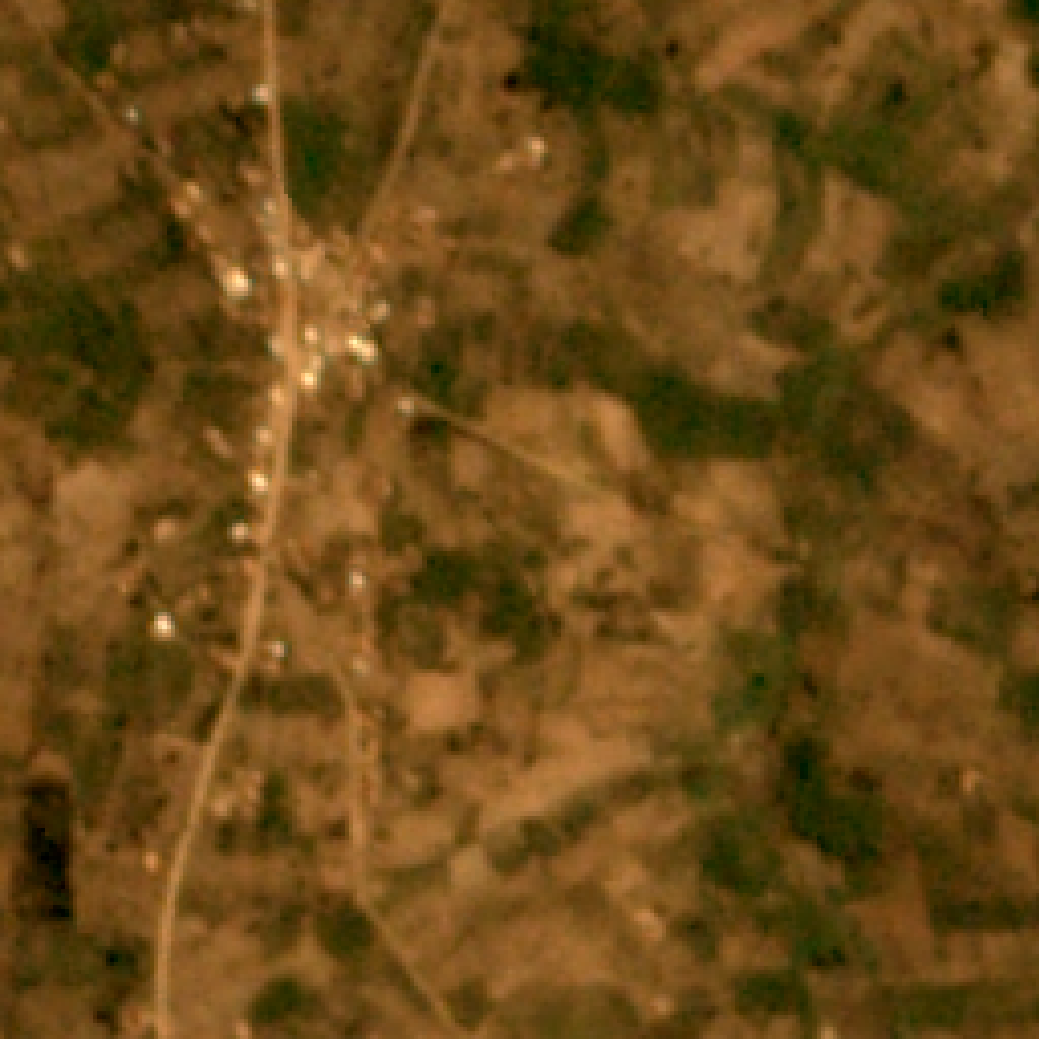}  
&
\includegraphics[width=\wimg \linewidth, height= \wimg \linewidth]{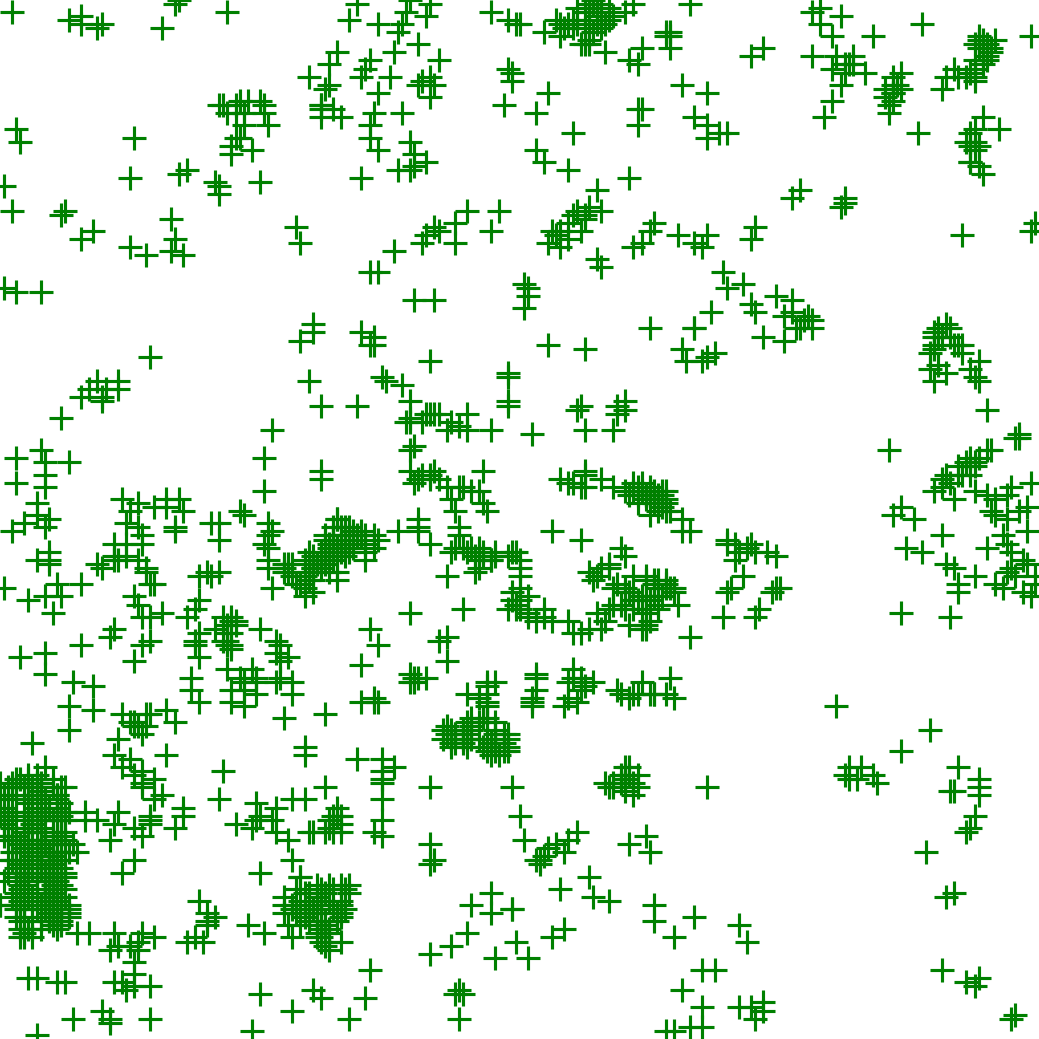} 
&
\includegraphics[width=\wimg \linewidth, height= \wimg \linewidth]{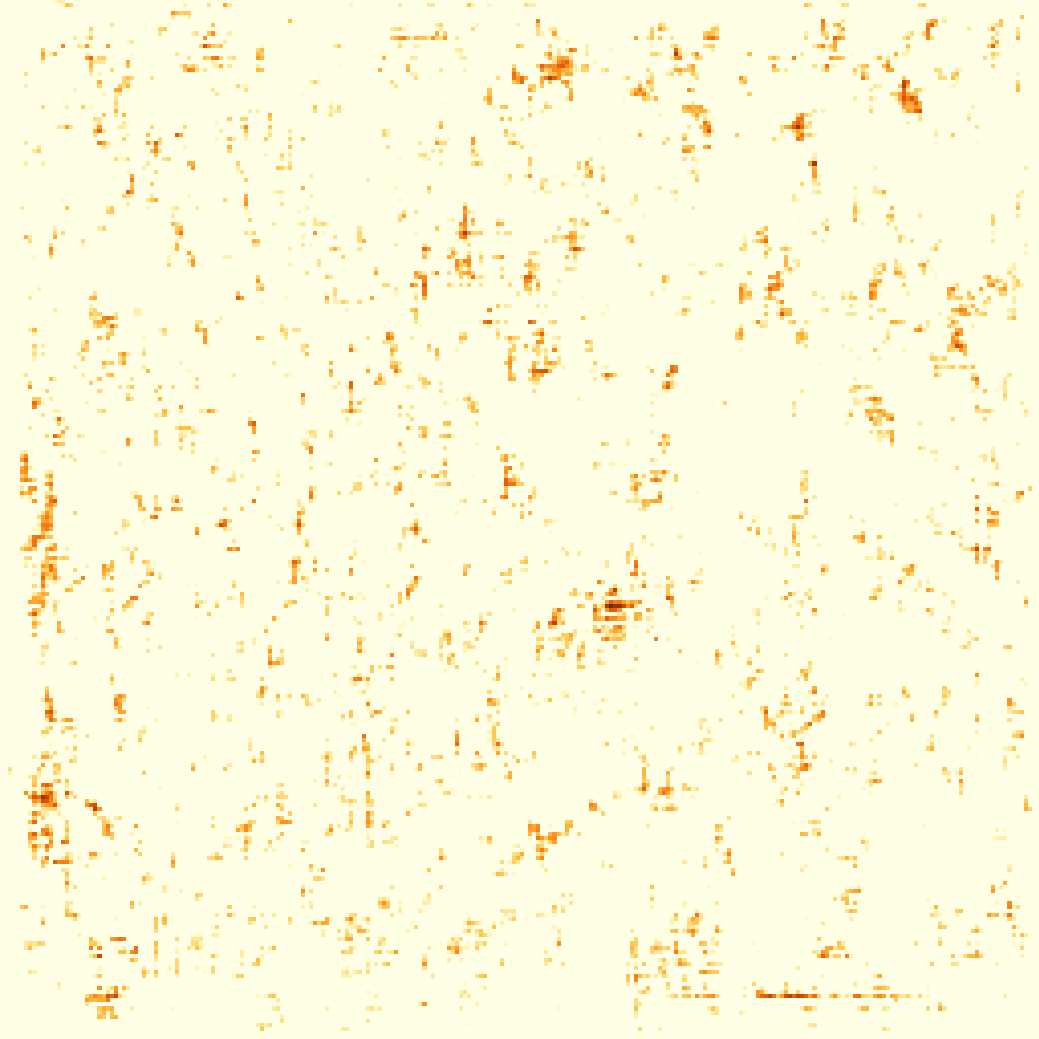} 
&
\includegraphics[width=\wimg \linewidth, height= \wimg \linewidth]{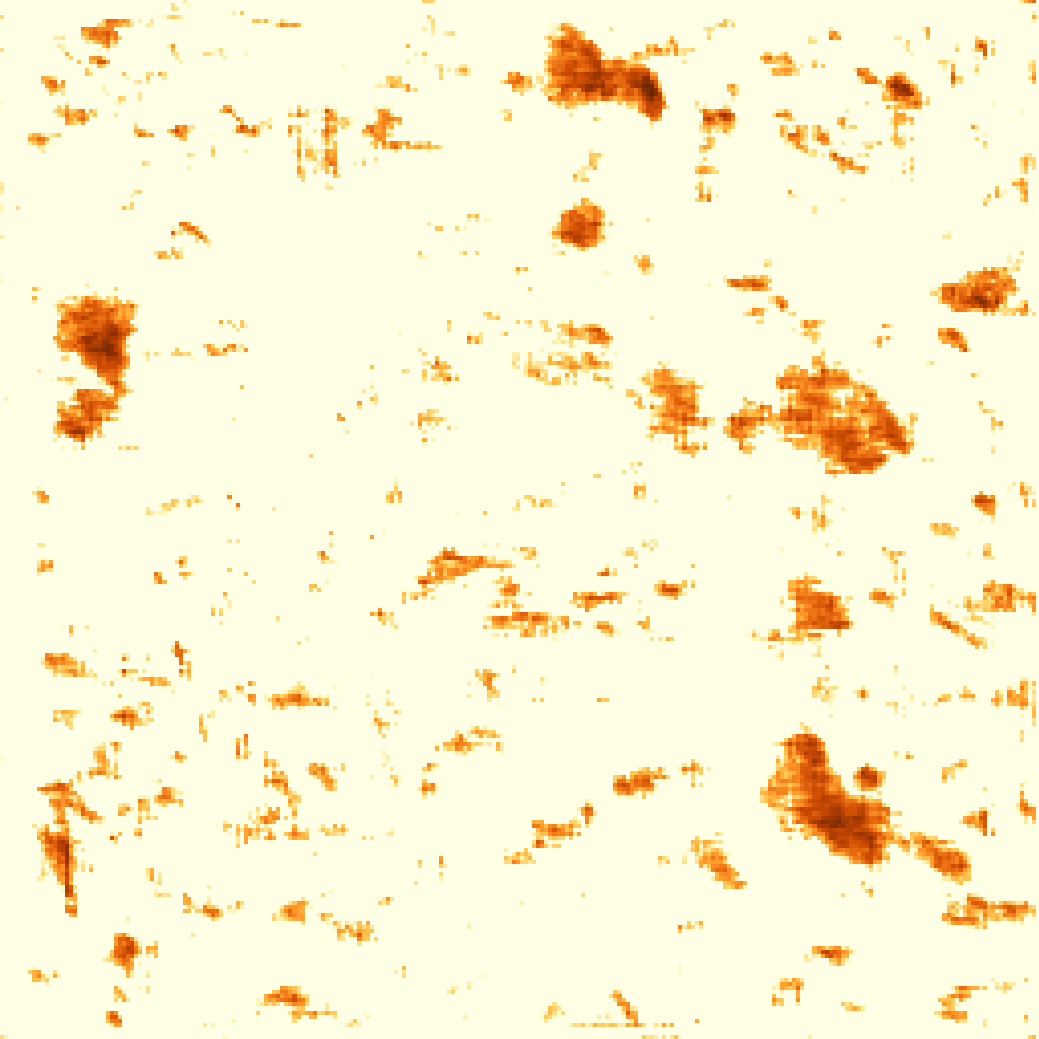} 
&
\includegraphics[width=\wimg \linewidth, height= \wimg \linewidth]{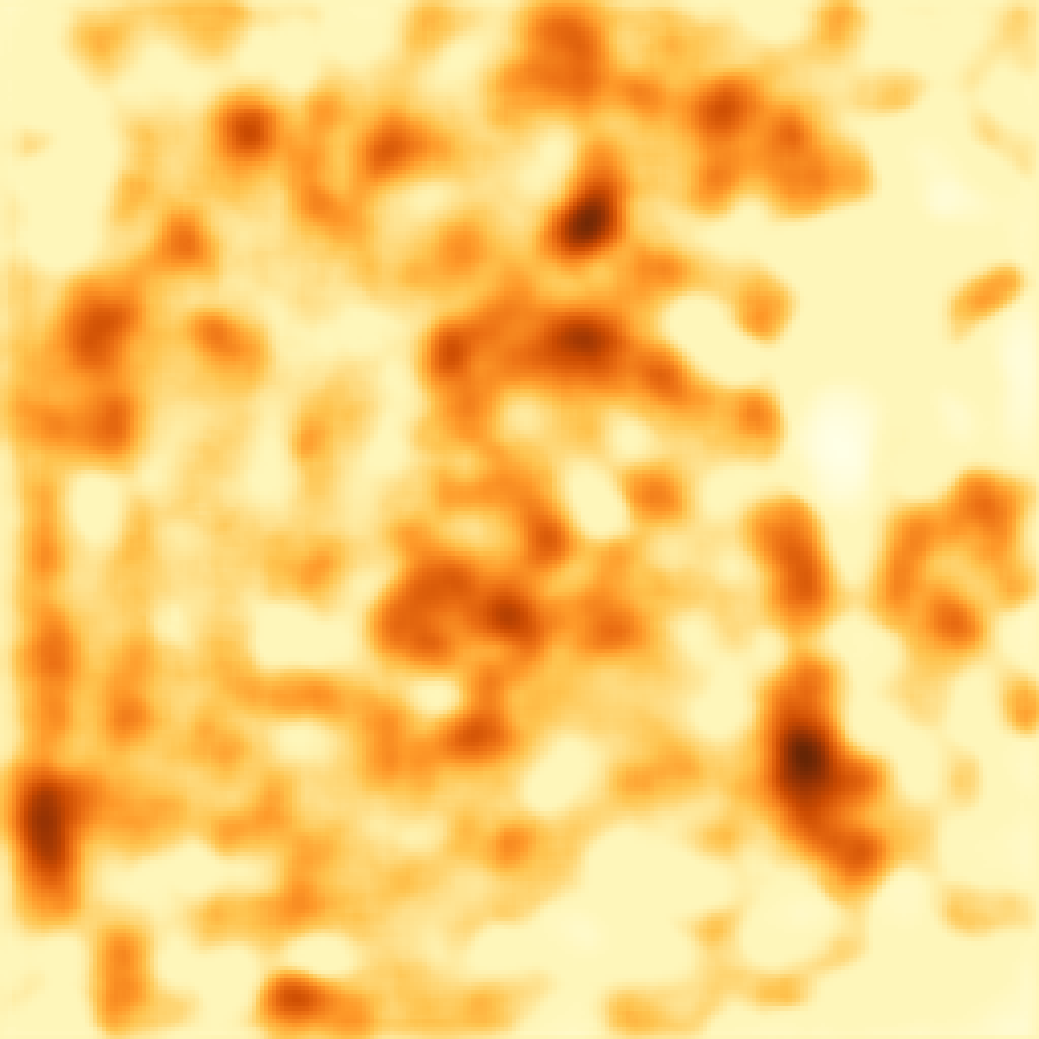}
&
\includegraphics[width=\wimg \linewidth, height= \wimg \linewidth]{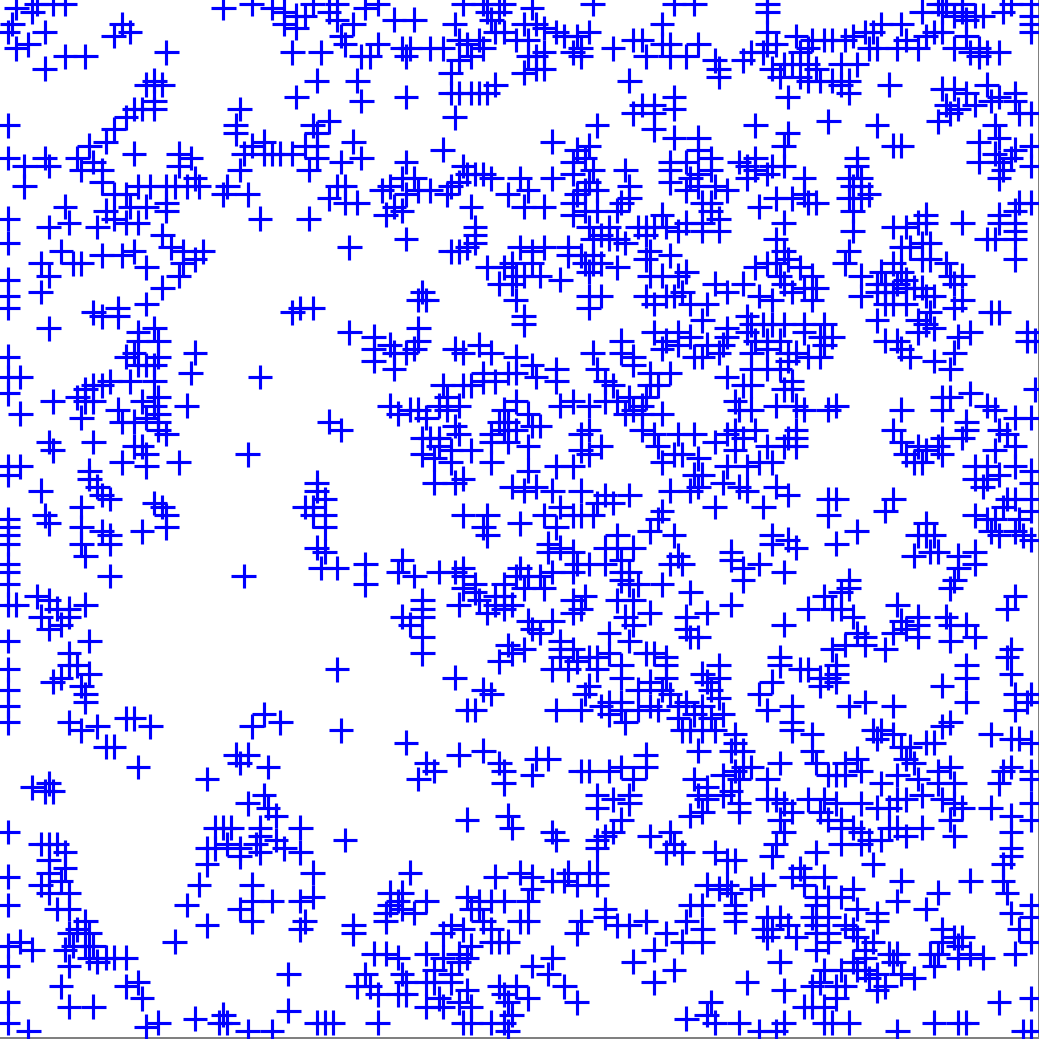}
\\
& & 1471 trees & 1326.7 trees & 2182.0 trees & 2567.1 trees & 1672 trees\\
\scriptsize \rotatebox{90}{\quad\;\;\shortstack{France \\ SPOT6}}
&
\includegraphics[width=\wimg \linewidth, height= \wimg \linewidth, trim={4mm 4mm 4mm 4mm},clip]{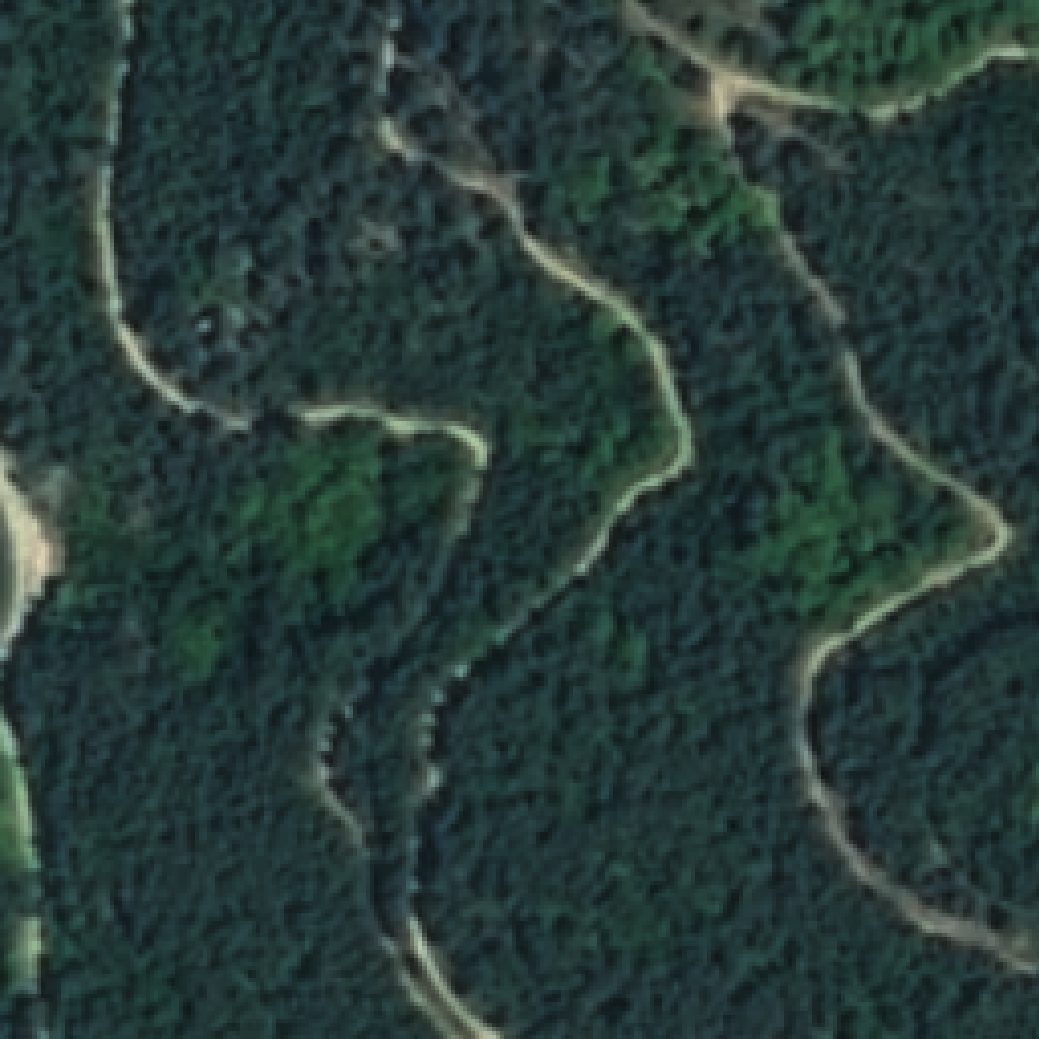}   
&
\includegraphics[width=\wimg \linewidth, height= \wimg \linewidth, trim={4mm 4mm 4mm 4mm},clip]{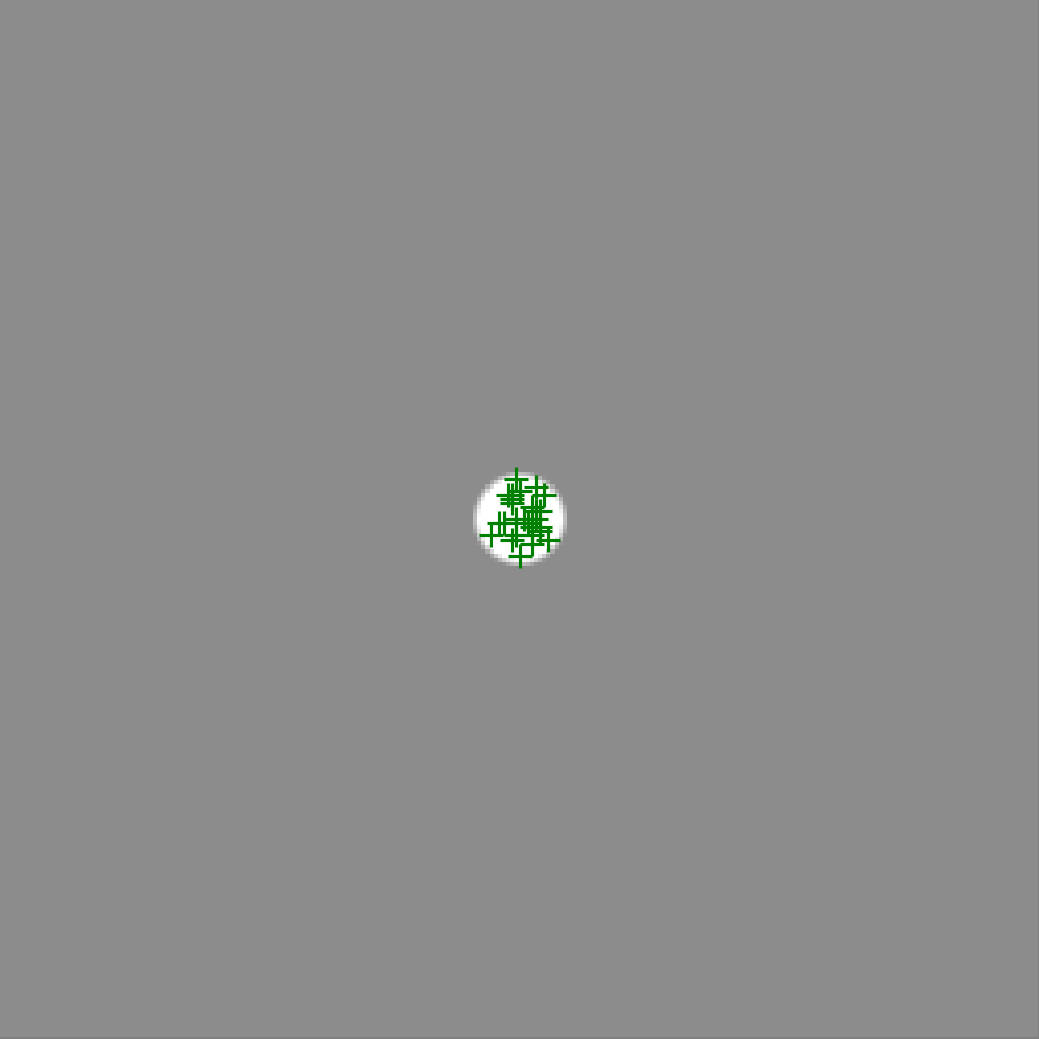} 
&
\includegraphics[width=\wimg \linewidth, height= \wimg \linewidth, trim={4mm 4mm 4mm 4mm},clip]{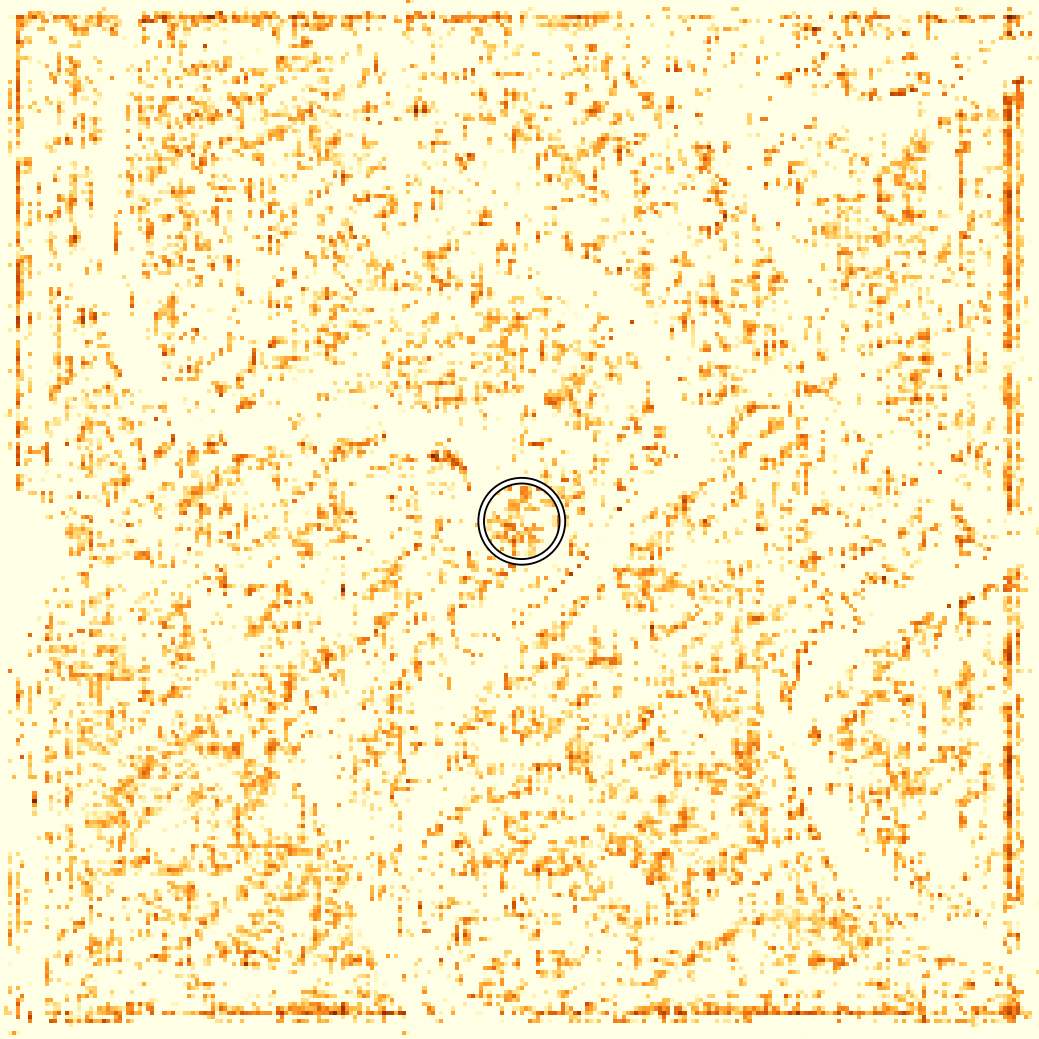} 
&
\includegraphics[width=\wimg \linewidth, height= \wimg \linewidth, trim={4mm 4mm 4mm 4mm},clip]{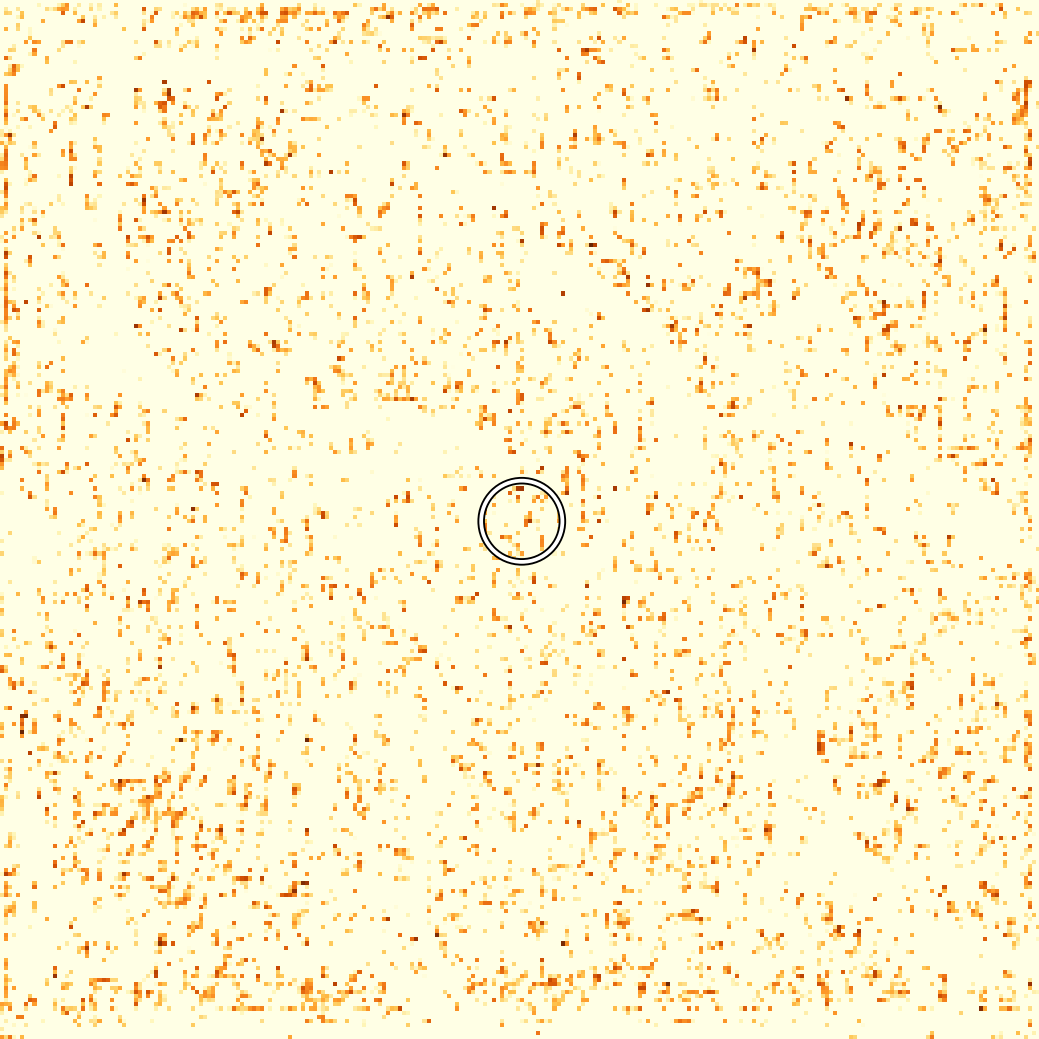} 
&
\includegraphics[width=\wimg \linewidth, height= \wimg \linewidth, trim={4mm 4mm 4mm 4mm},clip]{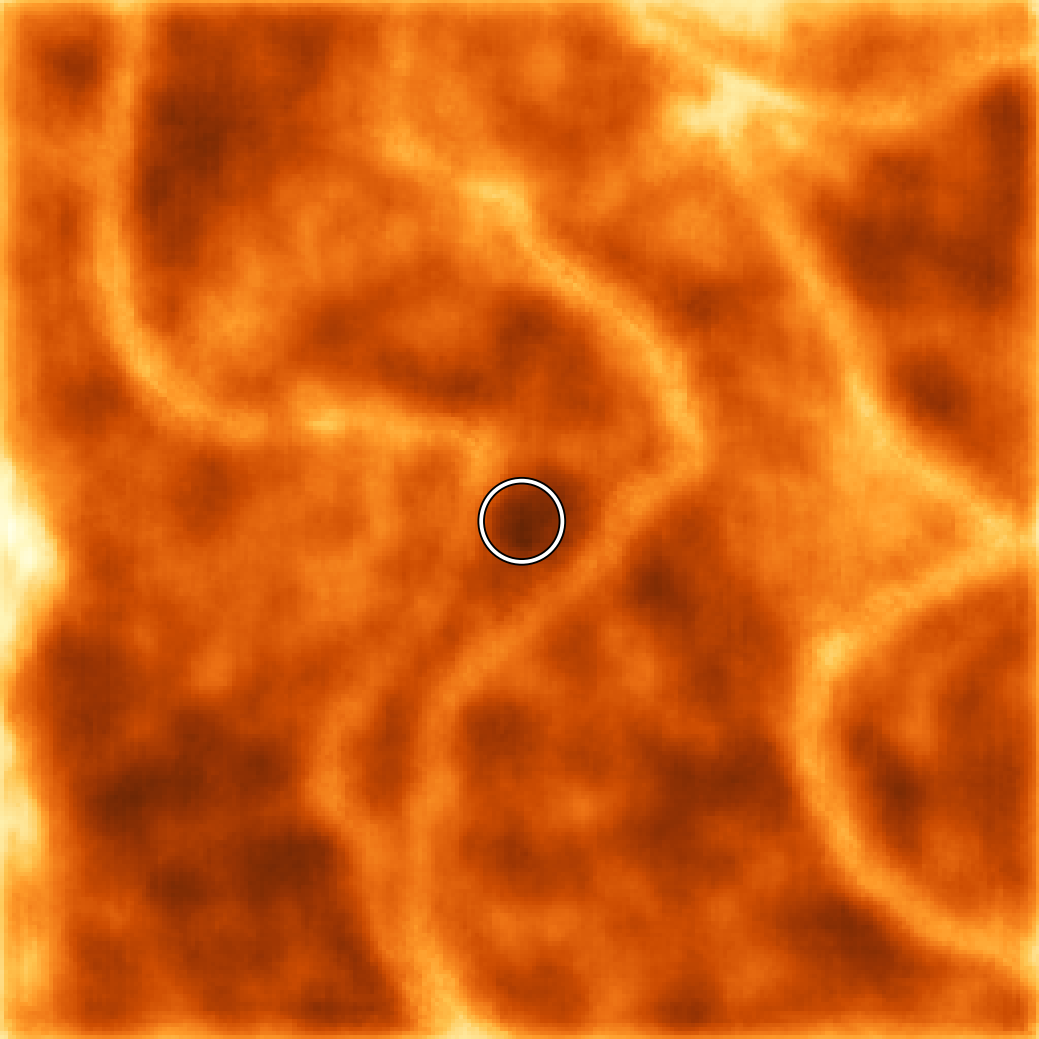}
&
\includegraphics[width=\wimg \linewidth, height= \wimg \linewidth, trim={4mm 4mm 4mm 4mm},clip]{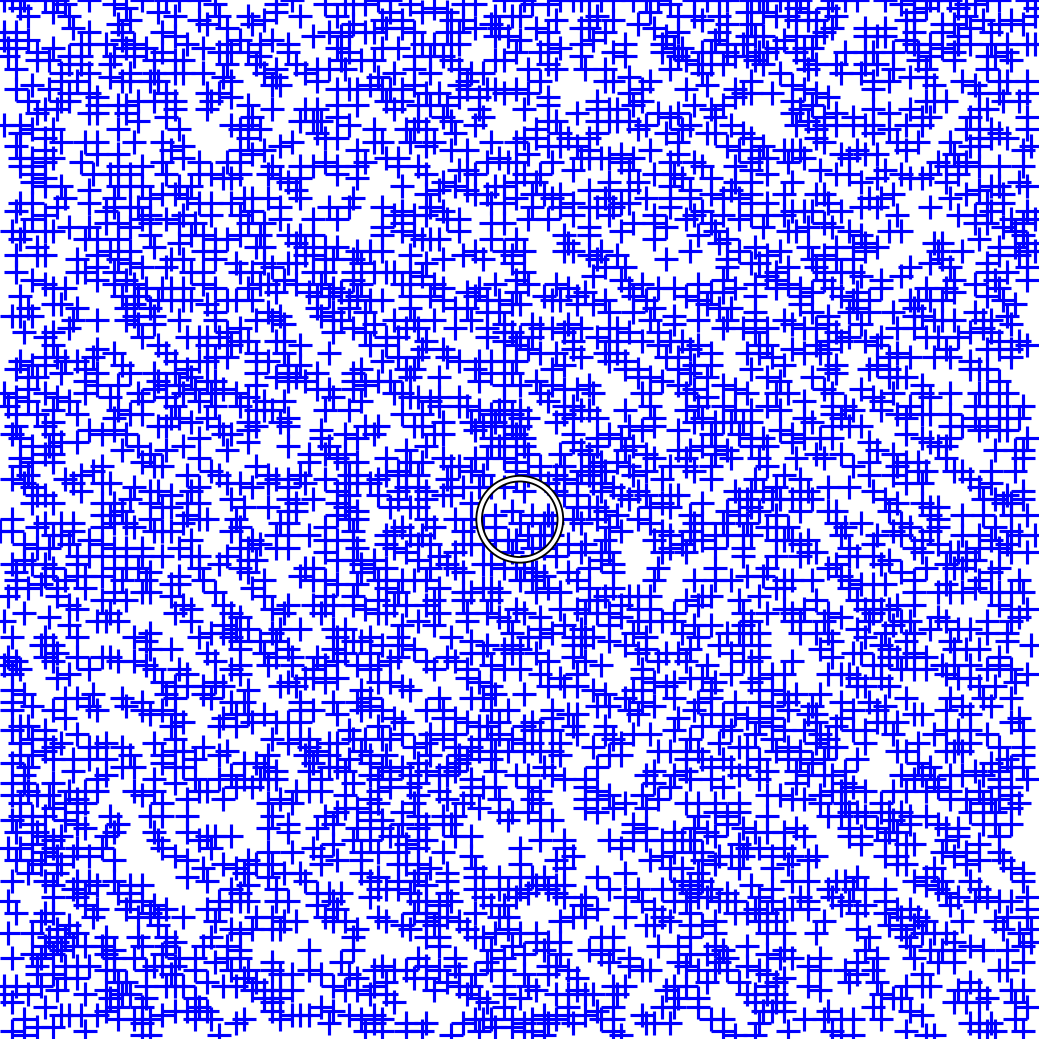}
\\
& & 27 trees & 23.7 trees & 17.0 trees & 18.3 trees & 21 trees
\\
&
\begin{subfigure}{\wimg\linewidth}
  \caption{~\\{Input} {RGB}}
  \label{fig:quali:a}
\end{subfigure}
&
\begin{subfigure}{\wimg\linewidth}
  \caption{{Strong} {labels}}
  \label{fig:quali:b}
\end{subfigure}
&
\begin{subfigure}{\wimg\linewidth}
  \caption{{\textsc{TreeMatch}}}
  \label{fig:quali:d}
\end{subfigure}
&
\begin{subfigure}{\wimg\linewidth}
  \caption{~\\{DM-} {Count}}
  \label{fig:quali:e}
\end{subfigure}
&
\begin{subfigure}{\wimg\linewidth}
  \caption{ {Density} {Regression}}
  \label{fig:quali:f}
\end{subfigure}
&
\begin{subfigure}{\wimg\linewidth}
  \caption{ {CenterNet}}
  \label{fig:quali:f}
\end{subfigure}
\end{tabular}
    \caption{{\bf Qualitative results across regions and sensors.}
Each row corresponds to a different dataset: China (Gaofen2), Rwanda (PlanetScope), and France (SPOT6). In the France split, ground-truth annotations are only available on 15m radius inventory plots.
From left to right: input RGB image (\subref{fig:quali:a}), strong manual annotations (\subref{fig:quali:b}), \textsc{TreeMatch}'s predictions (\subref{fig:quali:d}), DM-Count (\subref{fig:quali:d}), density regression baseline (\subref{fig:quali:e}), and CenterNet detector (\subref{fig:quali:f}).
\textsc{TreeMatch} produces spatially coherent density estimates across both isolated and dense tree regimes, while baselines either oversmooth dense regions or miss trees in cluttered areas.}
    \label{fig:quali}
\end{figure}

\subsection{Baselines and Metrics}
\label{subsec:baselines}

\paragraph{\textbf{Baselines and Competing Methods.}}
Following common practice in forest analysis, all  models use a UNet architecture~\cite{ronneberger2015u} with a ResNet50 encoder~\cite{he2016deep}. We chose this design fot its stability and ease of training, and evaluate other backbones in the ablation.
We evaluate the following training strategies:
\begin{enumerate}
    \item \textbf{Density Regression.}
    Tree annotations are converted into density maps using fixed-width Gaussian kernels, and the model is trained with an $\ell_2$ loss~\cite{wan_adaptive_2019, shi_counting_2019, ranjan_iterative_2018}. 

    \item \textbf{Count Regression.}
    Global sum pooling is applied to the 1-channel output of UNet to directly regress the total tree count for each image.

    \item \textbf{Detection.}
    We evaluate both box-based and point-based detectors, including YOLOv8~\cite{jocher_ultralytics_2023}, P2P~\cite{song_rethinking_2021}, TreeFormer \cite{amirkolaee_treeformer_2023}, and CenterNet~\cite{zhou_objects_2019}.

    \item \textbf{Distribution Matching.}
    We evaluate VRSNet \cite{luo2024vrsnet}, and DM-Count~\cite{wang_distribution_2020}, a crowd-counting method based on balanced optimal transport.

    \item \textbf{Bayesian Approach.}
    We evaluate a Bayesian appraoch designed for crow counting \cite{ma_bayesian_2019}.
    In addition, we reimplemented an uncertainty-aware dense regression baseline~\cite{lang_high-resolution_2023}, in which the model predicts both the mean and variance of the density at each pixel.
\end{enumerate}

\paragraph{\textbf{Evaluation Metrics.}}
We assess counting accuracy using the following metrics:
\begin{enumerate}
    \item \textbf{RMSE.} Image-level root mean squared error on tree counts, in \emph{trees per hectare}.
    
    \item \textbf{nMAE.} Dataset-level normalized mean absolute error.
    
    \item \textbf{\bf $\mathbf{R^2}$.} Coefficient of determination between predicted and ground-truth counts, computed at the image-level.
\end{enumerate}

\subsection{Results and Analysis}
\label{subsec:results}

\begin{table*}[t]
\centering
\caption{{\bf Quantitative Evaluation on \textsc{TinyTrees}.} 
We compare regression baselines (top), detection-based vision models (middle), and distribution-matching methods (bottom).
We indicate the type of supervision used during training for the best configuration: strong manual annotations and/or weak ALS-derived labels.
\textsc{TreeMatch} consistently outperforms competing approaches. \\
- \textit{did not converge} \quad 
n/a \textit{official implementation not compatible with partial annotation}}
\resizebox{\linewidth}{!}{
\begin{tabular}{l cc ccc ccc ccc}
 \toprule
 &
 \multirow{2}{*}{\rotatebox{90}{strong}}
 &
 \multirow{2}{*}{\rotatebox{90}{\!\!\!\!\!\!weak}}
 & \multicolumn{3}{c}{\small China / Gaofen2}
 &
 \multicolumn{3}{c}{\small Rwanda / PlanetScope}
 &
 \multicolumn{3}{c}{\small France / Spot6}\\
 \cmidrule(lr){4-6}
 \cmidrule(lr){7-9}
 \cmidrule(lr){10-12}
Method &  & & 
\small RMSE $\downarrow$  & \small \(R^2\) $\uparrow$ & \small nMAE $\downarrow$
&
\small RMSE $\downarrow$  & \small \(R^2\) $\uparrow$ & \small nMAE $\downarrow$
&
\small RMSE $\downarrow$  & \small \(R^2\) $\uparrow$ & \small nMAE $\downarrow$\\
\midrule
Density regression & \faCheck &  & 64.6 & 0.55 & 37.4 &  79.3 & 0.36 & 54.0 & 163.4 & 0.20 & 40.2 \\
\rowcolor{gray!10} Count regression & \faCheck  & \faCheck & 67.9 & 0.50 & 43.7 & 75.8 & 0.42 & 53.8 & 149.2 & 0.33 & 39.1 \\
Uncertainty-aware reg. & \faCheck & & 63.3 & 0.57 & 37.9 & 80.5 & 0.34 & 55.3 & 163.0 & 0.20 & 40.6 \\
\greyrule
\rowcolor{gray!10} YOLOv8 \cite{jocher_ultralytics_2023}& \faCheck  &  & - & - & - & \negphantom{4}448.7 & \negphantom{-}-0.50 & 85.0 & n/a & n/a & n/a \\
CenterNet \cite{zhou_objects_2019} & \faCheck & \faCheck & 82.9 & 0.26 & 51.8 & 98.9 & 0.01 & 69.8 & 181.5 & 0.01 & 45.6\\
\rowcolor{gray!10} P2P \cite{song_rethinking_2021} & \faCheck & \faCheck & 98.7 & -0.05 & 60.4 & - & - & - & 177.2 & 0.06 & 45.6 
\\
TreeFormer \cite{amirkolaee_treeformer_2023} & \faCheck & & 68.7 & 0.49 & 41.2 & 86.0 & 0.25 & 59.8 & 179.7 & 0.03 & 45.1 \\
\rowcolor{gray!10} Bayes. Crowd Count. \cite{ma_bayesian_2019} & \faCheck & & 168.8 &-2.06 & 134.1 & 80.5 & 0.34 & 60.9 & 162.0 & 0.21 & 43.7
\\\greyrule
VRSNet \cite{luo_vrsnet_2024} & \faCheck & & 67.5 & 0.51 & 41.2 & 81.7 & 0.32 & 56.9 & 165.1 & 0.18 & 40.3 \\
\rowcolor{gray!10} DM-Count \cite{wang_distribution_2020} & \faCheck & \faCheck & 65.0 & 0.55 & 39.6 & 76.8 & 0.40 & 53.4 & 158.0 & 0.25 & 40.9\\
\bf \textsc{TreeMatch} (ours) & \faCheck & \faCheck & \bf 60.6 & \bf 0.60 & \bf 36.6 & \bf 72.4 & \bf 0.47 & \bf 51.1 & \bf 147.2 & \bf 0.35 & \bf 37.4\\
 \bottomrule
  \end{tabular}
}

  \label{table:comparison}
\end{table*}

We report the performance of all methods across all splits in \cref{table:comparison}.
Overall, the task has proven to be challenging, with $R^2$ values remaining below $0.6$ for all models and regions.
This contrasts with related tasks such as crowd counting, where instance-level metrics can reach substantially higher accuracy.
Across all datasets and evaluation metrics, \textsc{TreeMatch} achieves the best overall performance.
Improvements are consistent in terms of RMSE, $R^2$, and normalized MAE, indicating robustness across continents, biomes, and imaging conditions.
Qualitative results are shown in \cref{fig:quali}.

\paragraph{\bf Performance Analysis.}
We make several observations:
\begin{itemize}

\item {\bf Regression baselines.}
Simple regression approaches provide surprisingly strong baselines.
In particular, image-level count regression performs competitively, highlighting the importance of global count supervision. Density regression also performs well, although its optimal Gaussian std of 4px leads to blurry outputs (see \cref{fig:quali:f}).
However, these methods are not well-suited to heterogeneous annotation quality.
In particular, incorporating weak supervision degrades the performance of density regression.

\item {\bf Detection-based methods.}
Detection models such as YOLOv8 and CenterNet perform poorly in this setting.
They struggle particularly in dense forest regions where tree boundaries are ambiguous and individual crowns are not clearly separable (see \cref{fig:quali:f}).
In several cases, these models either fail to converge or produce unstable predictions, confirming that classical instance detection is ill-suited to satellite-based tree counting across heterogeneous forest regimes.

\item {\bf Distribution-matching methods.}
DM-Count~\cite{wang_distribution_2020} performs relatively well, confirming the relevance of transport-based formulations for counting tasks.
However, its reliance on balanced OT and TV losses lead to over-confident activations (see \cref{fig:quali:f}) and limit robustness under partial or noisy annotations.
Our unbalanced formulation is blurrier but generalizes better, consistently improving upon DM-Count across all datasets.

\end{itemize}

\paragraph{\textbf{Efficiency.}}
Our model is computationally efficient at inference, requiring only a single forward pass.
It processes approximately 99M pixels/s, corresponding to about 887\,km$^2$/s for PlanetScope imagery.
Training is also fast and stable: a full training run on a single data split completes in under one hour on a single NVIDIA 3090 GPU.
These characteristics suggest that the approach is well-suited for large-scale deployment.

\paragraph{\textbf{Spatial Generalization.}}
In the previous experiments, we trained one model per region–sensor pair.
An important question is whether a single model can generalize across biomes and satellite platforms.
\Cref{tab:gene} reports cross-region evaluation results, as well as the performance of a jointly trained global model.
We observe that some regions are substantially easier to learn from than others.
In particular, the model trained on the China split achieves the best overall performance.
This can likely be attributed to the higher spatial resolution of Gaofen imagery and the quality of the photo-interpreted strong labels.
In contrast, the model trained on the Rwanda split generalizes poorly, likely due to the specificity of its tropical biome and the calibration of its sensor.

While a model jointly trained on all splits performs slightly worse than specialized models on their training split, it still achieves strong overall results.
This suggests that cross-region generalization is feasible and could be further improved by leveraging sensor-agnostic architectures~\cite{xiong2024dofa,astruc2024anysat} and larger, more diverse training datasets.

\begin{table}[t]
    \centering
    \begin{tabular}{c@{\hspace{.1\linewidth}}c}
\begin{minipage}{0.35\linewidth}
\caption{{\bf Cross-Region Spatial Generalization.}
Each row corresponds to a model trained on a specific region (or jointly on all regions), and each column reports performance (nMAE) on a target region.}
    \label{tab:gene}
\end{minipage}
\begin{minipage}{0.60\linewidth}\vspace{-3mm}
\resizebox{\linewidth}{!}{
\begin{tabular}{l ccc c}
\toprule
 \diagbox{Training}{Evaluation}    & China & Rwanda & France &  Overall \\\midrule
China     & \applycolorchina{36.6} & \applycolorafrica{99.9} & \applycolorfrance{76.9} & \applycolorall{84.8}\\
Rwanda    & \applycolorchina{1066} & \applycolorafrica{51.1} & \applycolorfrance{180} & \applycolorall{287} \\
France    & \applycolorchina{154} & \applycolorafrica{99.9} & \applycolorfrance{37.4} & \applycolorall{111}  \\\greyrule
Joint Training & \applycolorchina{42.4} & \applycolorafrica{51.7} & \applycolorfrance{42.9} & \applycolorall{49.3} \\\bottomrule
\end{tabular}}
\end{minipage}
\end{tabular}
\end{table}



\subsection{Ablation Study and Perspectives}
\label{subsec:ablation}

We conduct an ablation study to quantify the contribution of each component of our framework.
Results are reported in Table~\ref{tab:ablation}.

\begin{itemize}

\item[\textsc{a})] \textbf{Role of supervision sources.}
Removing weak, ALS- or model-derived supervision (\emph{No Train-weak}) increases RMSE from 72.4 to 76.1, showing that large-scale noisy labels provide a valuable complementary training signal.
Removing strong, manual supervision (\emph{No Train-strong}) leads to a larger degradation, with RMSE increasing to 82.4, confirming the importance of accurate counts for calibrating the model. Uniform sampling slightly improves performance on the Rwanda split (71.0 RMSE), although the effect was not consistent on other splits due to varying strong/weak ratios.

\item[\textsc{b})] \textbf{Loss formulation.}
Training with count supervision only ($\mathcal{L}_{\text{count}}$ \emph{only}) increases RMSE to 75.5, highlighting the importance of spatial distribution matching.
Replacing unbalanced OT with balanced OT further degrades performance (78.3 RMSE), confirming that explicitly tolerating mass mismatch is critical under weak supervision.
Removing residual correction (\emph{No residuals}) increases RMSE to 79.0, demonstrating the benefit of our correction mechanism for noisy annotations.

\item[\textsc{c})] \textbf{Choice of backbone.}
Using a ViT-S backbone yields performance nearly identical to the baseline (72.5 RMSE), indicating limited sensitivity to moderate capacity reduction.
A hierarchical Swin-S backbone slightly improves performance (71.0 RMSE), consistent with prior findings that hierarchical encoders are well-suited to forest imagery analysis \cite{fogel_open-canopy_2025}.
\end{itemize}

\paragraph{\textbf{Impact of Annotation Type.}}
For the China split, weak supervision is provided as continuous forest cover maps with constant density rather than pointwise annotations, due to the low resolution of the available canopy height data.
For comparison, we derived pointwise weak annotations using the local peak detection procedure employed for France.
This alternative leads to a slight performance decrease (nMAE = 39.6 vs.\ 36.6).
These results indicate that our model can accommodate both pointwise and continuous weak labels, and that adapting the annotation form to its corresponding modality improves performance. 

\paragraph{\textbf{Limitations and Perspectives.}}

Optimal transport losses incur a quadratic memory complexity $O(N^2)$ with respect to the number of pixels $N = H \times W$, which limits scaling to larger images with broader spatial context. Sinkhorn-based implementations alleviate part of this issue through efficient computation, but further improvements could be achieved through more memory-efficient optimal transport algorithms~\cite{sun2024map}.

Tree counting and canopy height estimation are closely related problems, suggesting strong potential for joint modeling. Tree counts constrain stand density, while height maps capture vertical forest structure. These signals are linked through well-established ecological relationships, such as allometric scaling between tree height and volume~\cite{jucker_allometric_2017}, and self-thinning dynamics relating stand volume and density~\cite{yoda_self-thinning_1963}. Combining height and density estimates could therefore lead to more consistent estimates of forest structure and biomass. The increasing availability of large-scale canopy height products~\cite{fogel_open-canopy_2025, liu_overlooked_2023, pauls_estimating_2024} makes such joint modeling increasingly feasible at continental scales.

\begin{table*}[t]
\centering
\begin{tabular}{c@{\hspace{.03\linewidth}}c}
\raisebox{-2mm}
{\begin{minipage}{0.36\linewidth}
 \caption{{\bf Ablation Study.}
Effects of supervision strategy (\textsc{a}), loss formulation (\textsc{b}), and backbone architecture (\textsc{c}) on performance. All results are reported for the Rwanda split.}
\label{tab:ablation}
\end{minipage}}
&
\begin{minipage}{0.6\linewidth}
\resizebox{\linewidth}{!}{
\begin{tabular}{cl cccc}
\toprule
     && RMSE $\downarrow$ & R2 $\uparrow$ & nMAE $\downarrow$ \\\midrule
   &Used config. & 72.4 & 0.47 & 51.1 \\\greyrule
   \textsc{a} &No Train-weak & 76.1 & 0.41 & 52.9\\
   \textsc{a} &No Train-strong& 82.4 & 0.31 & 57.0\\
   \textsc{a} &Uniform sampling& 71.4 & 0.48 & 50.8 \\
   \textsc{b} &$\mathcal{L}_{\text{count}}$ only & 75.5 & 0.42 & 53.5\\
   \textsc{b} &Balanced OT & 78.3 & 0.38 & 53.6\\
   \textsc{b} &No residuals & 79.0 & 0.37 & 55.0\\
   \textsc{c} &ViT-S backbone & 72.5 & 0.47 & 51.6 \\
   \textsc{c} &SWIN-S backbone & \bf 71.0 & \bf 0.49 & \bf 50.6\\\bottomrule
\end{tabular}
}
\end{minipage}
\end{tabular}
\end{table*}

\section{Conclusion}
In this work, we addressed the problem of counting individual trees from satellite imagery in the presence of intrinsic ambiguity and heterogeneous supervision. We proposed an approach based on unbalanced optimal transport that can naturally accommodate localization ambiguity, uncertain total counts, and annotations of varying quality. 
Experiments on our proposed \textsc{TinyTrees} benchmark, spanning three continents and satellite sensors, demonstrate consistent improvements over detection-based, regression-based, and distribution-matching baselines. These results suggest that unbalanced optimal transport provides a principled and flexible foundation for large-scale tree counting from satellite imagery under real-world annotation constraints. 

\section*{Acknowledgement}

This work was funded by the National Key Research and Development Program of China (grant agreement no. 2023YFF1305700) and the Novo Nordisk Foundation (grant number NNF21OC0069116).
This work was supported by the ANR project SHARP ANR-23-PEIA-0008 in the context of the PEPR IA and Hi! PARIS (ANR-23-IACL-0005), and the Danish National Research Foundation through the Center for Remote Sensing and Deep Learning of Global Tree Resources (TreeSense, DNRF192).

We thank the French Forestry Office (ONF) for providing the forest inventory data used in the France split.
%
%
\bibliographystyle{unsrtnat}
\bibliography{ref}

\clearpage
\pagebreak
\setcounter{section}{0}
\renewcommand*{\theHsection}{App.\the\value{section}}
\renewcommand*{\thesection}{A.\arabic{section}}
\setcounter{figure}{0}
\renewcommand*{\theHfigure}{App.\thefigure}
\renewcommand\thefigure{A.\arabic{figure}}
\setcounter{table}{0}
\renewcommand*{\theHtable}{App.\thetable}
\renewcommand\thetable{A.\arabic{table}}
\maketitlesupplementary
\FloatBarrier

This appendix provides additional ablations and qualitative illustrations
(\cref{sec:sup:ablation}), implementation details (\cref{sec:sup:details}),
and further information about the data used to train our models
(\cref{sec:sup:qualif}).

\section{Additional Analysis}
\label{sec:sup:ablation}

\paragraph{\textbf{Impact of Curriculum.}}
We evaluate the impact of disabling the curriculum schedule $\lambda(t)$ used for the correction mechanism by fixing $\lambda(t)=0.8$ for all $t$.
Results are reported in \cref{tab:sup:ablation}.
We observe that relying on the correction mechanism too early, before the model has learned a reasonable density estimate, leads to a decrease in performance.

\begin{table}
    \caption{{\bf Impact of Curriculum}.}
    \label{tab:sup:ablation}
    \centering
    \begin{tabular}{l cccc}
\toprule
     & RMSE $\downarrow$ & R2 $\uparrow$ & nMAE $\downarrow$ \\\midrule
   \bf With Curriculum & \bf 72.4 & \bf 0.47 & \bf 51.1 \\\greyrule
   No curriculum & 76.3 & 0.41 & 52.6
   \\\bottomrule
\end{tabular}
\end{table}

\begin{figure}
\centering
\begin{tabular}{ccc}
     \begin{subfigure}{0.32\linewidth}
         \includegraphics[width=\linewidth]{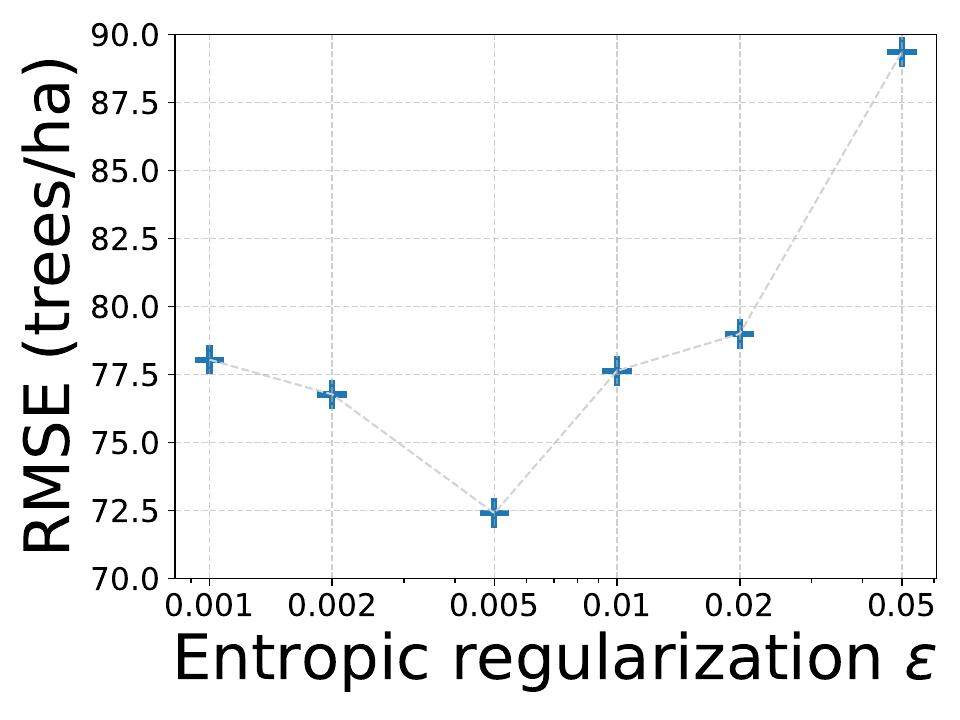}
     \end{subfigure}
     &
     \begin{subfigure}{0.32\linewidth}
         \includegraphics[width=\linewidth]{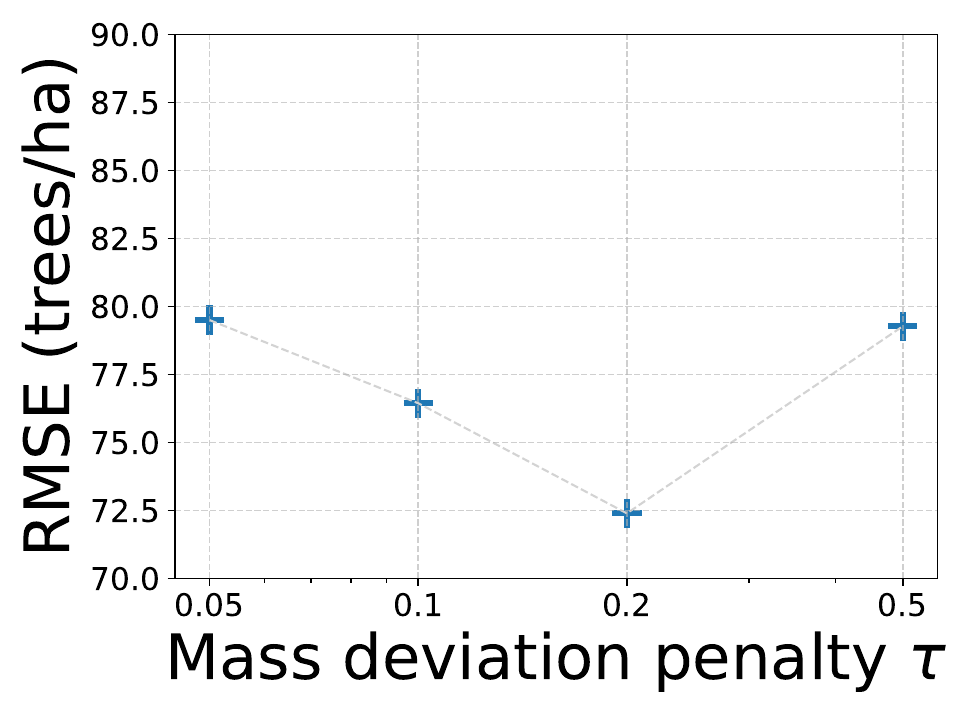}
     \end{subfigure}
     &
     \begin{subfigure}{0.32\linewidth}
         \includegraphics[width=\linewidth]{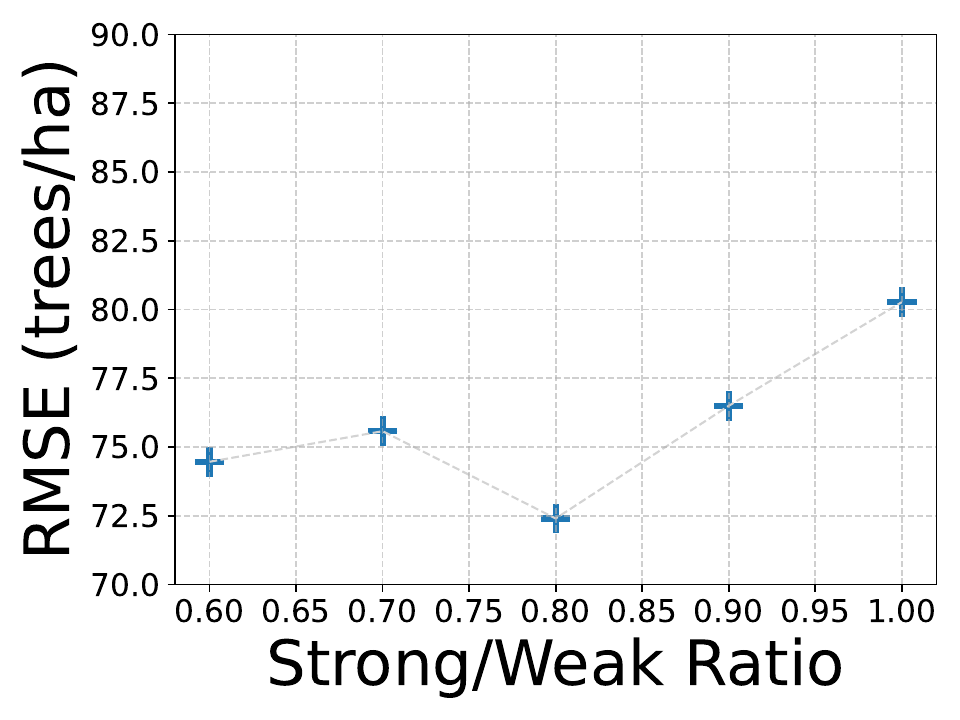}
     \end{subfigure}
\end{tabular}
\caption{{\bf Sensitivity analysis.} Impact of key hyperparameters on model performance (PlanetScope split).}
\label{fig:suppl:sensitivity}
\end{figure}

\paragraph{\textbf{Sensitivity Analysis.}}
\Cref{tab:sup:parameter} reports the selected values of the main hyperparameters: the sampling ratio between strong and weak annotations, the entropic regularization strength $\epsilon$, the mass deviation penalty $\tau$, and the maximum value $\alpha$ reached by the marginal correction scheduler $\lambda(t)$. 
We analyze the sensitivity of the model to these parameters in \Cref{fig:suppl:sensitivity}. 

Overall, $\epsilon$ has a stronger influence on performance than $\tau$ and the sampling ratio. The selected values of $\tau$ vary across dataset splits, likely reflecting differences in the quality of the weak supervision. In contrast, identical values of $\epsilon$ and $\alpha$, and similar sampling ratios, consistently perform well across all splits, suggesting that these parameters are relatively stable.

\begin{table}
  \caption{{\bf Selected hyperparameters.}
    Values of the main hyperparameters used across datasets.
    \emph{Strong/Weak} denotes the sampling ratio between strong and weak annotations.
    $\epsilon$ is the entropic regularization strength,
    $\tau$ is the mass deviation penalty,
    and $\alpha$ is the maximum value reached by the marginal correction scheduler $\lambda(t)$.}
    \label{tab:sup:parameter}
    \centering
    \begin{tabular}
{l c@{\qquad}c@{\qquad}c@{\qquad}c}
\toprule
     & \makecell{Strong/weak \\ ratio} & $\epsilon$ & $\tau$ & $\alpha$ \\\midrule
   Gaofen & 80\% & 0.005 & 0.1 & 0.8\\
   PlanetScope & 80\% & 0.005 & 0.2& 0.8\\
   SPOT & 70\% & 0.005 & 0.05& 0.8\\\bottomrule
\end{tabular}
\end{table}

\begin{figure}
    \centering
    \begin{tabular}{c@{\hspace{.1\linewidth}}c}
\begin{minipage}{0.35\linewidth}
\caption{{\bf Computational cost.}
We report the computational cost of a training step (ms) as a function of batch size, with the same feature extractor (UNet-R50).
}
    \label{fig:sup:trainstep}
\end{minipage}
\begin{minipage}{0.60\linewidth}\vspace{-3mm}
\resizebox{\linewidth}{!}{
    \includegraphics[width=0.5\linewidth]{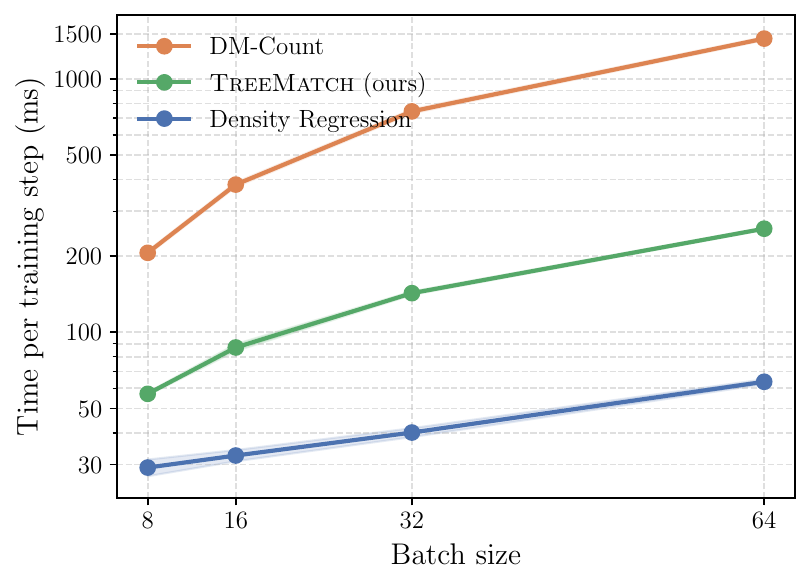}}
\end{minipage}
\end{tabular}
\end{figure}

\begin{figure}
    \centering
     \captionsetup[subfigure]{%
   justification=centering,
   labelfont=small,
   textfont=small,
}
\def\wimg{0.18}
\scriptsize
\begin{tabular}{l cc cc}
\scriptsize \rotatebox{90}{\quad\;\;\shortstack{China \\ Gaofen2}}
&
\includegraphics[width=\wimg \linewidth, height= \wimg \linewidth]{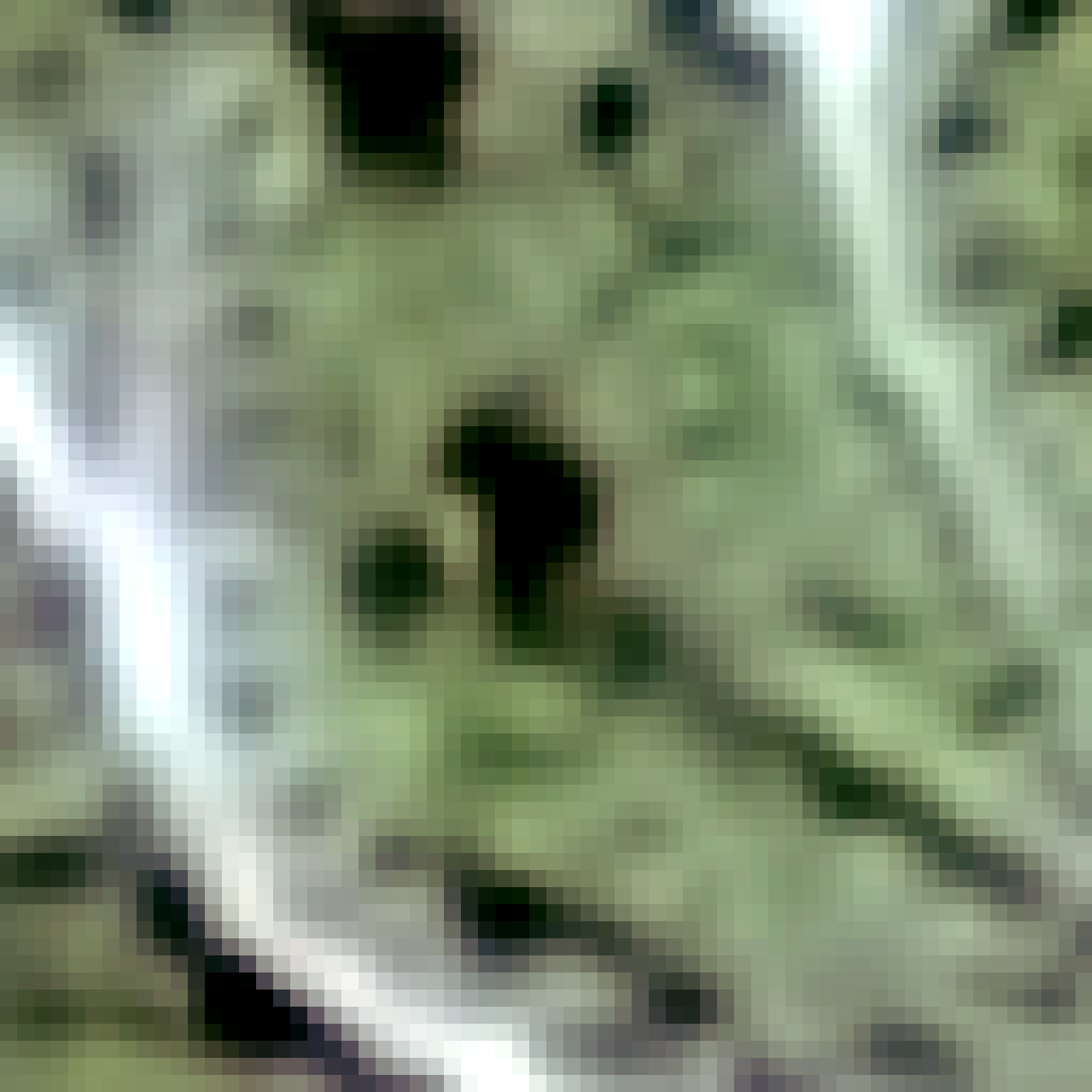}   
&
\includegraphics[width=\wimg \linewidth, height= \wimg \linewidth]{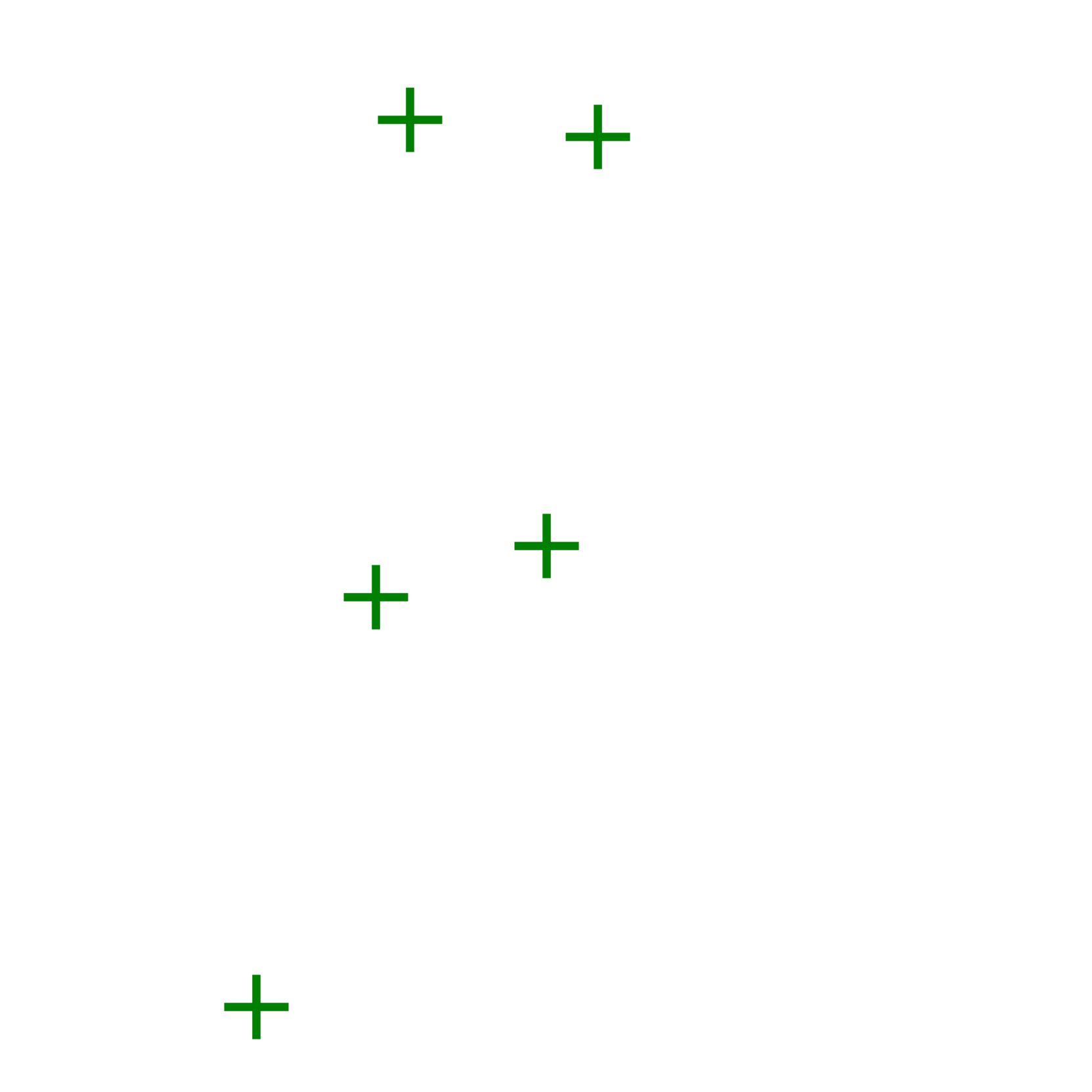} 
&
\includegraphics[width=\wimg \linewidth, height= \wimg \linewidth]{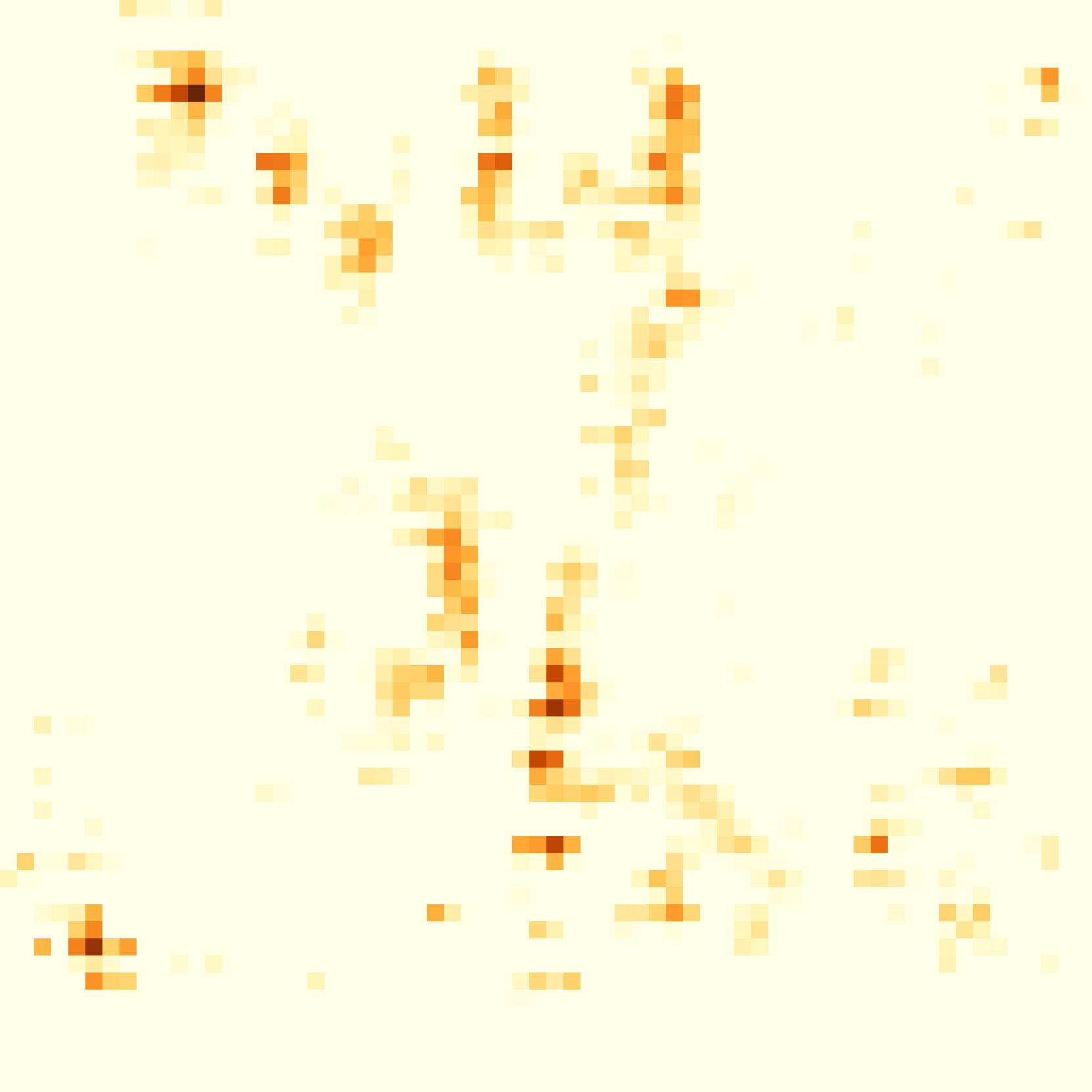} 
&
\includegraphics[width=\wimg \linewidth, height= \wimg \linewidth]{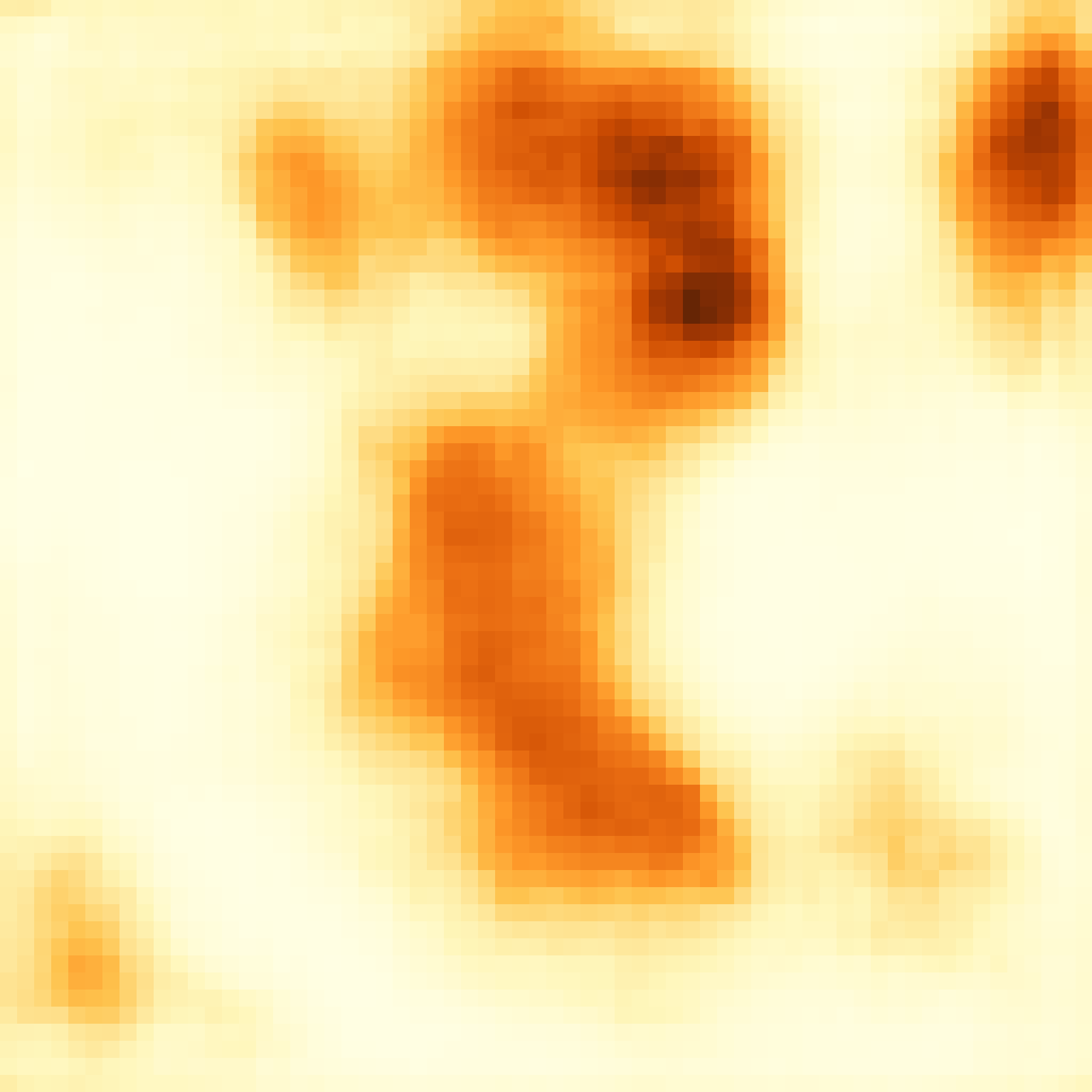} 
\\
& & 5 trees & 24.0 trees & 14.4 trees
\\
\scriptsize \rotatebox{90}{\;\;\shortstack{Rwanda \\ PlanetScope}}
&
\includegraphics[width=\wimg \linewidth, height= \wimg \linewidth]{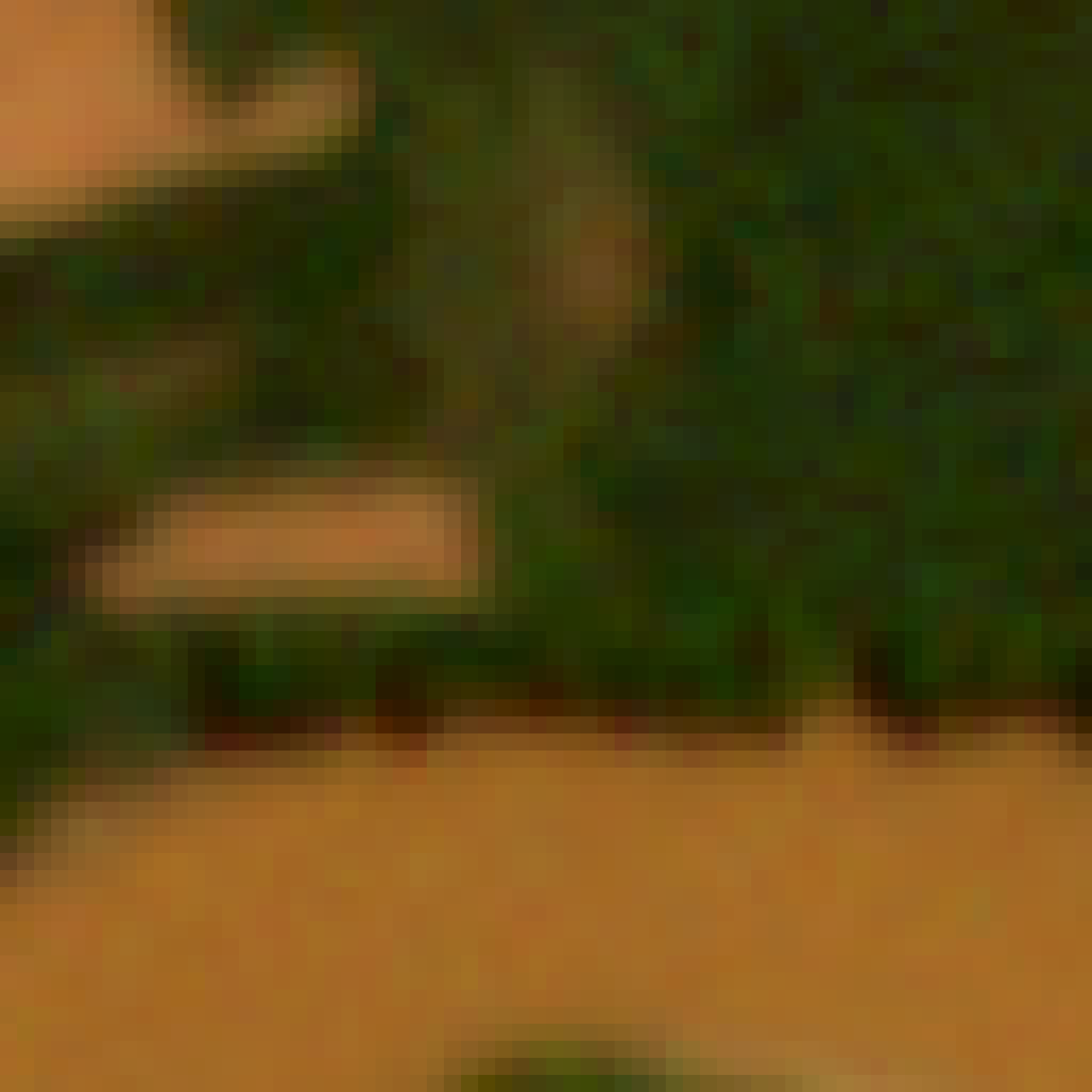}  
&
\includegraphics[width=\wimg \linewidth, height= \wimg \linewidth]{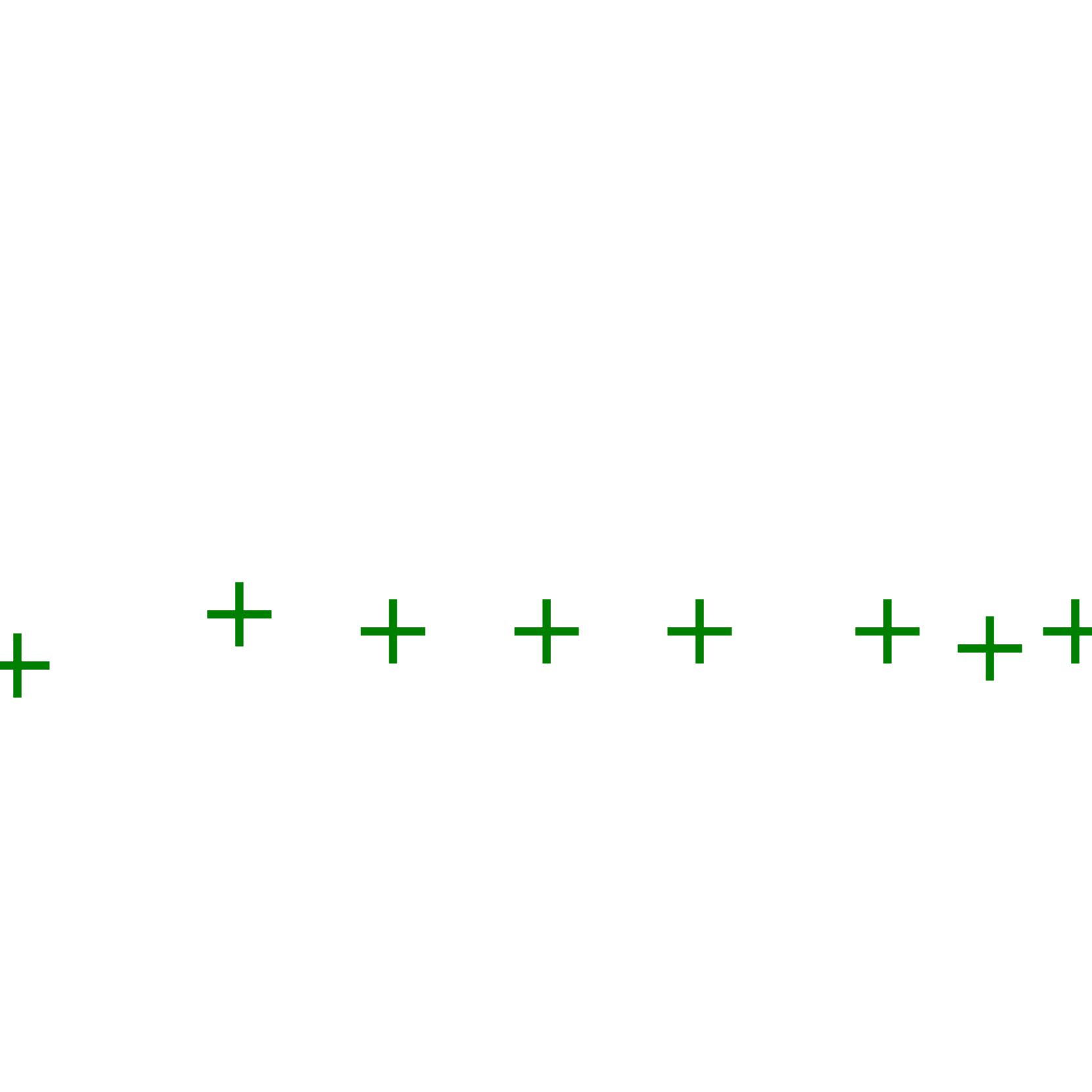} 
&
\includegraphics[width=\wimg \linewidth, height= \wimg \linewidth]{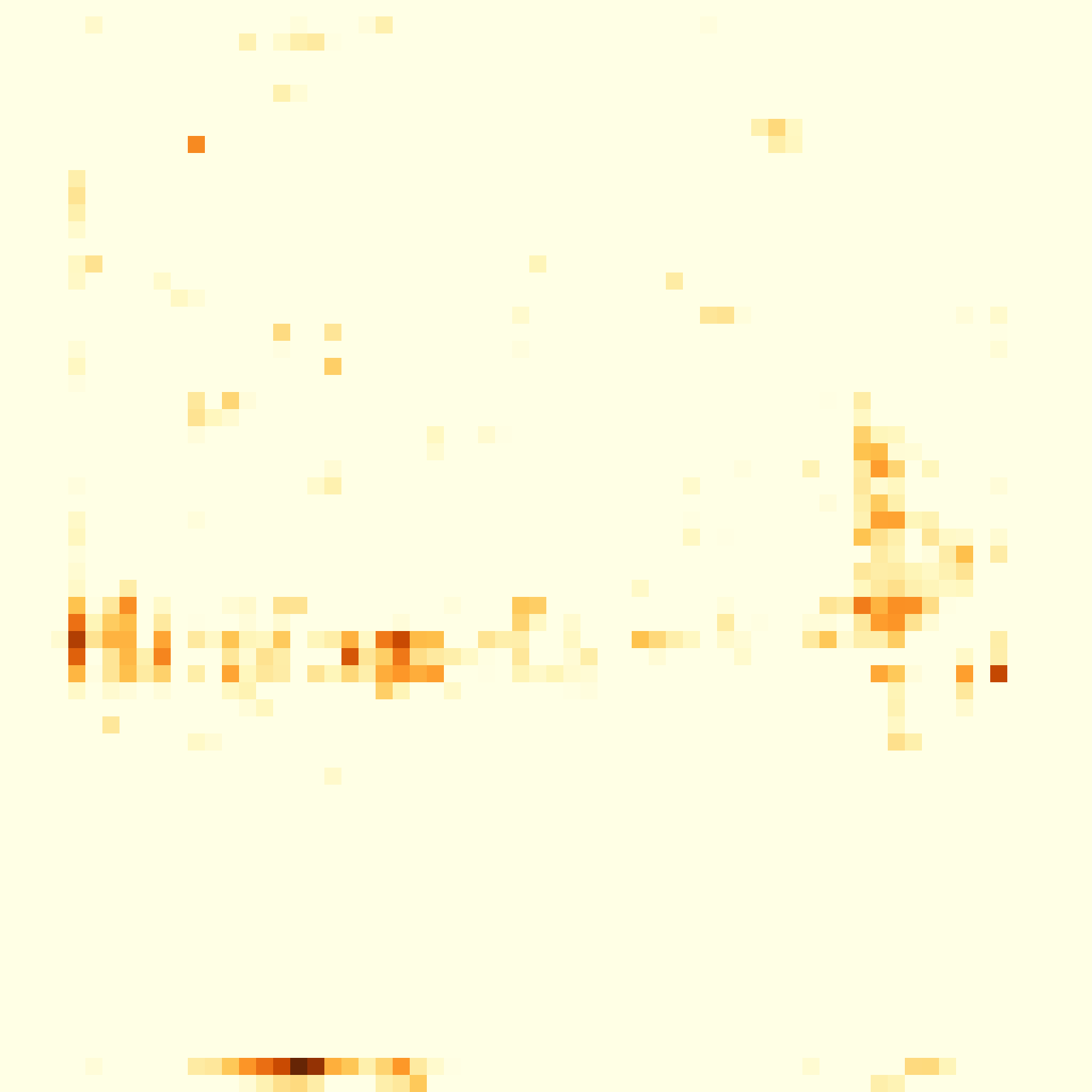} 
&
\includegraphics[width=\wimg \linewidth, height= \wimg \linewidth]{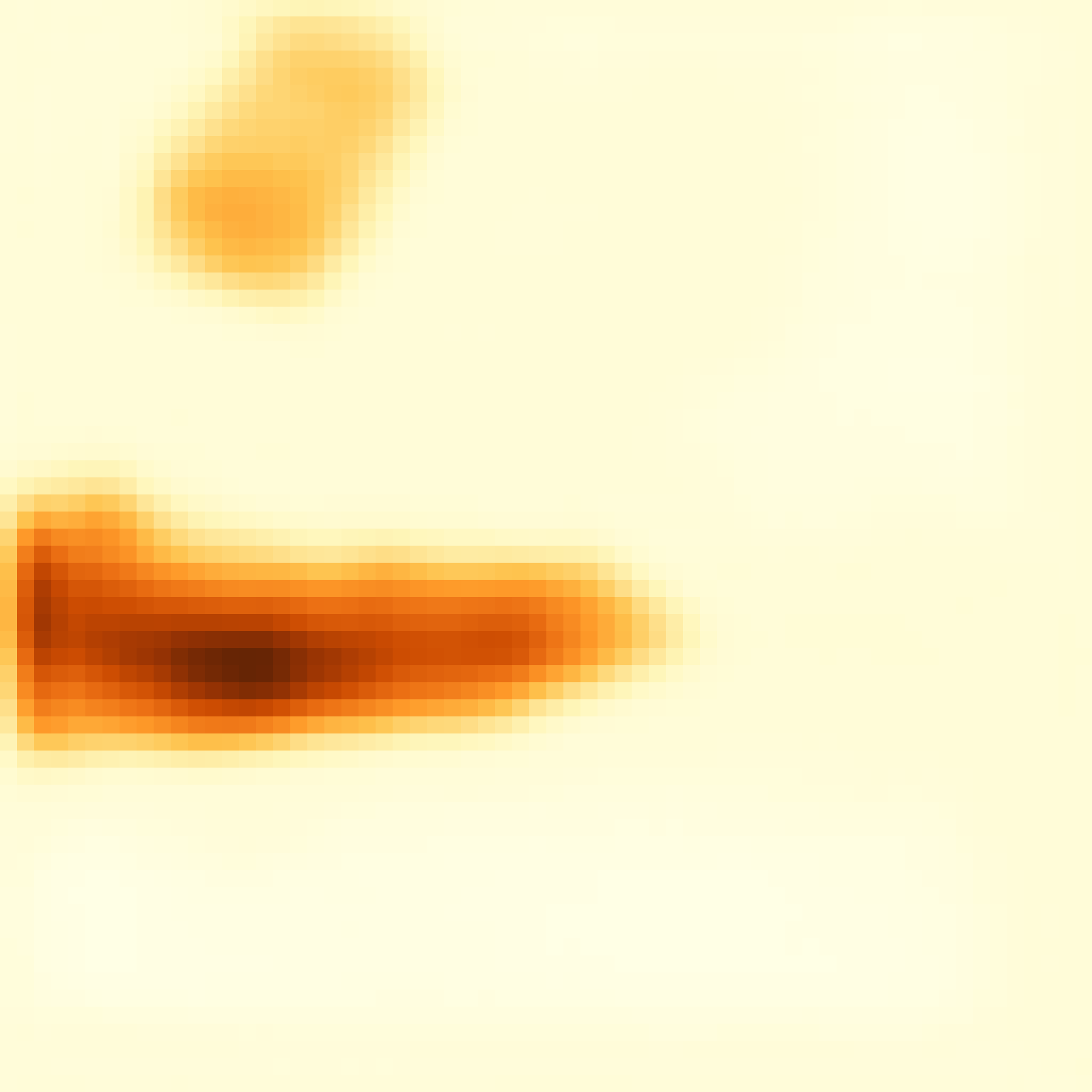} 
\\
& & 8 trees & 86.1 trees & 63.2 trees\\
\scriptsize \rotatebox{90}{\quad\;\;\shortstack{France \\ SPOT6}}
&
\includegraphics[width=\wimg \linewidth, height= \wimg \linewidth]{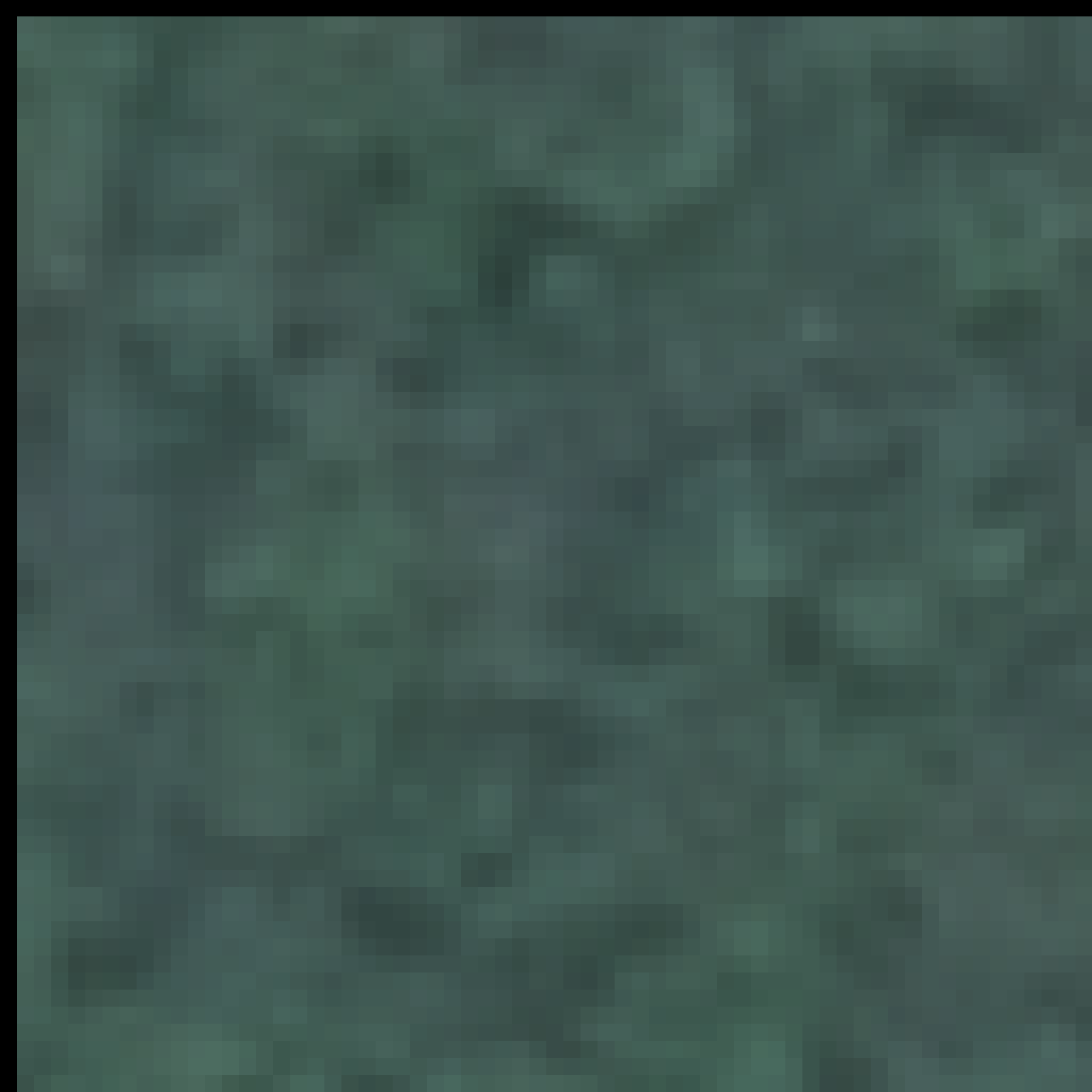}   
&
\includegraphics[width=\wimg \linewidth, height= \wimg \linewidth]{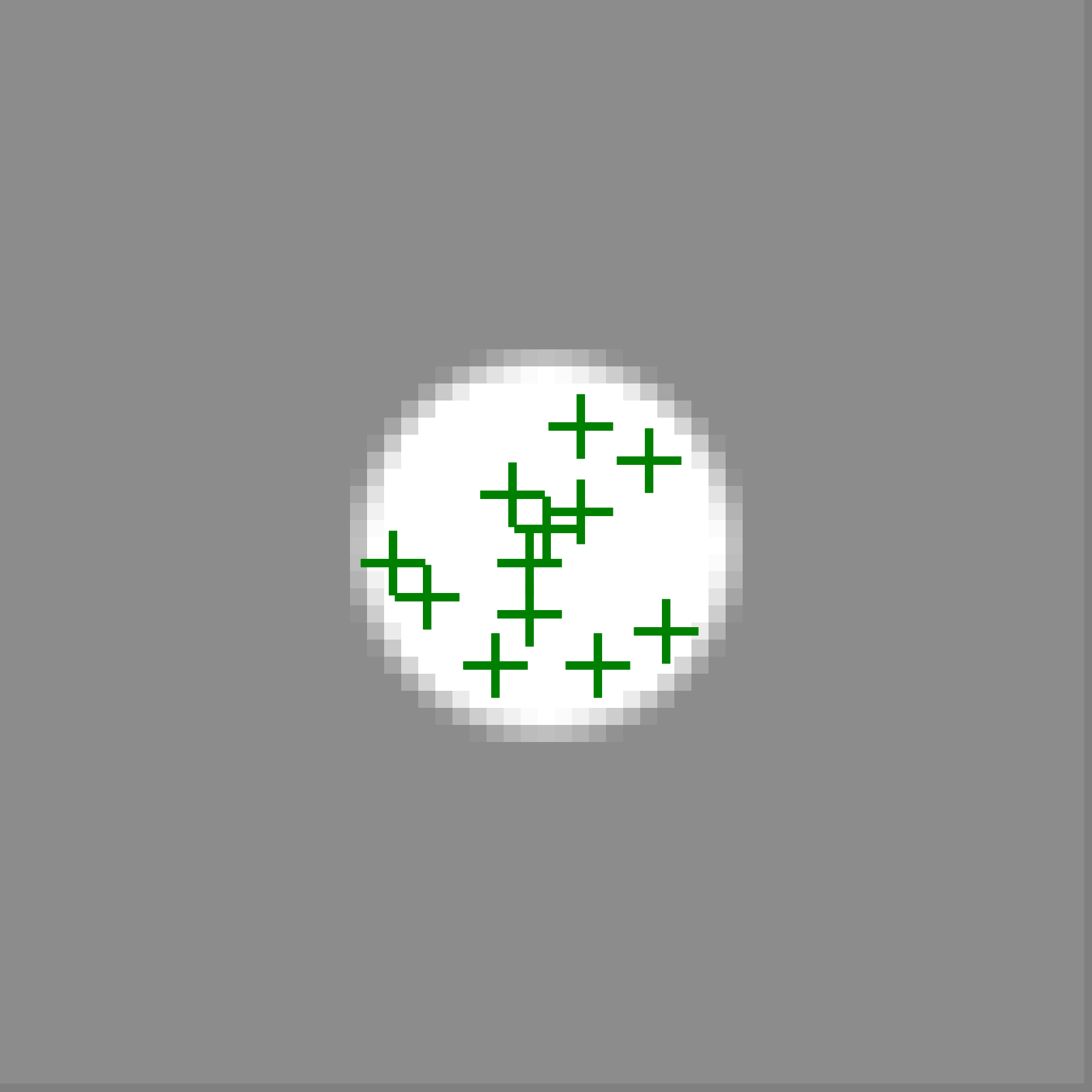} 
&
\includegraphics[width=\wimg \linewidth, height= \wimg \linewidth]{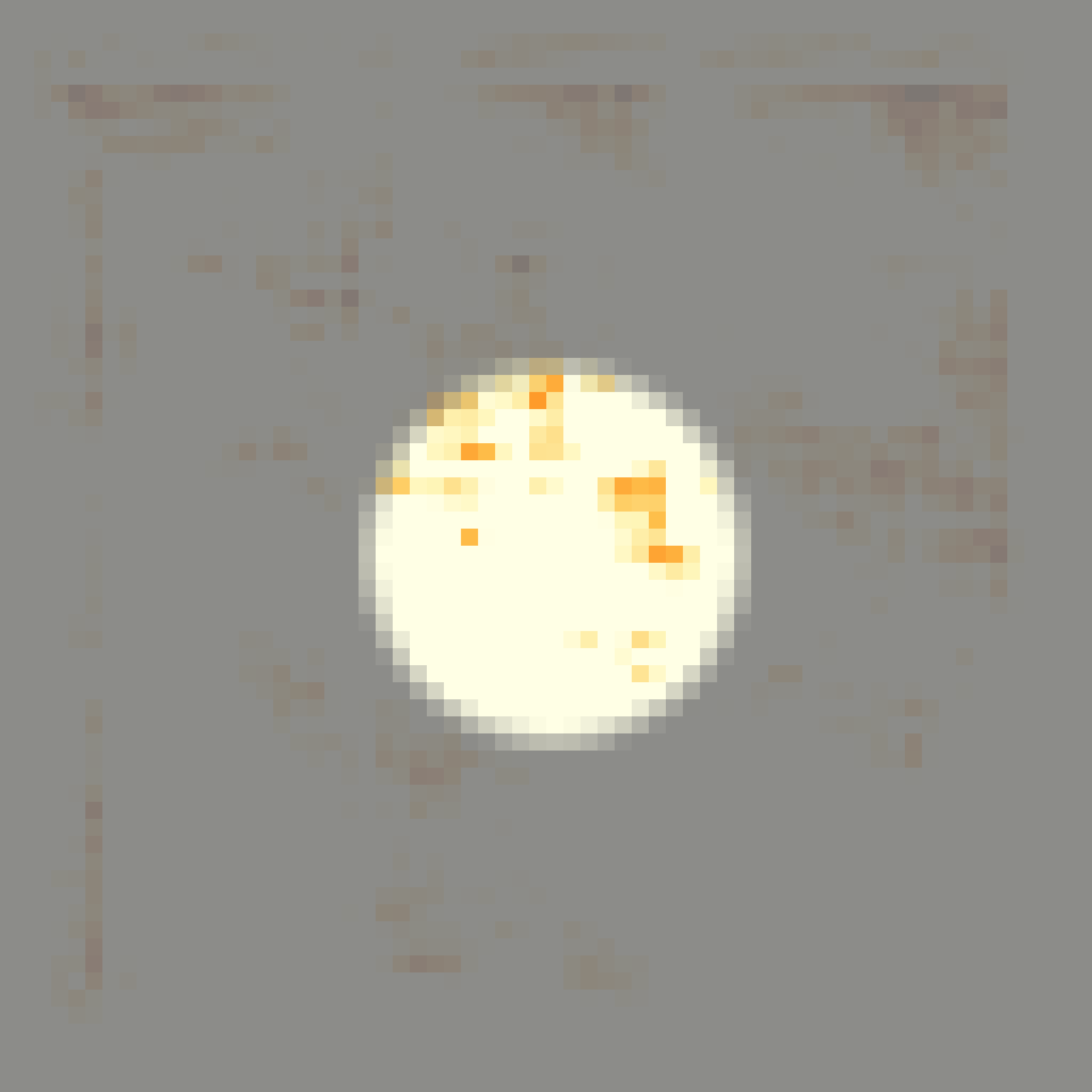} 
&
\includegraphics[width=\wimg \linewidth, height= \wimg \linewidth]{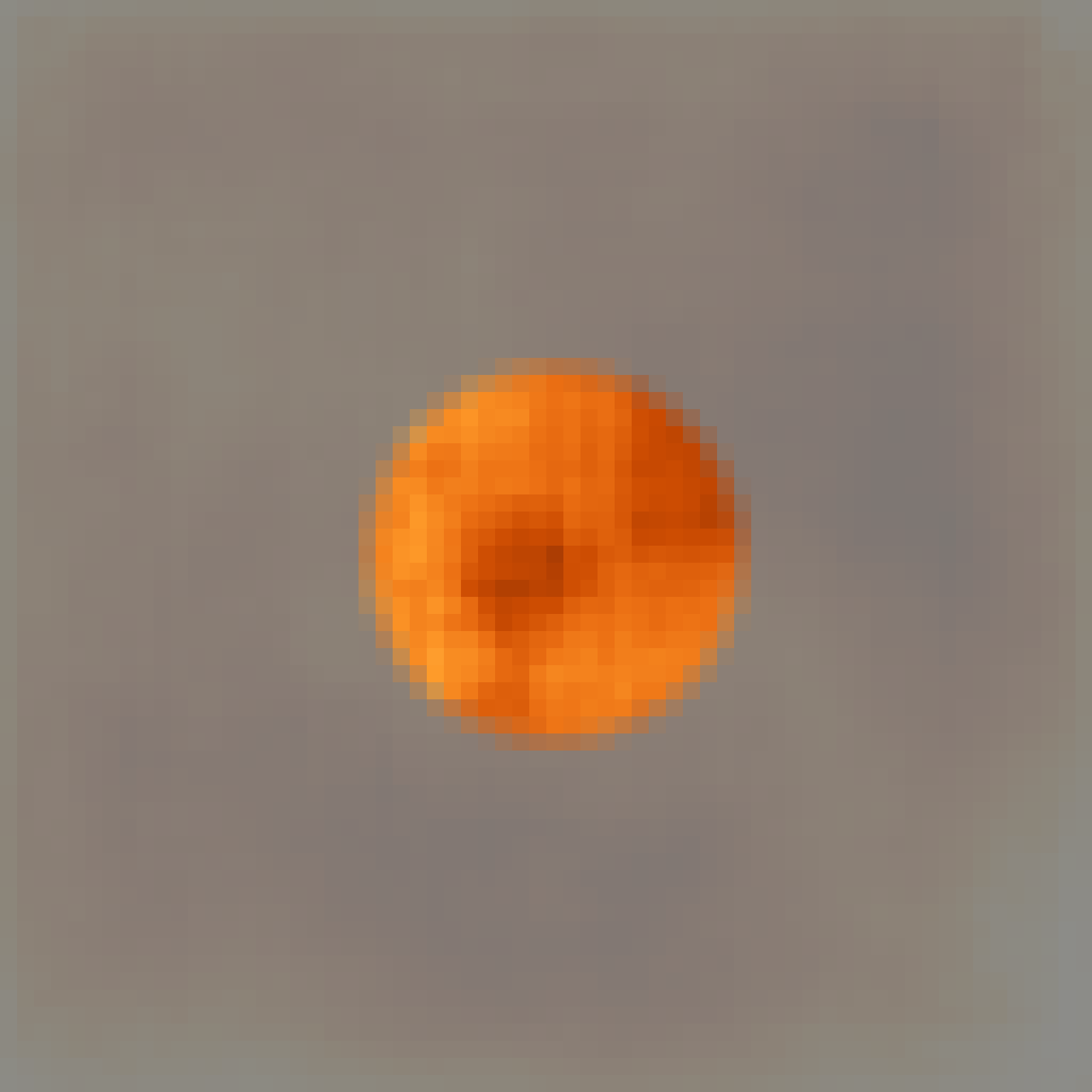} 
\\
& & 12 trees & 7.1 trees & 9.0 trees
\\
&
\begin{subfigure}{\wimg\linewidth}
  \caption{~\\{Input} {RGB}}
  \label{fig:supp:quali:a}
\end{subfigure}
&
\begin{subfigure}{\wimg\linewidth}
  \caption{{Strong} {labels}}
  \label{fig:supp:quali:b}
\end{subfigure}
&
\begin{subfigure}{\wimg\linewidth}
  \caption{{\textsc{TreeMatch}}}
  \label{fig:supp:quali:c}
\end{subfigure}
&
\begin{subfigure}{\wimg\linewidth}
  \caption{ {Density} {Regression}}
  \label{fig:supp:quali:d}
\end{subfigure}
\end{tabular}
    \caption{{\bf Failure Case Illustrations.}
Rows, top to bottom: China (Gaofen2), Rwanda (PlanetScope), and France (SPOT6).
From left to right: input RGB image (\subref{fig:supp:quali:a}), strong manual annotations (\subref{fig:supp:quali:b}), \textsc{TreeMatch}'s predictions (\subref{fig:supp:quali:c}), density regression baseline (\subref{fig:supp:quali:d}).
The errors illustrate three common error sources: ambiguous visual patterns in heterogeneous landscapes (first row), local mass overestimation due to herbaceous terrain (second row), and cases where individual trees become indistinguishable within continuous canopy cover (third row).}
    \label{fig:sup:fail}
\end{figure}

\paragraph{Failure Cases.}
We illustrate several representative failure cases in \cref{fig:sup:fail}.
Most errors occur in dense forest regions where individual crowns are visually indistinguishable, or in heterogeneous landscapes where non-tree structures resemble tree crowns.

\begin{figure}
    \centering
     \captionsetup[subfigure]{%
   justification=centering,
   labelfont=small,
   textfont=small,
}
\def\wimg{0.21}
\scriptsize
\begin{tabular}{l@{\;\;}cc cc}
\scriptsize \rotatebox{90}{\qquad\shortstack{China \\ Gaofen2}}
&
\includegraphics[width=\wimg \linewidth, height= \wimg \linewidth, trim={2mm 2mm 2mm 2mm},clip]{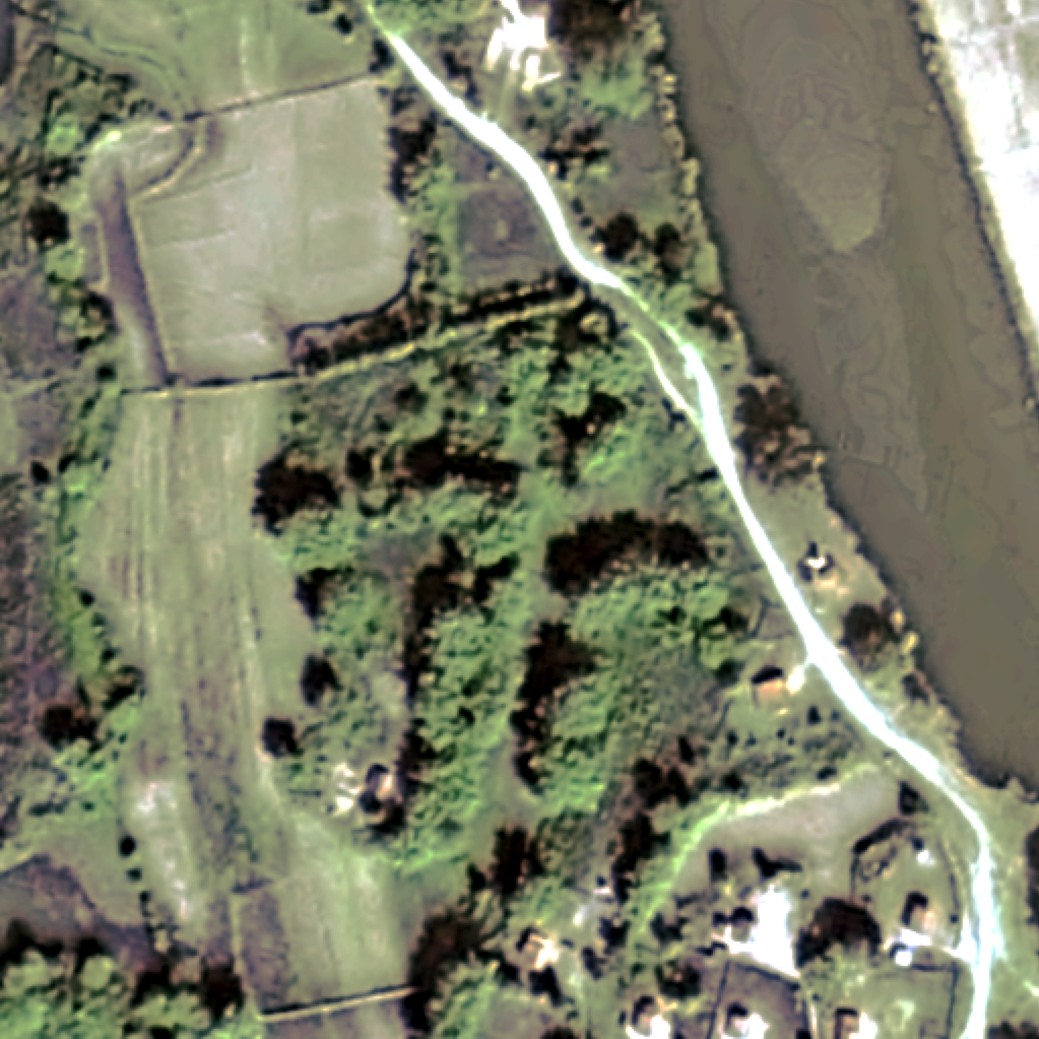}   
&
\includegraphics[width=\wimg \linewidth, height= \wimg \linewidth, trim={2mm 2mm 2mm 2mm},clip]{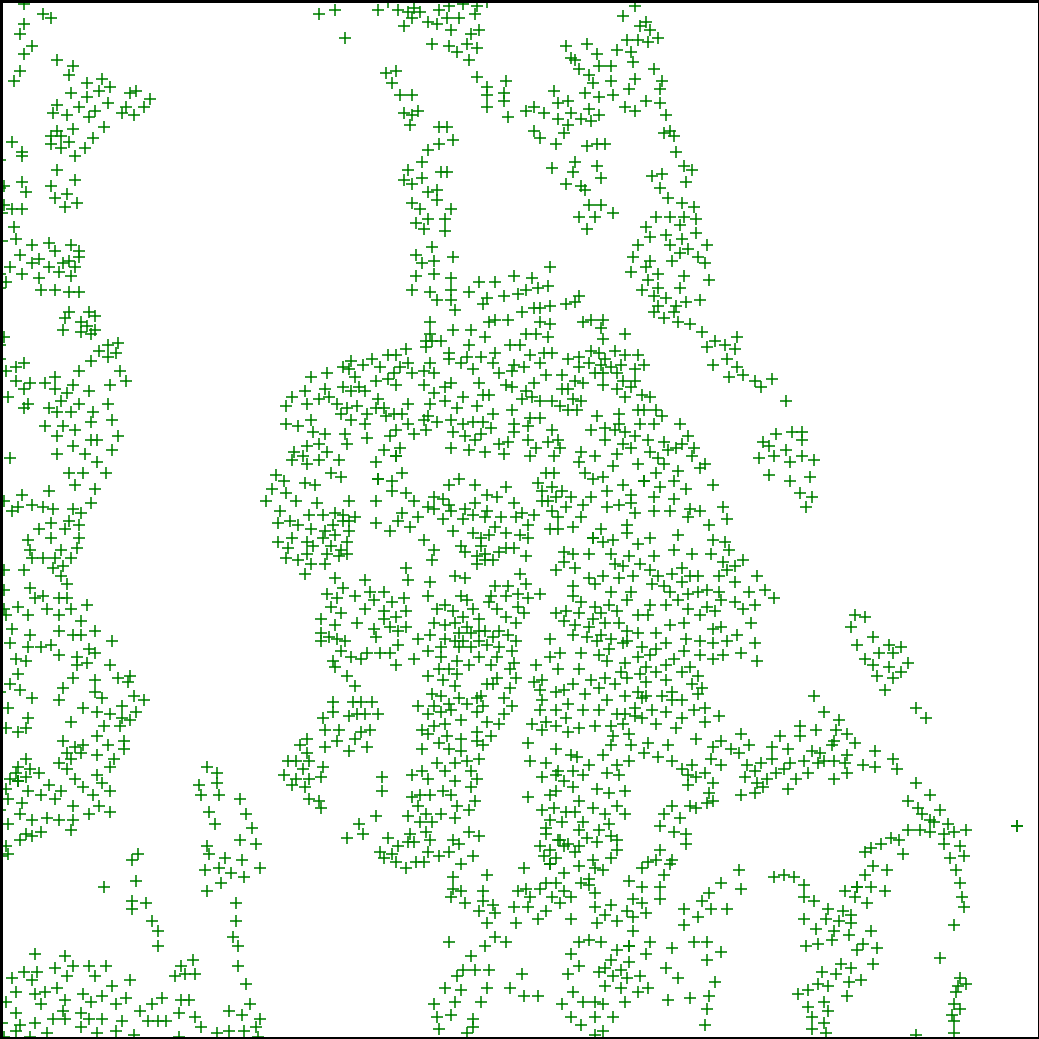} 
&
\includegraphics[width=\wimg \linewidth, height= \wimg \linewidth, trim={2mm 2mm 2mm 2mm},clip]{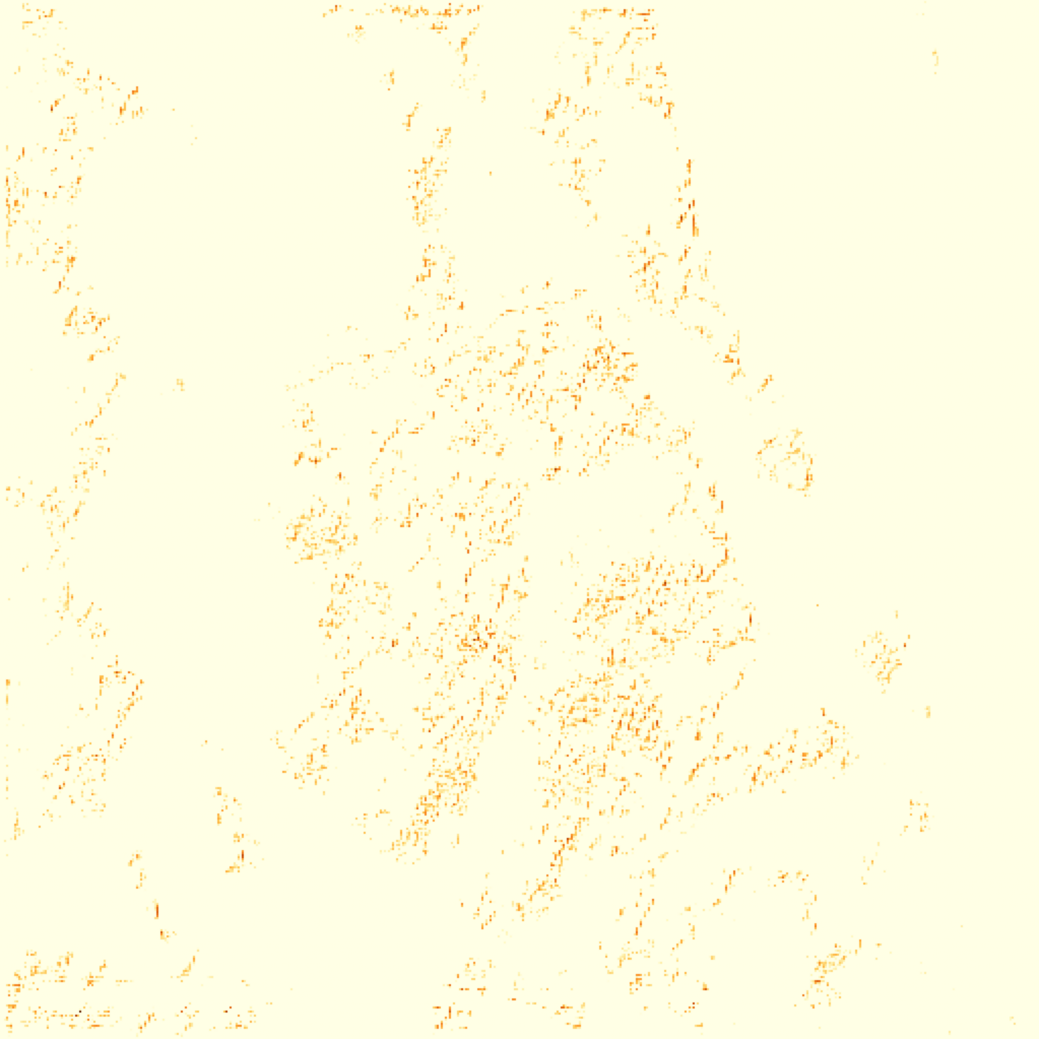} 
&
\includegraphics[width=\wimg \linewidth, height= \wimg \linewidth, trim={2mm 2mm 2mm 2mm},clip]{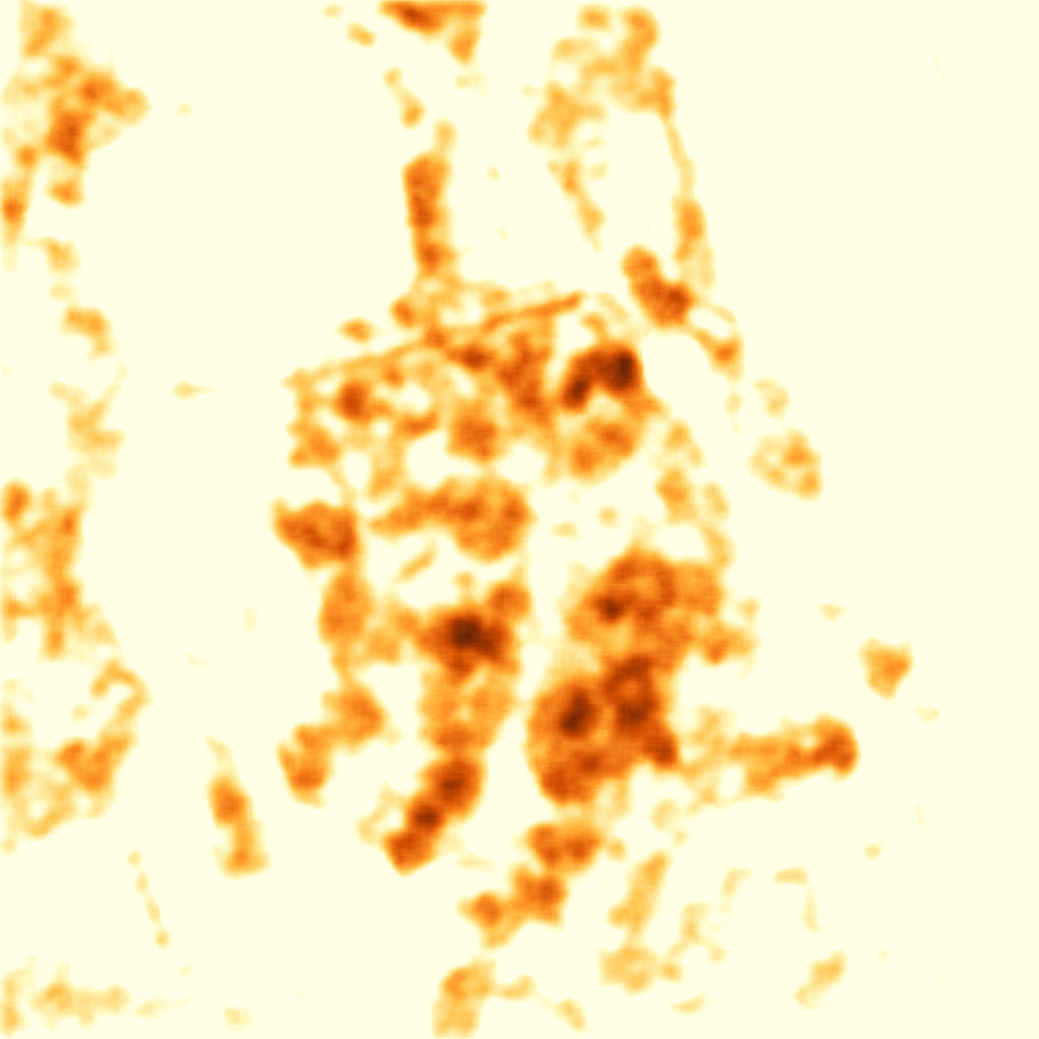} 
\\
& & 1918 trees & 1151.9 trees & 1100.8 trees
\\
\scriptsize \rotatebox{90}{\qquad\;\shortstack{Rwanda \\ PlanetScope}}
&
\includegraphics[width=\wimg \linewidth, height= \wimg \linewidth, trim={2mm 2mm 2mm 2mm},clip]{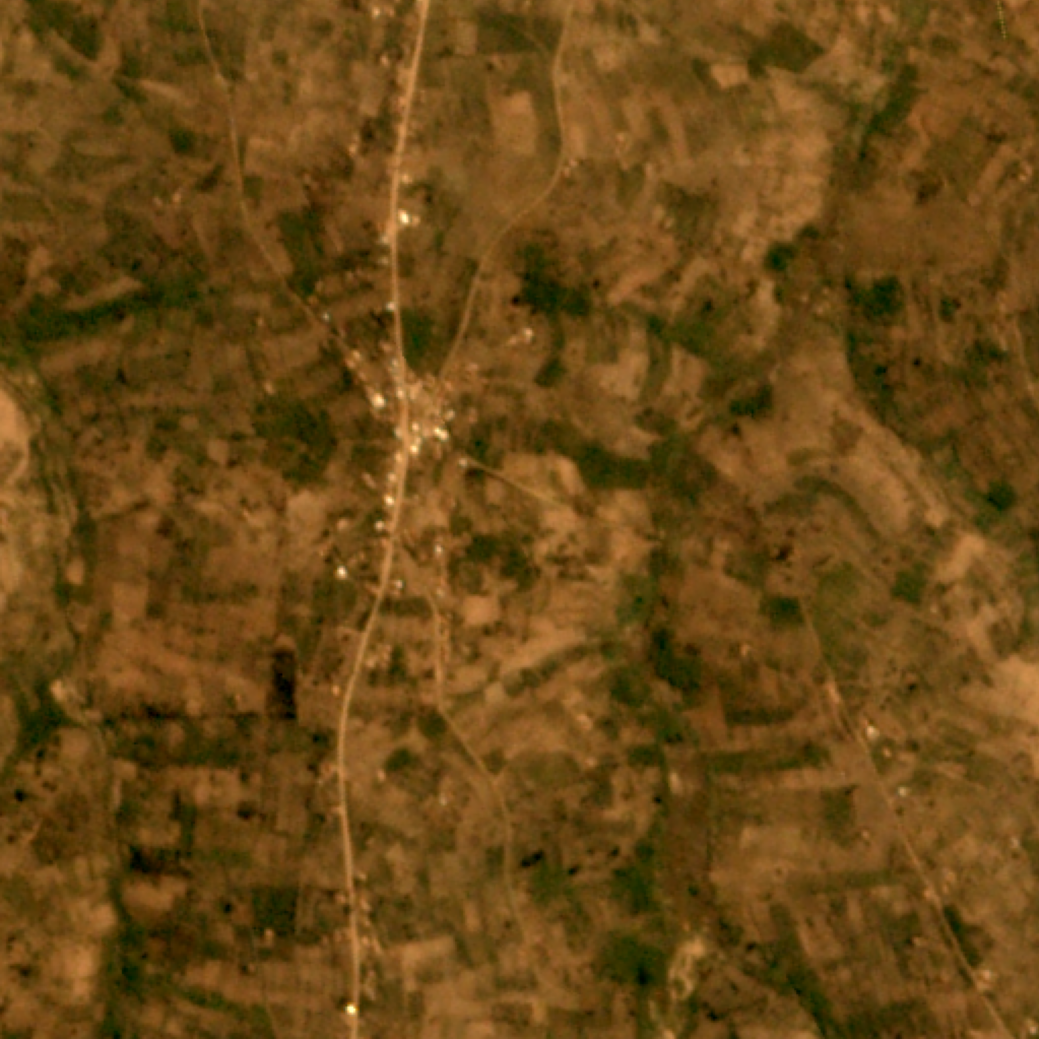}  
&
\includegraphics[width=\wimg \linewidth, height= \wimg \linewidth, trim={2mm 2mm 2mm 2mm},clip]{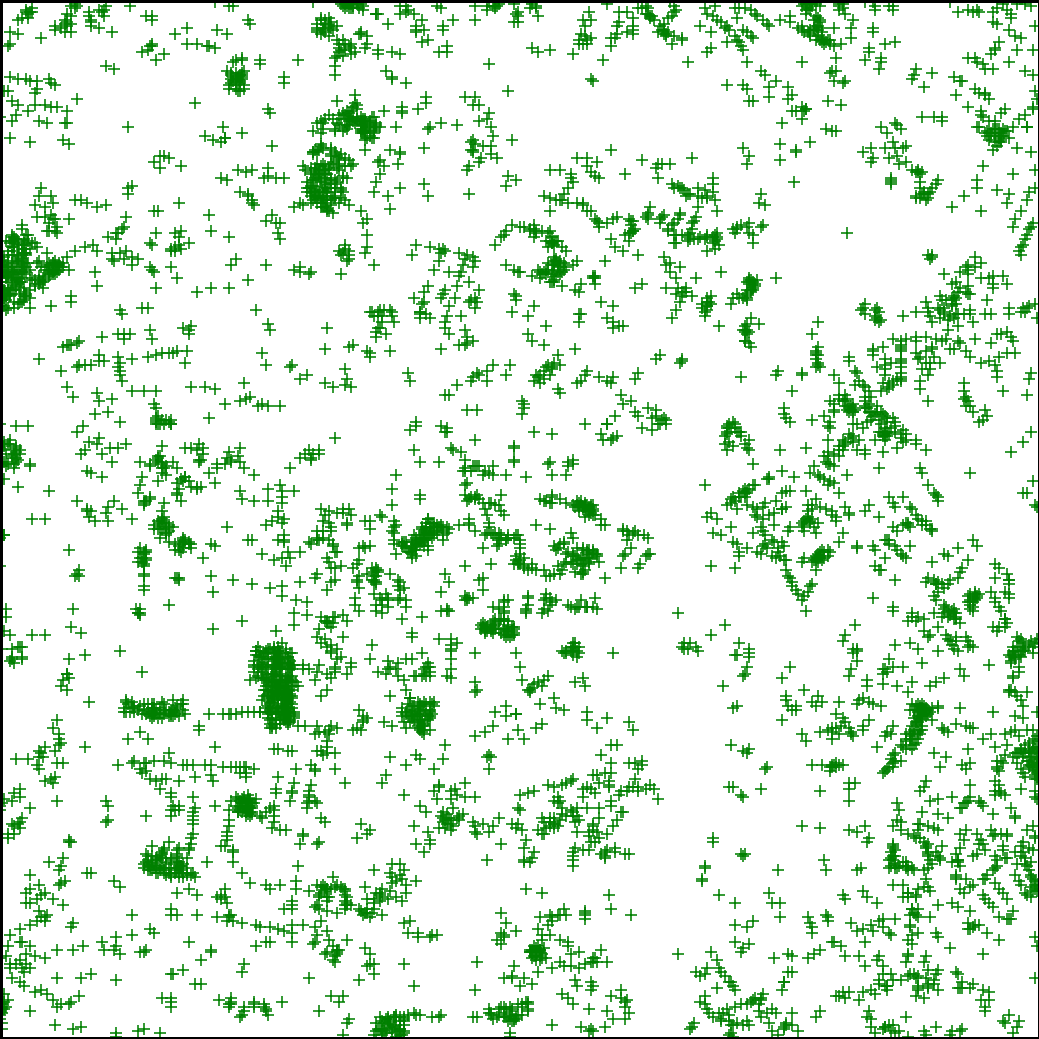} 
&
\includegraphics[width=\wimg \linewidth, height= \wimg \linewidth, trim={2mm 2mm 2mm 2mm},clip]{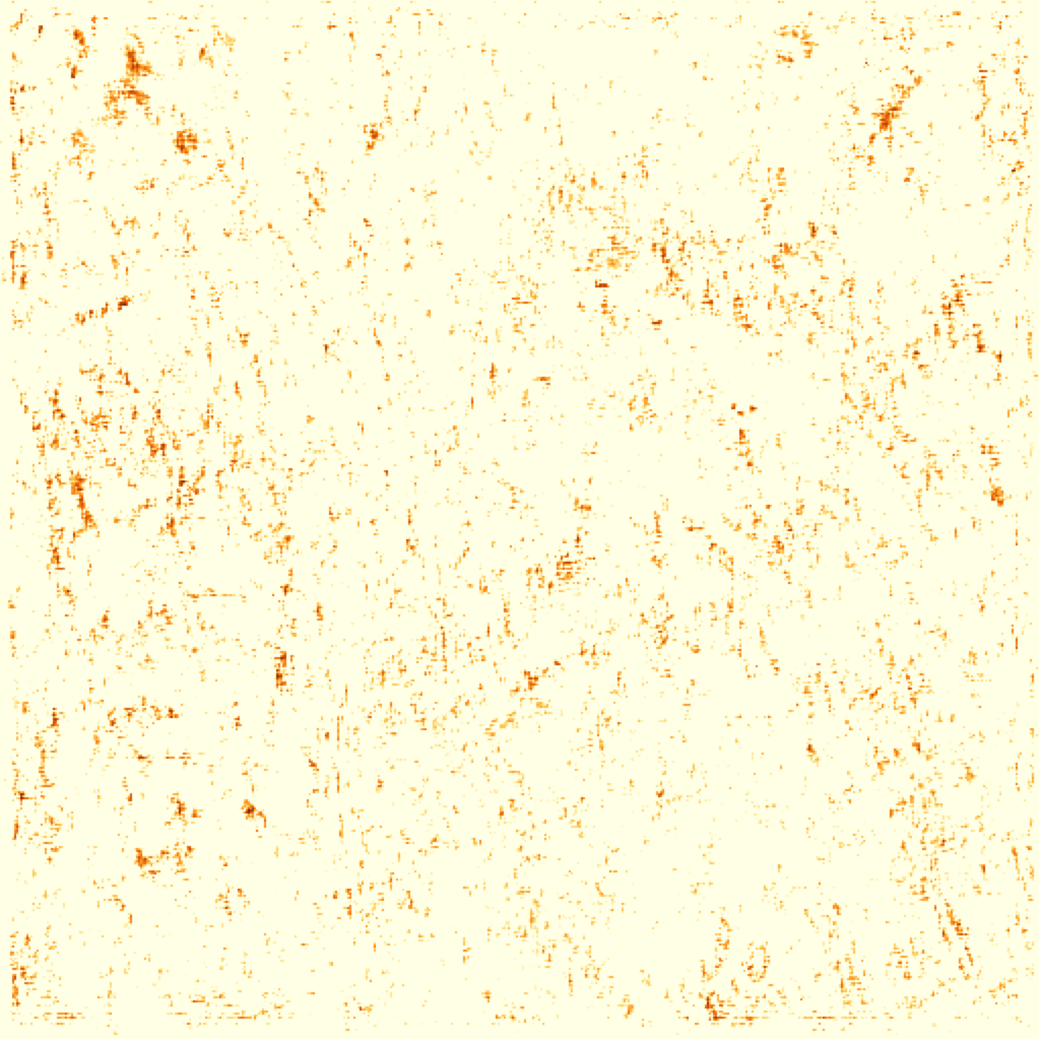} 
&
\includegraphics[width=\wimg \linewidth, height= \wimg \linewidth, trim={2mm 2mm 2mm 2mm},clip]{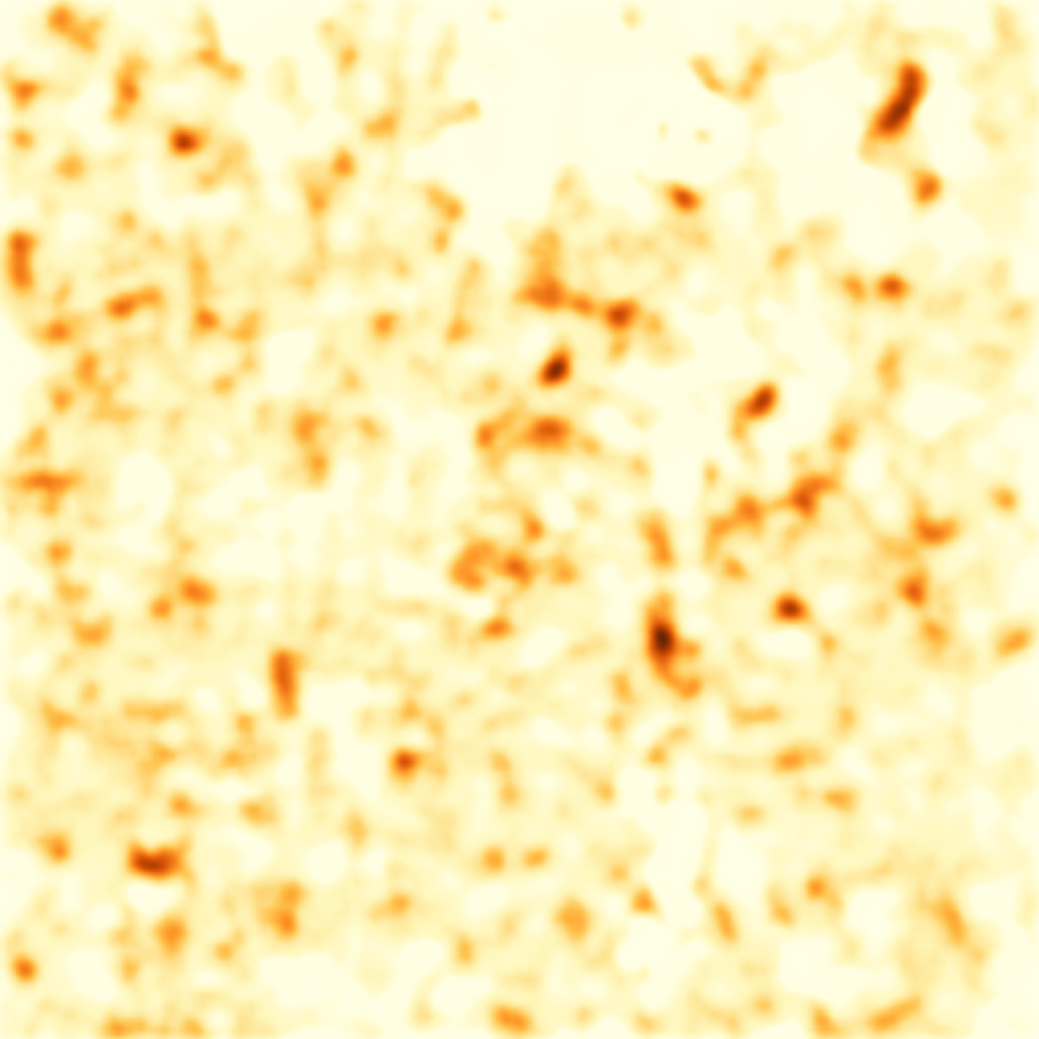} 
\\
& & 5555 trees & 5757.9 trees & 1021.0 trees\\
\scriptsize \rotatebox{90}{\qquad\;\;\shortstack{France \\ SPOT6}}
&
\includegraphics[width=\wimg \linewidth, height= \wimg \linewidth, trim={2mm 2mm 2mm 2mm},clip]{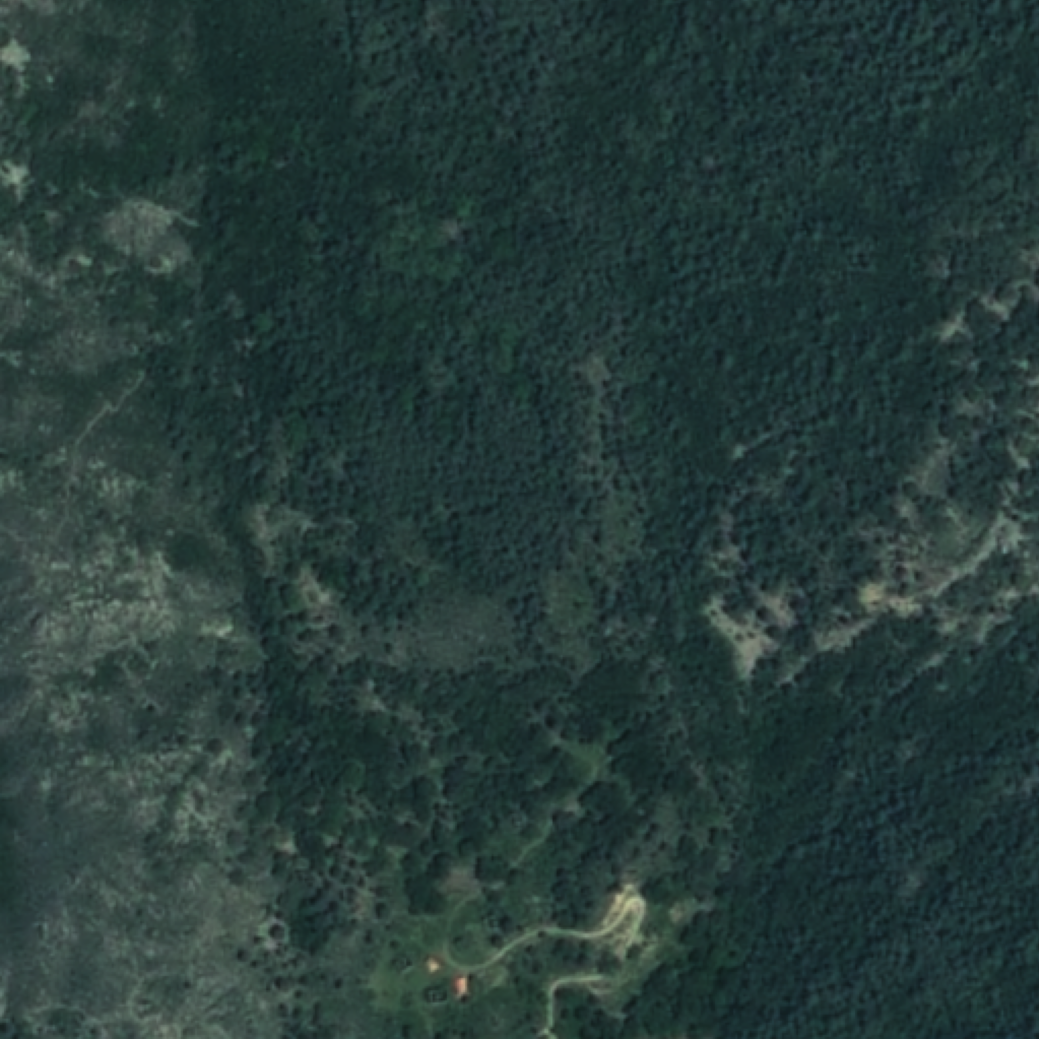}  
&
\includegraphics[width=\wimg \linewidth, height= \wimg \linewidth, trim={2mm 2mm 2mm 2mm},clip]{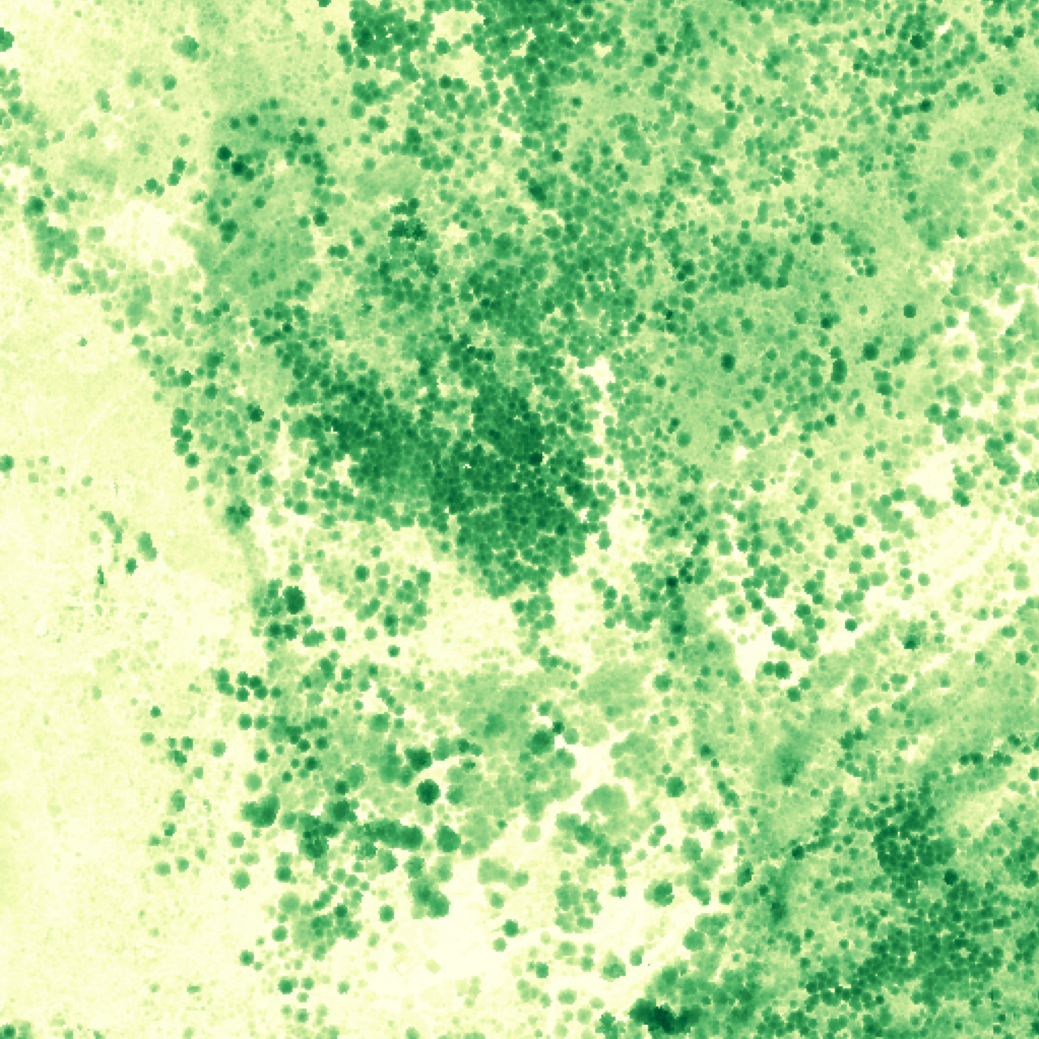} 
&
\includegraphics[width=\wimg \linewidth, height= \wimg \linewidth, trim={2mm 2mm 2mm 2mm},clip]{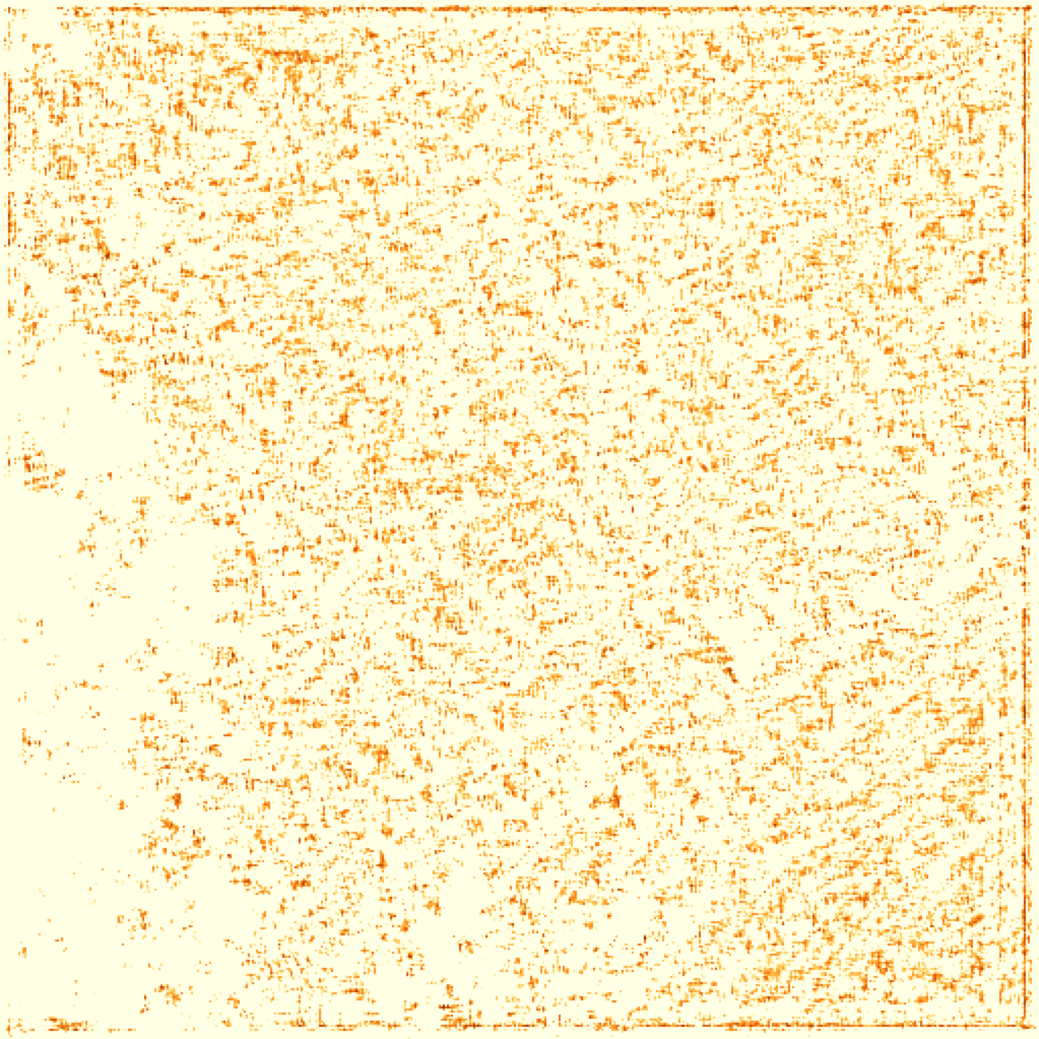} 
&
\includegraphics[width=\wimg \linewidth, height= \wimg \linewidth, trim={2mm 2mm 2mm 2mm},clip]{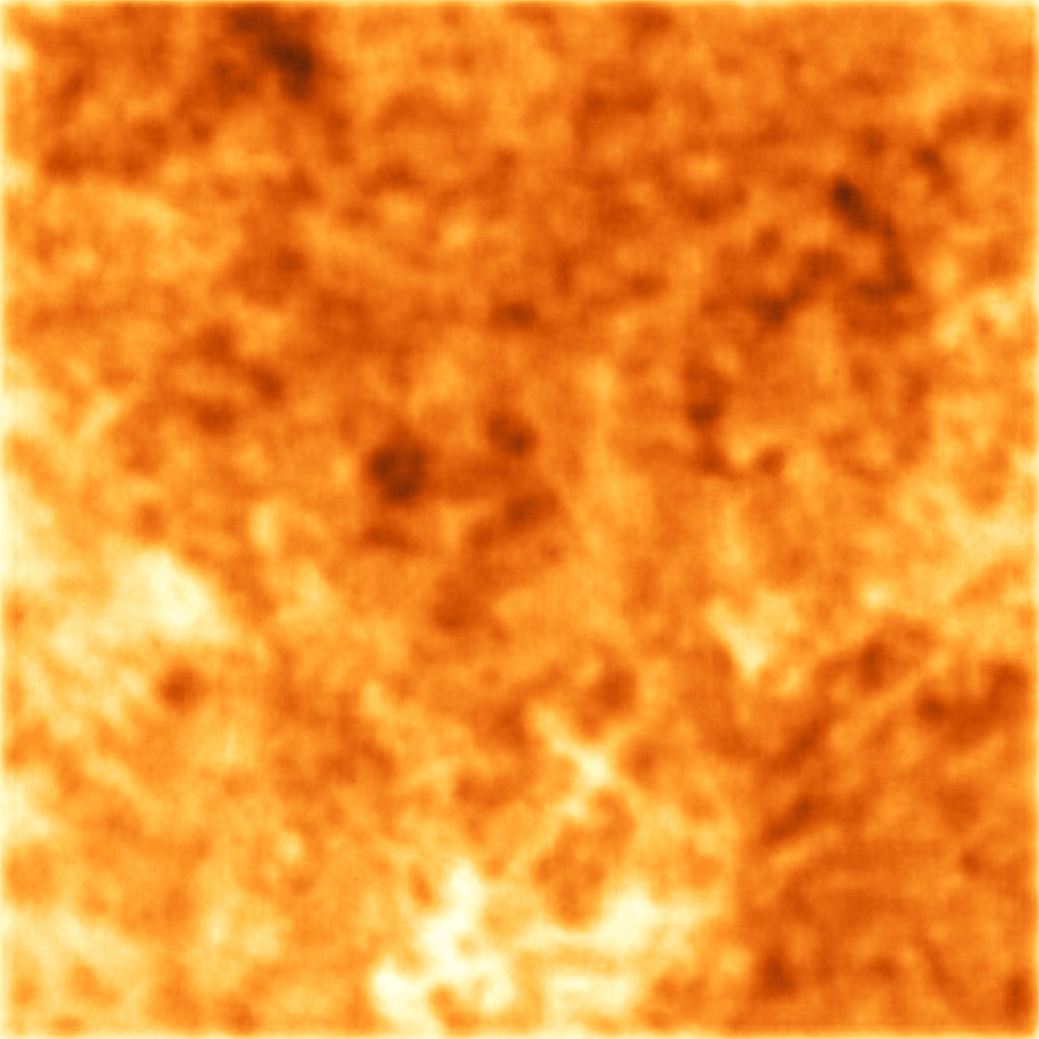} 
\\
& & - & 8723.9 trees & 1095.2 trees\\
&
\begin{subfigure}{\wimg\linewidth}
  \caption{{Input} {RGB}}
  \label{fig:supp:quali:a}
\end{subfigure}
&
\begin{subfigure}{\wimg\linewidth}
  \caption{{Strong} {labels}}
  \label{fig:supp:quali:b}
\end{subfigure}
&
\begin{subfigure}{\wimg\linewidth}
  \caption{{\textsc{TreeMatch}}}
  \label{fig:supp:quali:c}
\end{subfigure}
&
\begin{subfigure}{\wimg\linewidth}
  \caption{ {Density} {Regression}}
  \label{fig:supp:quali:d}
\end{subfigure}
\end{tabular}

    \caption{{\bf Extended qualitative comparison.}
Rows, top to bottom: China (Gaofen2), Rwanda (PlanetScope), and France (SPOT6).
From left to right: input RGB image (\subref{fig:supp:quali:a}), strong manual annotations (\subref{fig:supp:quali:b}), \textsc{TreeMatch}'s predictions (\subref{fig:supp:quali:c}), density regression baseline (\subref{fig:supp:quali:d}). For the France split, ground truth is only avialbale on a small inventory plot. We represent the canopy height map from OpenCanopy \cite{fogel_open-canopy_2025} instead.}
    \label{fig:sup:quali}
\end{figure}

\paragraph{Qualitative analysis.}
We plot predictions on wider (128x128) images in \cref{fig:sup:quali}. SPOT6 imagery, being of higher quality, leads to more localized predictions. Using a fully convolutional backbone such as UNet-R50 allows scaling up to arbitrary image sizes for inference.

\section{Additional Implementation Details}
\label{sec:sup:details}

\paragraph{\textbf{Metric Definitions.}}
We assess counting accuracy using the following metrics:
\begin{enumerate}
    \item \textbf{RMSE.} Image-level root mean squared error on tree counts, in \emph{trees per hectare}.
    \begin{align}
        \mathrm{RMSE}
        =
        \frac{10^4}{HW{\Delta x}^2}
        \sqrt{
        \frac{1}{n}
        \sum_{i=1}^n
        \left( y_i - z_i \right)^2
        }~,
    \end{align} with $\Delta x$ the pixel size in meters and $H, W$ image height and width.
    
    \item \textbf{nMAE.} Dataset-level normalized mean absolute error.
    \begin{align}
        \mathrm{nMAE}
        =
        \frac{1}{n}
        \sum_{i=1}^n
        \frac{\left| y_i - z_i \right|}{\bar{y}}~,
    \end{align}
    with \(\bar{y}\) the mean ground-truth count.
    
    \item \textbf{\bf $\mathbf{R^2}$.} Coefficient of determination between predicted and ground-truth counts, computed at the image-level.
    \begin{align}
    R^2
    =
    1
    -
    \frac{
        \sum_{i=1}^n (y_i - z_i)^2
    }{
        \sum_{i=1}^n (y_i - \bar{y})^2
    }~,
\end{align} 
\end{enumerate}

\paragraph{\textbf{Computational Cost and Scalability.}}
Training \textsc{TreeMatch} requires solving entropy-regularized optimal transport problems.
Thanks to an efficient Sinkhorn implementation, this remains tractable in practice: training \textsc{TreeMatch} on our largest split (Rwanda) takes $27$ minutes on a single RTX 3090 GPU, compared to $18$ minutes for direct regression and $1$h$40$ for the OT-based DM-Count baseline. \Cref{fig:sup:trainstep} indicates that this trend is consistent accross batch sizes, with \textsc{TreeMatch} introducing a moderate overhead over regression while remaining significantly faster than DM-Count.
Importantly, our method does not affect inference: prediction requires only a standard forward pass.

\section{Pseudo-labels Qualification}
\label{sec:sup:qualif}
We evaluate the quality of the weak, automated tree labels by comparing them with our strong annotations on the France split. 
\Cref{table:pseudolabels} reports the parameters used for extracting automated tree labels from the canopy height model via peak detection: the minimum canopy height threshold (\emph{Thresh.}) and the minimum distance allowed between two detected tree tops (\emph{Min. dist.}). 

This simplistic procedure produces a highly noisy signal. When compared with field-collected tree positions, the resulting labels yield an $R^2$ close to zero. In isolation, this signal is therefore largely insufficient for accurate tree counting. Nonetheless, our experiments show that this weak supervision still improves training when used within our self-correcting training scheme (see Table~3 and the ablation study in the main paper).

\paragraph{Temporal / Resolution Match.}
Tree annotations are point annotations, which we rasterize at the spatial resolution of the corresponding satellite imagery.
Temporal alignment varies across regions: annotations are contemporaneous with the imagery for the China and Rwanda splits, while in France the field measurements can precede the imagery by up to three years due to the lower frequency of in situ acquisition campaigns.


\begin{table*}
\centering
\caption{{\bf Qualification of Weak Labels.} Parameters and validation metrics for CHM pseudo-label extraction.}
\begin{tabular}{cccccccc}
 \toprule
Sensor & Split & Thresh. (m) & Min. dist. (m) & RMSE & R2 & nMAE \\
\midrule
\multirow{1}{*}{Spot6} & Train-weak & 3 & 4 & 236.3 & 0.07 & 47.0\\\bottomrule
  \end{tabular}
\label{table:pseudolabels}
\end{table*}
\end{document}